\documentclass[times, review, 10pt]{elsarticle}
\usepackage{amssymb}
\usepackage{a4wide}
\usepackage{graphicx}
\usepackage{amsmath}
\usepackage{longtable}
\usepackage{algorithm} 
\usepackage{algpseudocode} 
\usepackage{mathrsfs}
\usepackage{subcaption}
\usepackage{hyperref}
\usepackage{mathtools}
\usepackage{pifont}
\usepackage{color}
\usepackage{lineno}
\usepackage{makecell} 
\usepackage{pdflscape}
\usepackage{adjustbox}
\usepackage[utf8]{inputenc}
\usepackage{tabularx}
\usepackage{blindtext}
\usepackage{setspace}
\usepackage{notoccite} 
\usepackage{lscape} 
\usepackage{caption} 
\usepackage{mwe}
\usepackage{booktabs}
\usepackage{amsthm}
\newtheorem{theorem}{Theorem}[section]

\newcommand{\RNum}[1]{\lowercase\expandafter{\romannumeral #1\relax}}
\newcommand{\RNumU}[1]{\uppercase\expandafter{\romannumeral #1\relax}}
\usepackage{natbib}
\journal{Pattern Recognition, Elsevier: 28 November 2023; revised 24 May 2024; accepted 28 May 2024.}

\begin{document}
\date{}

\begin{frontmatter}
\title{Advancing Supervised Learning with the Wave Loss Function: A Robust and Smooth Approach}
\author[inst1]{Mushir Akhtar}
\author[inst1]{M. Tanveer\corref{Correspondingauthor}}
\author[inst1]{Mohd. Arshad}
\author[]{for the Alzheimer’s Disease Neuroimaging
Initiative\corref{ADNI information}}
\affiliation[inst1]{organization={Department of Mathematics},
            addressline={Indian Institute of Technology Indore}, 
            city={Simrol, Indore},
            postcode={453552}, 
            country={India}}
            \cortext[Correspondingauthor]{Corresponding author}
            \cortext[ADNI information]{The data for this article were sourced from the Alzheimer’s Disease Neuroimaging Initiative (ADNI) database (adni.loni.usc.edu). While investigators within the ADNI contributed to the design and implementation of ADNI and provided data, they were not involved in the analysis or writing of this report.}

\begin{abstract}
Loss function plays a vital role in supervised learning frameworks. The selection of the appropriate loss function holds the potential to have a substantial impact on the proficiency attained by the acquired model. The training of supervised learning algorithms inherently adheres to predetermined loss functions during the optimization process. In this paper, we present a novel contribution to the realm of supervised machine learning: an asymmetric loss function named wave loss. It exhibits robustness against outliers, insensitivity to noise, boundedness, and a crucial smoothness property. Theoretically, we establish that the proposed wave loss function manifests the essential characteristic of being classification-calibrated. Leveraging this breakthrough, we incorporate the proposed wave loss function into the least squares setting of support vector machines (SVM) and twin support vector machines (TSVM), resulting in two robust and smooth models termed as Wave-SVM and Wave-TSVM, respectively. To address the optimization problem inherent in Wave-SVM, we utilize the adaptive moment estimation (Adam) algorithm, which confers multiple benefits, including the incorporation of adaptive learning rates, efficient memory utilization, and faster convergence during training. It is noteworthy that this paper marks the first instance of Adam's application to solve an SVM model.
Further, we devise an iterative algorithm to solve the optimization problems of Wave-TSVM. To empirically showcase the effectiveness of the proposed Wave-SVM and Wave-TSVM, we evaluate them on benchmark UCI and KEEL datasets (with and without feature noise) from diverse domains. Moreover, to exemplify the applicability of Wave-SVM in the biomedical domain, we evaluate it on the Alzheimer’s Disease Neuroimaging Initiative (ADNI) dataset. The experimental outcomes unequivocally reveal the prowess of Wave-SVM and Wave-TSVM in achieving superior prediction accuracy against the baseline models. The source codes of the proposed models are publicly available at \url{https://github.com/mtanveer1/Wave-SVM}.

\end{abstract}

\end{frontmatter}

\section{Introduction and Motivation}
In the realm of machine learning, supervised learning stands as a cornerstone, which empowers to build models that can make accurate predictions and classifications. A critical component within this paradigm is the choice of an appropriate loss function, which plays a crucial role in guiding the training process. The loss function quantifies the disparity between predicted outcomes and true labels, providing an evident indication of a model's performance \cite{wang2020comprehensive}. Its role extends beyond mere evaluation; it fundamentally shapes the model's learning trajectory, influencing its generalization ability, and aligning its predictions with the intrinsic goals of the task.
\par
Support Vector Machine (SVM) \cite{cortes1995support} emblematize a stalwart supervised learning algorithm.
It is grounded in the principle of structural risk minimization (SRM) and originates from statistical learning theory (SLT), consequently having a solid theoretical foundation and demonstrating better generalization capabilities.
It has extensive applications across various domains, such as image classification \cite{han2023ml}, wind speed prediction \cite{hu2022novel}, face recognition \cite{lopez2022incremental}, handwritten digit recognition \cite{niu2012novel}, and so forth.
\par
Consider the training set $ \mathcal{D}=\left\{x_i,y_i\right\}_{i=1}^l$, where $x_i \in \mathbb{R}^n$ represents the sample vector and $y_i \in\{-1,1\}$ signifies the corresponding class label associated with the sample. The core idea that serves as the foundation of SVM involves the construction of a decision hyperplane $w^\intercal x+b=0$, where $b \in \mathbb{R}$ represents the bias and $w \in \mathbb{R}^n$ signifies the weight vector. These parameters are estimated through the process of training using available data. For a test data point $\widetilde{x}$, the associated class label $\widetilde{y}$ is predicted as $1$ if $w^\intercal \widetilde{x}+b \geq 0$ and $-1$ otherwise. The pursuit of an optimal hyperplane hinges upon two distinct scenarios within the input space: first, when training data is linearly separable, and second, when training data is linearly inseparable.
\par
For a linearly separable scenario, the process of attaining the optimal parameters $w$ and $b$ involves addressing the following SVM model:
\begin{align} \label{hardmarginSVM}
\underset{ w, b}{min} \hspace{0.2cm} &\frac{1}{2}\|w\|^2 \nonumber \\
 \text { subject to }\hspace{0.2cm}  & y_i\left(w^\intercal x_i+b\right) \geq 1, ~\forall~ i=1,2, \ldots,l.
\end{align}
The model described in equation (\ref{hardmarginSVM}) is named as hard-margin SVM, as it mandates the correct classification of each individual training sample. In cases where the data is linearly inseparable, the commonly employed strategy involves allowing for misclassification while imposing penalties for these errors. This is achieved by incorporating a loss function into the objective function, which leads to the subsequent unconstrained optimization problem:
\begin{align} \label{softmarginSVM}
\underset{ w, b}{min} \hspace{0.2cm} &\frac{1}{2}\|w\|^2 + C \sum_{i=1}^l \mathfrak{L}\biggl(1-y_i \left(w^\intercal x_i+b\right) \biggr), 
\end{align}
where $C > 0$ is a trade-off parameter and $\mathfrak{L}(u)$ with $u$:$=1-y_i \left(w^\intercal x_i+b\right)$ denotes the loss function. Since model (\ref{softmarginSVM}) permits the misclassification of samples, it is identified as a soft-margin SVM model \cite{cortes1995support}. The objective of SVM is to obtain an optimal hyperplane that separates data points of distinct classes by maximum possible margin. This primarily involves solving a quadratic programming problem (QPP) whose complexity is proportional to the cube of the training dataset size. Twin SVM (TSVM) \cite{khemchandani2007twin}, a variant of the SVM,  tackles this problem by solving two smaller QPPs instead of one large QPP, thereby reducing computational costs by approximately $75\%$ compared to traditional SVM methods. To enhance the generalization performance of TSVM, several variants have been proposed in the literature. As an example, to tackle the imbalance problem, \citet{ganaie2022large} proposed the large-scale fuzzy least squares TSVM. Further, to address the noise-sensitivity of TSVM \citet{tanveer2022large} proposed the large-scale pin-TSVM by utilizing the pinball loss function. Both of the aforementioned algorithms eliminate the requirement of matrix inversion, making them suitable for large-scale problems. To delve deeper into the variants of TSVM models, readers can refer to \cite{tanveer2022comprehensive}.

Extensive research endeavors have been dedicated to the development of novel loss functions for the purpose of enhancing the effectiveness of SVM models. Based on the smoothness of loss functions, we classify them into two distinct classes. The first category encompasses loss functions characterized by their smoothness, while the latter consists of those with non-smooth attributes. Here, we only review a subset of widely recognized loss functions, chosen to provide sufficient motivation for the work presented in this paper.
\subsection{Non-smooth loss functions}
\begin{figure*}
\centering
    \subcaptionbox{     \label{fig:hinge }} { %
      \includegraphics[width=0.48\textwidth,keepaspectratio]{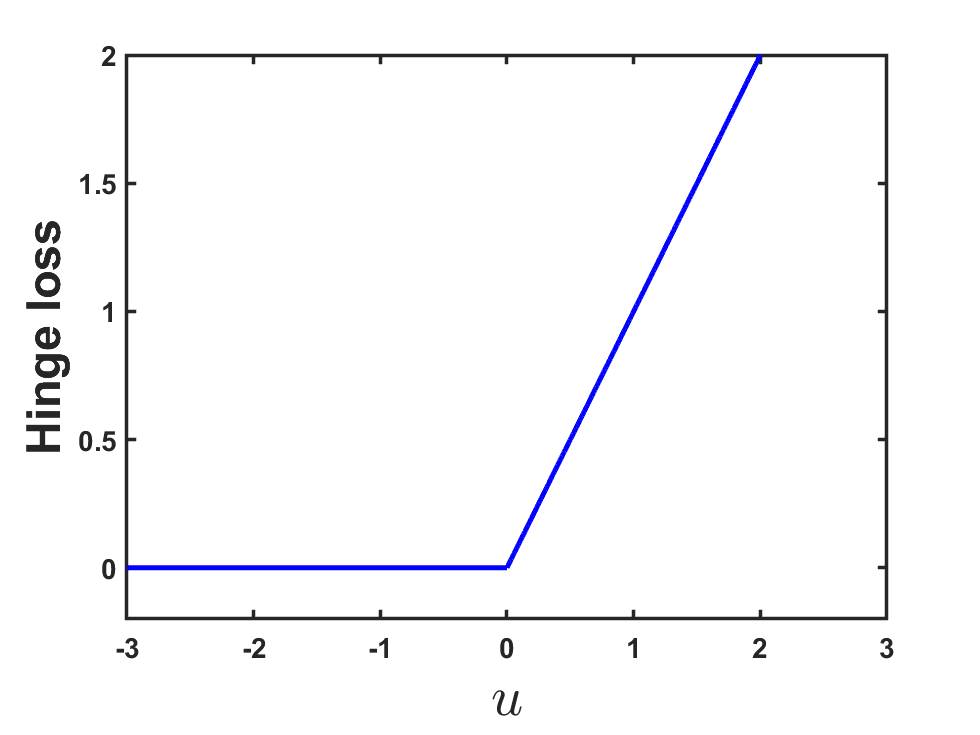}}
      \hfill
      \subcaptionbox{   \label{fig:pinball }} { %
      \includegraphics[width=0.48\textwidth,keepaspectratio]{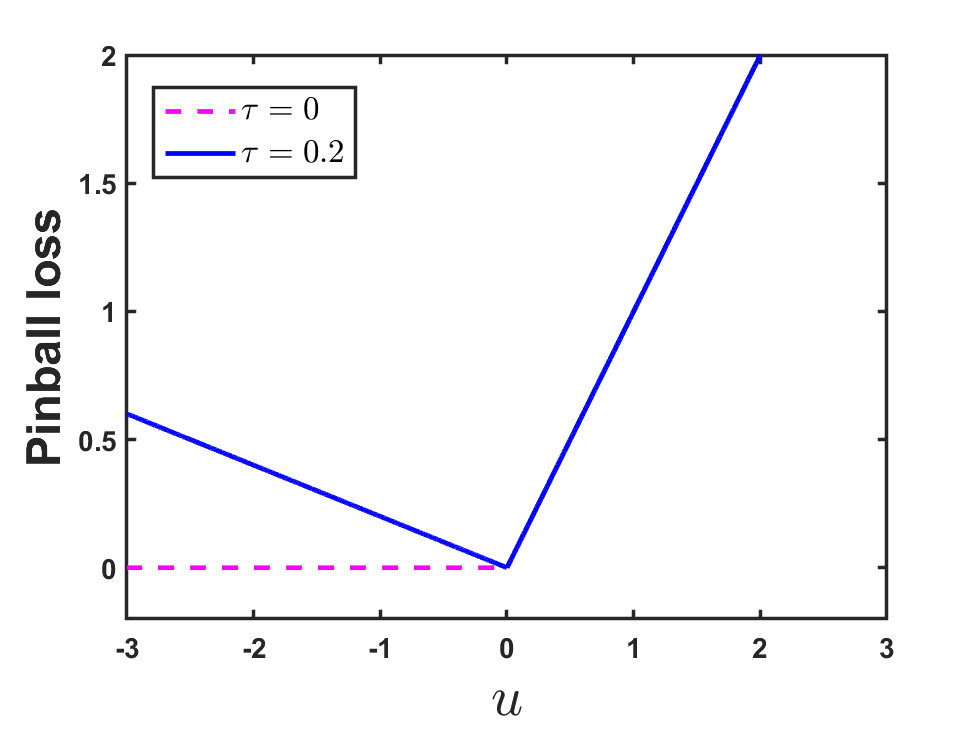}}
\\
      \subcaptionbox{  \label{fig:ramp loss}} { %
      \includegraphics[width=0.48\textwidth,keepaspectratio]{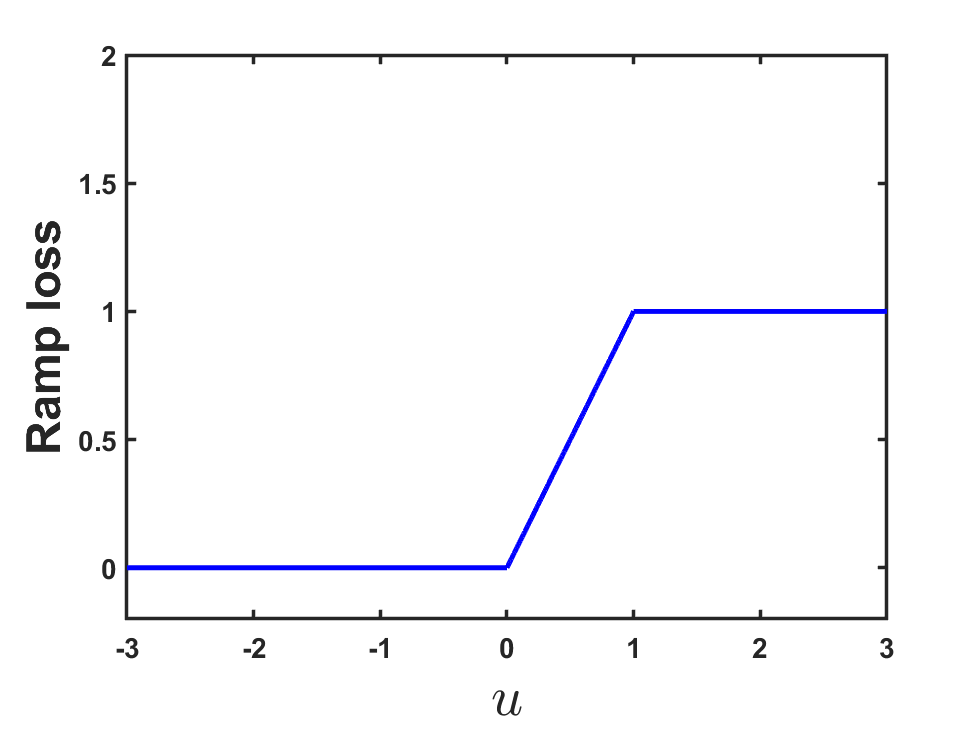}}
      \hfill
      \subcaptionbox{  \label{fig:squared hinge}} { %
      \includegraphics[width=0.48\textwidth,keepaspectratio]{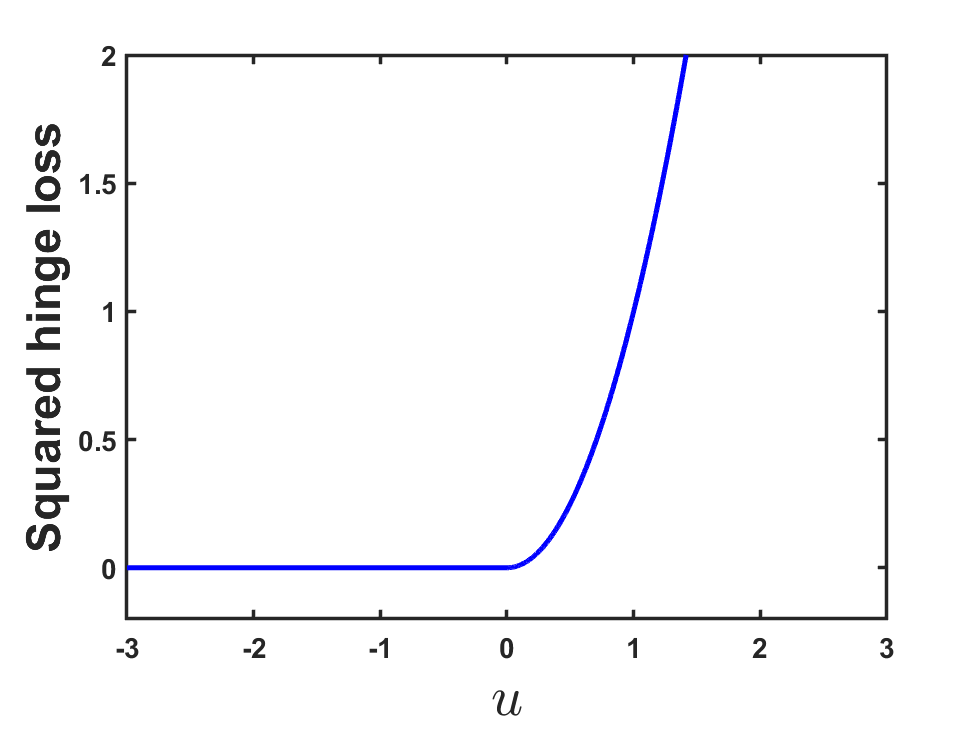}}
\\
    \subcaptionbox{     \label{fig:smoothpin }} { %
      \includegraphics[width=0.48\textwidth,keepaspectratio]{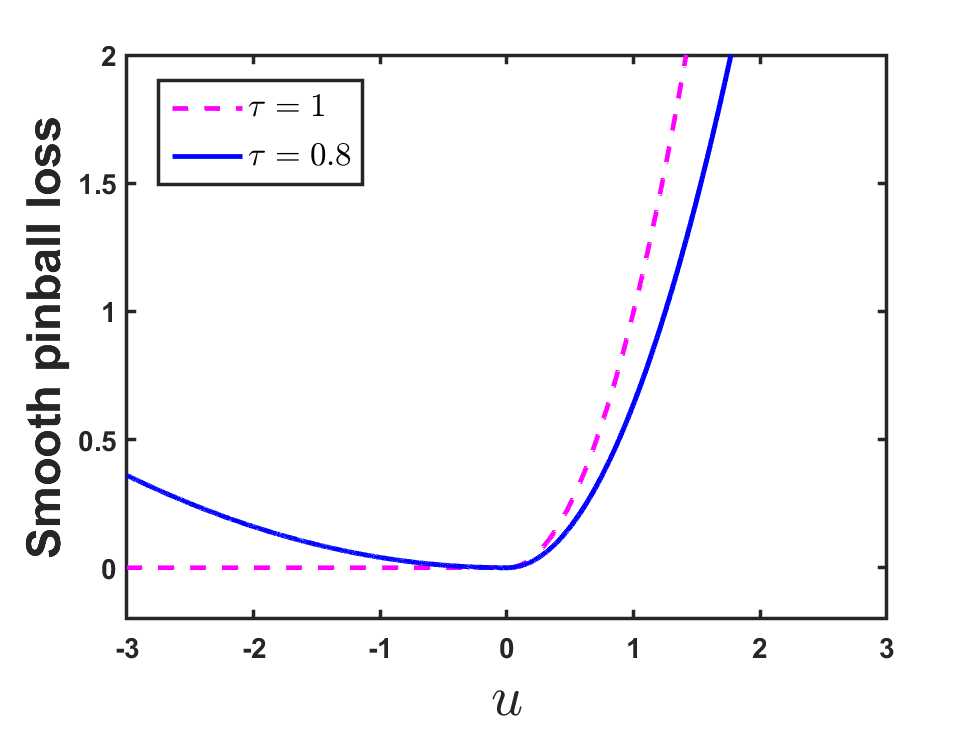}}
      \hfill
      \subcaptionbox{   \label{fig:linex }} { %
      \includegraphics[width=0.48\textwidth,keepaspectratio]{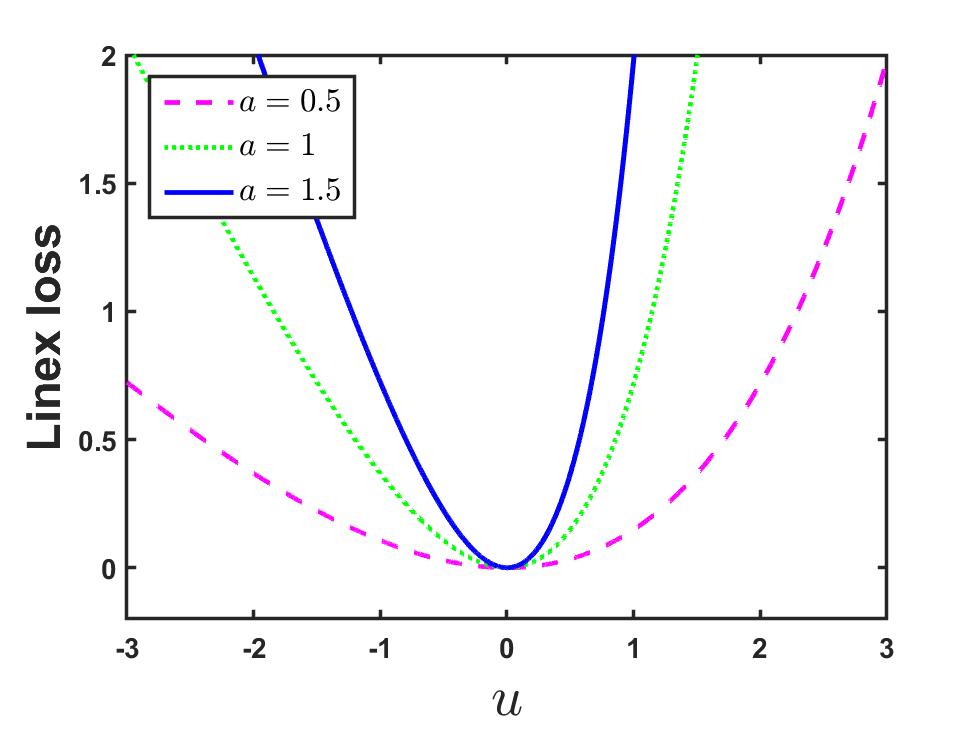}}
      \caption{Visual illustration of baseline loss functions. (a) Hinge loss function. (b) Pinball loss function with $\tau=0$ and $\tau=0.2$. (c) Ramp loss function with $\theta=1$. (d) Squared hinge loss function (e) Smooth pinball loss function with $\tau=0.8$ and $\tau=1$. (f) LINEX loss function with $a=0.5$, $a=1$, and $a=1.5$.}
    \label{fig:Baseline Loss functions}
 \end{figure*}

\begin{itemize}
    \item \textbf{Hinge loss function:} SVM with hinge loss function (C-SVM) was first proposed by \citet{cortes1995support}. The mathematical expression for the hinge loss function is given as:
\begin{align}
\mathfrak{L}_{hinge}(u)=
\begin{cases}
u, & u > 0, \\
0, & u \leq 0. 
\end{cases}
\end{align}
Figure \ref{fig:hinge } depicts the visual representation of the hinge loss function. It is non-smooth at $u=0$ and unbounded.
\item \textbf{Pinball loss function:} To improve the efficacy of C-SVM, \citet{huang2013support} proposed SVM with pinball loss function (Pin-SVM). The mathematical formulation of the pinball loss is expressed as:
\begin{align}
\mathfrak{L}_{pin}(u)=
\begin{cases}
u, & u > 0, \\
-\tau u, & u \leq 0, 
\end{cases}
\end{align}
where $\tau \in \left[0,1\right]$. It penalizes both correctly classified and misclassified samples so that if there is noise near the decision boundary, it can be adjusted to strike a trade-off between accuracy and noise insensitivity. For $\tau=0$, the pinball loss function (See Figure \ref{fig:pinball }) is reduced to the hinge loss function. It is also non-smooth at $u=0$ and unbounded.

\item \textbf{Ramp loss function:} To enhance the robustness of C-SVM against outliers, \citet{brooks2011support} incorporate the ramp loss function into the SVM setting. The mathematical representation of the ramp loss function is articulated as follows: 
\begin{align}    
\mathfrak{L}_{ramp}(u)=
\begin{cases}
\theta, & u \geq \theta, \\
u, & u \in \left(0,\theta\right),\\
0, & u \leq 0,
\end{cases}
\end{align}
where $\theta \geq 1$. It imposes a limitation on penalties beyond a specific threshold. The ramp loss function (see Figure \ref{fig:ramp loss}) is also non-smooth at $u=0$ but bounded.
\end{itemize}
Some other non-smooth loss functions that have been incorporated into the framework of SVM to improve its efficiency include the rescaled hinge loss \cite{xu2017robust}, generalized ramp loss \cite{wang2024fastC}, flexible pinball loss \cite{kumari2024diagnosis}, and so forth.

\subsection{Smooth loss functions}
\begin{itemize}
\item \textbf{Squared hinge loss function:} To improve the smoothness of the hinge loss function, \citet{cortes1995support} also designed the smooth version of the hinge loss function termed as squared hinge loss. The mathematical expression of the squared hinge loss function is expressed as:
\begin{align}
\mathfrak{L}_{s-hinge}(u)=
\begin{cases}
u^2, & u > 0, \\
0, & u \leq 0. 
\end{cases}
\end{align}
Figure \ref{fig:squared hinge} presents the visual illustration of the squared hinge loss function. It is smooth and unbounded.    
\item \textbf{Smooth pinball loss function:} To enhance the insensitivity to noise of squared hinge loss function, \citet{liu2021smooth} designed the smooth pinball loss function. The mathematical formulation of the smooth pinball loss is given as:
\begin{align}
\mathfrak{L}_{s-pin}(u)=
\begin{cases}
(\tau u)^2, & u > 0, \\
\left[(1-\tau) u\right]^2, & u \leq 0, 
\end{cases}
\end{align}
where $\tau \in \left[0,1\right]$. For $\tau=1$, the 
 smooth pinball loss function (see Figure \ref{fig:smoothpin }) is reduced to the squared hinge loss function. It is also smooth and unbounded.

\item \textbf{LINEX loss function:} LINEX loss is an asymmetric function that combines linear and exponential components \cite{parsian2002estimation}. Due to the merits of LINEX loss, \citet{ma2019linex} first incorporated it into the SVM framework and introduced a fast SVM model named LINEX-SVM. The mathematical expression for the LINEX loss is given as:
\begin{align} \label{Linexloss}   
\mathfrak{L}_{LINEX}(u)= e^{au}-au-1, ~\forall~ u \in \mathbb{R},
\end{align}
where $a \neq 0$ is the loss parameter that controls the penalty for classified and misclassified samples. Figure \ref{fig:linex } presents the visual illustration of the LINEX loss function. It is smooth and unbounded.
\item \textbf{RoBoSS loss function:} To enhance the robustness against outliers of smooth and unbounded loss functions, \citet{10685140} proposed the RoBoSS loss function. The mathematical formulation of RoBoSS loss is given as follows:
\begin{align}     
\mathfrak{L}_{rbss}(u)=
\begin{cases}
\lambda\left(1-(au+1)e^{-au}\right), & u > 0, \\
0, & u \leq 0,
\end{cases}
\end{align}
where $a>0$, $\lambda >0$ are shape and bounding parameters, respectively. It is smooth and bounded.
\end{itemize}
In addition to these, recent advancements in smooth loss functions include the smooth truncated H$\epsilon$ loss \cite{shi2023robust}, HawkEye loss \cite{akhtar2024hawkeye}, and so forth.

Through a comparative analysis of prevalent loss functions in the existing literature, we find that bounded loss functions demonstrate robustness by imposing a fixed penalty on all misclassified instances beyond a specified margin. Further, the strategic imposition of penalties upon both correctly classified and misclassified samples facilitates the calibration of a delicate equilibrium between accuracy and resistance against noise. In addition to robustness to outliers and insensitivity towards noise, smoothness is also a crucial aspect of loss function. Generally, SVM utilizes non-smooth loss functions, which suffer from high computational costs as they have to solve a QPP. It is evident that if the loss function is not smooth, SVM is also not smooth \cite{feng2016robust}. Most of the robust and insensitive to noise models are not convex, so the optimization problem of these models cannot be solved using the Wolfe dual method. The convenience of the smooth loss function is that there are fast optimization techniques that are specific to smooth models.
\par
Taking inspiration from prior research endeavors, this article proposes a novel asymmetric loss function, referred to as the wave loss function (see Figure \ref{fig:Proposed Loss}). It is precisely engineered to exhibit robustness against outliers, insensitivity to noise, and a propensity for smoothness. Subsequently, by integrating the proposed wave loss into the least squares framework of SVM and TSVM, we present two novel models, namely Wave-SVM and Wave-TSVM. The optimization problem associated with the Wave-SVM is effectively addressed through the utilization of the adaptive moment estimation (Adam) algorithm. Further, to solve the optimization
problems of the Wave-TSVM an efficient
iterative algorithm is utilized. The major contributions of this paper can be outlined as follows: 
\begin{itemize}
\item We propose a novel asymmetric loss function named wave loss, designed to exhibit robustness against outliers, insensitivity to noise, and smooth characteristics. Further, we delve into the theoretical aspect of the wave loss function and validate its capability to maintain a vital classification-calibrated property.
\item We amalgamate the proposed wave loss function into the least squares setting of SVM and TSVM and introduce two novel robust and smooth models, termed Wave-SVM and Wave-TSVM, respectively.
\item We address the optimization problem inherent to Wave-SVM by employing the Adam algorithm, known for its lower memory requirements and efficacy in handling large-scale problems. To our knowledge, this is the first time Adam has been used to solve an SVM problem. Further, we utilized an efficient iterative algorithm to solve the optimization problems of Wave-TSVM. 
 
\item We perform comprehensive numerical experiments using benchmark UCI and KEEL datasets (with and without feature noise) from various domains. The outcomes vividly showcase the outstanding performance of the proposed Wave-SVM and Wave-TSVM when contrasted with baseline models.
\item To demonstrate the superiority of the proposed Wave-SVM and Wave-TSVM models in the biomedical realm, we conducted experiments using the ADNI dataset. These empirical investigations furnish substantial evidence of the proposed models applicability in real-world medical scenarios.
\end{itemize}
The rest of this paper is structured as follows: Section 2 presents the proposed wave loss function and elucidates its distinctive characteristics. Further, Section 2 provides the formulations of Wave-SVM and Wave-TSVM, along with an analysis of their computational complexity.  Section 3 showcases the numerical findings. Lastly, Section 4 offers conclusions and outlines future research directions.

\section{Proposed work}
In this section, we introduce a groundbreaking asymmetric loss function named wave loss and elucidate its distinctive characteristics. Subsequently, we amalgamate the proposed wave loss into the frameworks of SVM and TSVM within the least squares setting and propose two novel models, namely Wave-SVM and Wave-TSVM. We also analyze the computational complexity associated with the proposed algorithms.

\subsection{Wave loss function}
\begin{figure*}[h]
\centering
    \subcaptionbox{     \label{fig:first }} { %
      \includegraphics[width=0.45\textwidth,keepaspectratio]{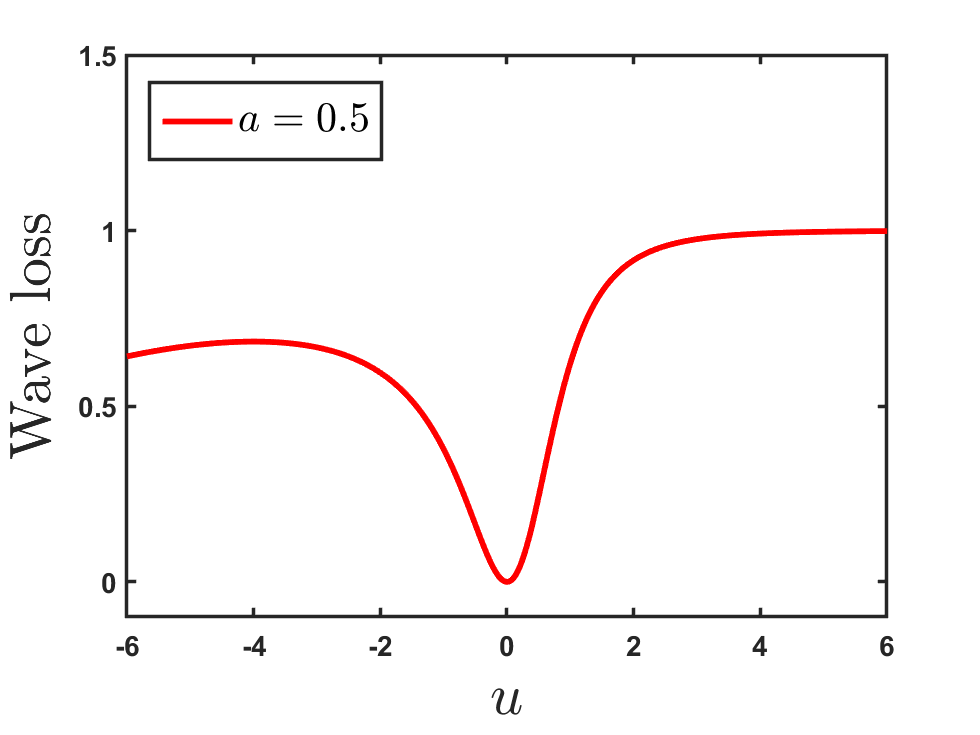}}
      \hfill
      \subcaptionbox{   \label{fig:second }} { %
      \includegraphics[width=0.45\textwidth,keepaspectratio]{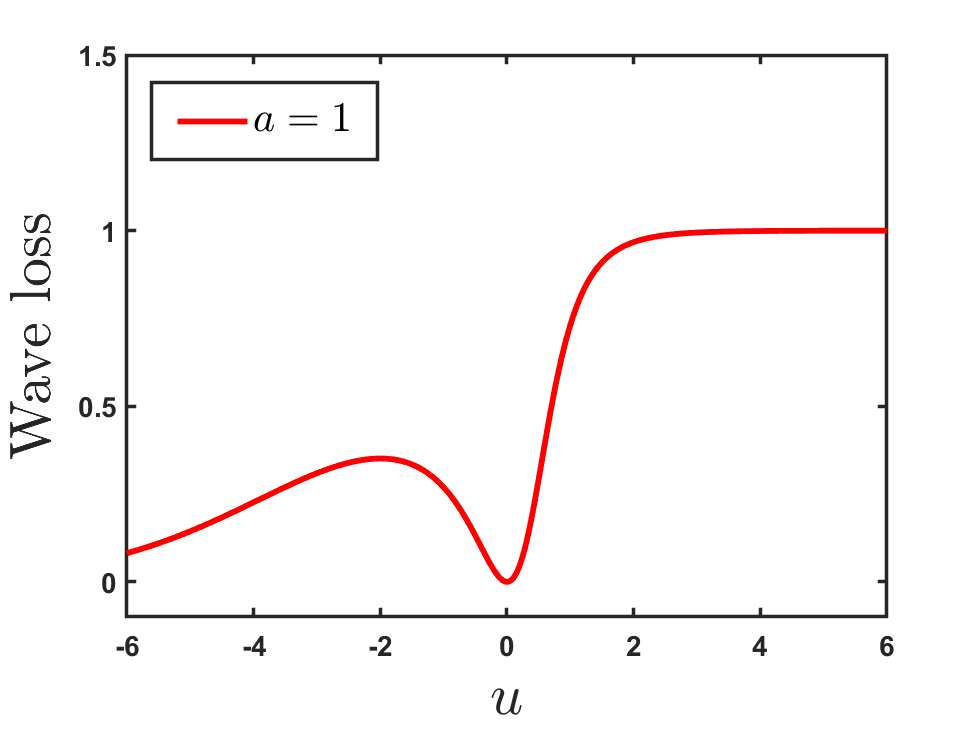}}
\\
      \subcaptionbox{  \label{fig:third}} { %
      \includegraphics[width=0.45\textwidth,keepaspectratio]{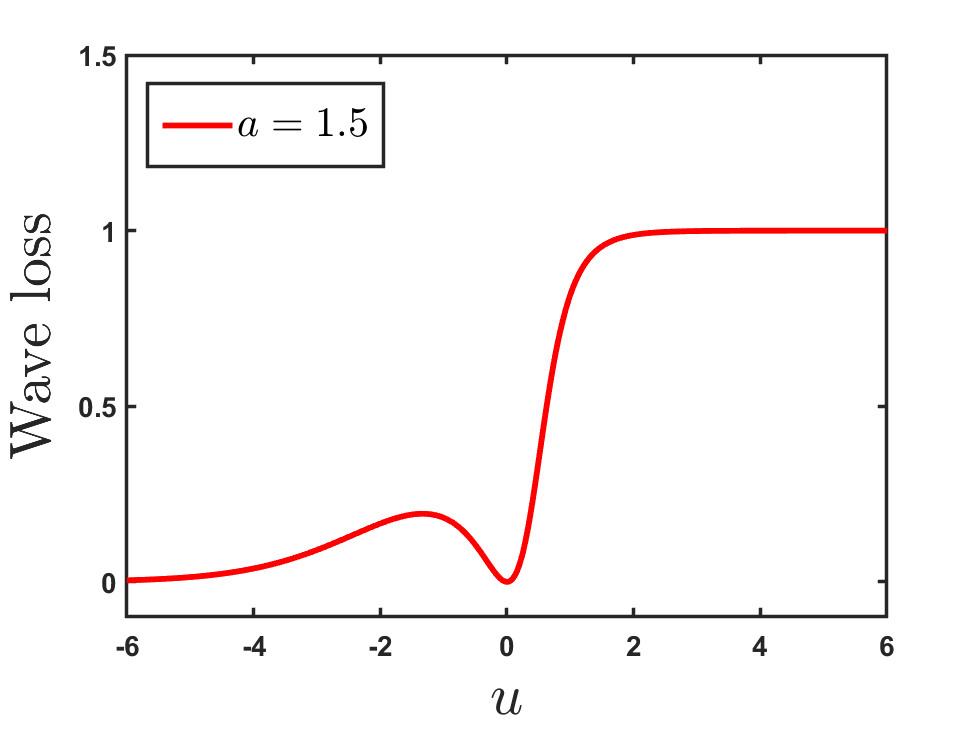}}
      \hfill
      \subcaptionbox{  \label{fig:four}} { %
      \includegraphics[width=0.45\textwidth,keepaspectratio]{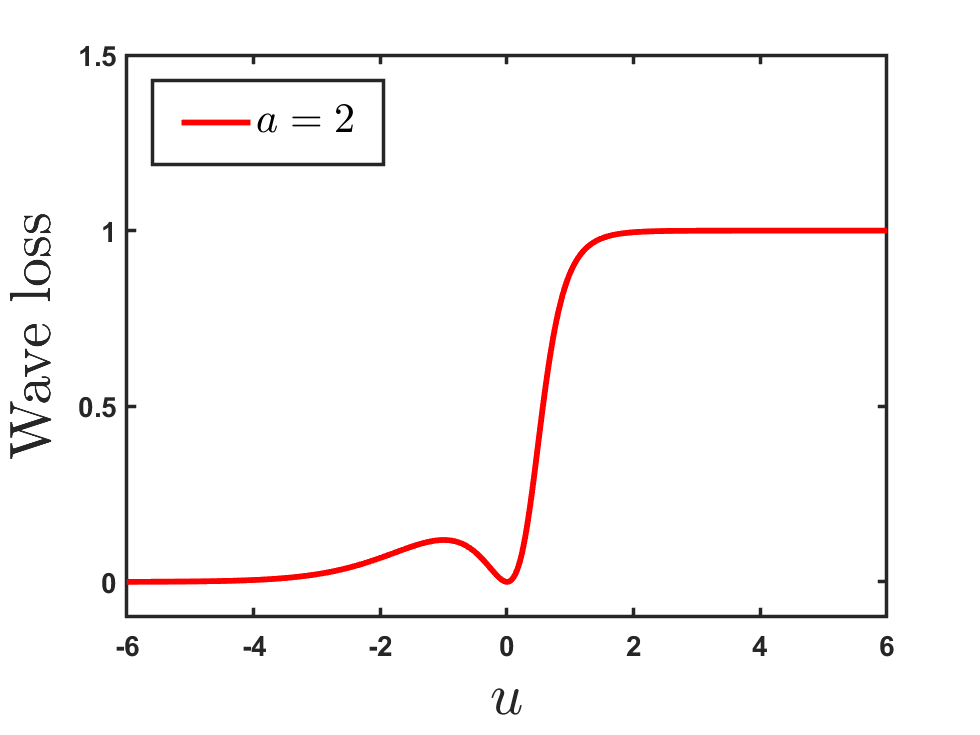}}
      \caption{Illustration of wave loss function for fixed $\lambda=1$ and  different values of $a$. Subfigures (a), (b), (c), and (d) demonstrate that the value of $a$ controls the strength of the penalty for correctly classified and misclassified samples.}
    \label{fig:Proposed Loss}
 \end{figure*}

We present a substantial advancement within the domain of supervised learning: a novel asymmetric loss function that embodies robustness against outliers, insensitivity to noise, and smoothness characteristics. This novel loss function, referred to as the wave loss function, is visually depicted in Figure \ref{fig:Proposed Loss}. The mathematical expression of the proposed wave loss function is articulated as follows:
\begin{align} \label{proposedloss}   
\mathfrak{L}_{wave}(u)= \frac{1}{\lambda} \biggl(1-\frac{1}{1+\lambda u^2e^{au}}\biggr), ~\forall~ u \in \mathbb{R},
\end{align}
where $a \in \mathbb{R}$ is shape parameter and $\lambda \in \mathbb{R}^+$ is bounding parameter. The wave loss function (\ref{proposedloss}), as delineated in this work, showcases the following inherent properties:
\begin{itemize}
\item It is robust to outliers and insensitive to noise. As it bound the loss to $1/\lambda$, which gives robustness to outliers, and it also gives loss to samples with $u \leq 0$, which produces insensitivity towards noise.
\item It is non-convex, smooth, and bounded.
\item It is infinitely differentiable and hence continuous for all $u \in \mathbb{R}$.
\item For given value of $\lambda$, when $a  \rightarrow +\infty $, the wave loss (\ref{proposedloss}) converges point-wise to the ``$0-1/\lambda$" loss which is defined as
$$  
\mathfrak{L}_{0-1}(u)=
\begin{cases}
\frac{1}{\lambda}, & u > 0, \\
0, & u \leq 0.
\end{cases}
$$
Furthermore, when $\lambda=1$ it converges to ``$0-1$" loss.
\item It has two advantageous parameters: shape parameter $(a)$ determines the loss function's shape, and  bounding parameter $(\lambda)$ sets thresholds for loss values.
\end{itemize}

It is crucial to note that the existing bounded loss functions set a strict limit on the maximum allowable loss for data points with substantial deviations. This effectively prevents noise and outliers from exerting excessive influence, thus enhancing the model's robustness. However, the majority of the existing loss functions use a hard truncation strategy to bound the loss, which makes them non-smooth. In contrast, the proposed wave loss function is smooth and bounded simultaneously (refer to Figure \ref{fig:Proposed Loss}). By adjusting the bounding parameter $\lambda$, the wave loss function smoothly bounds the loss to a predefined value. The primary benefit of the smoothness characteristic lies in its facilitation of the application of gradient-based optimization algorithms. This property ensures the existence of well-defined gradients, enabling the utilization of rapid and reliable optimization techniques.

 \subsection{Theoretical analysis of the wave loss function}
In this subsection, we analyze the theoretical characteristics inherent to the proposed wave loss function and provide evidence to demonstrate that the wave loss function exhibits a crucial classification-calibrated property. \citet{bartlett2006convexity} introduced the concept of classification-calibration to assess the statistical effectiveness of loss functions in the context of classification learning. It ensures that the model's predicted probabilities reflect the true likelihood of an event. For more details, readers can refer to \cite{bartlett2006convexity, sypherd2022tunable}. This attribute holds crucial importance in enhancing our understanding of the performance dynamics of the wave loss function within classification tasks.
\par
\begin{figure*}
\centering
\subcaptionbox{\label{fig:Calibrated1 }} { %
      \includegraphics[width=0.48\textwidth,keepaspectratio]{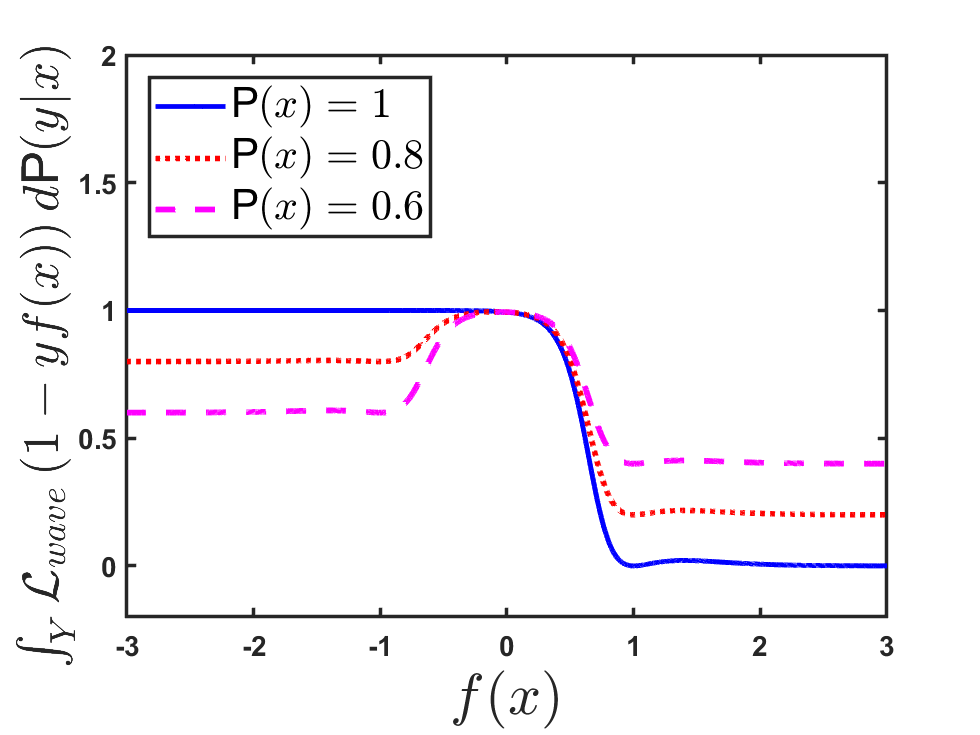}}
      \hfill
      \subcaptionbox{\label{fig:calibrated2 }} { %
      \includegraphics[width=0.48\textwidth,keepaspectratio]{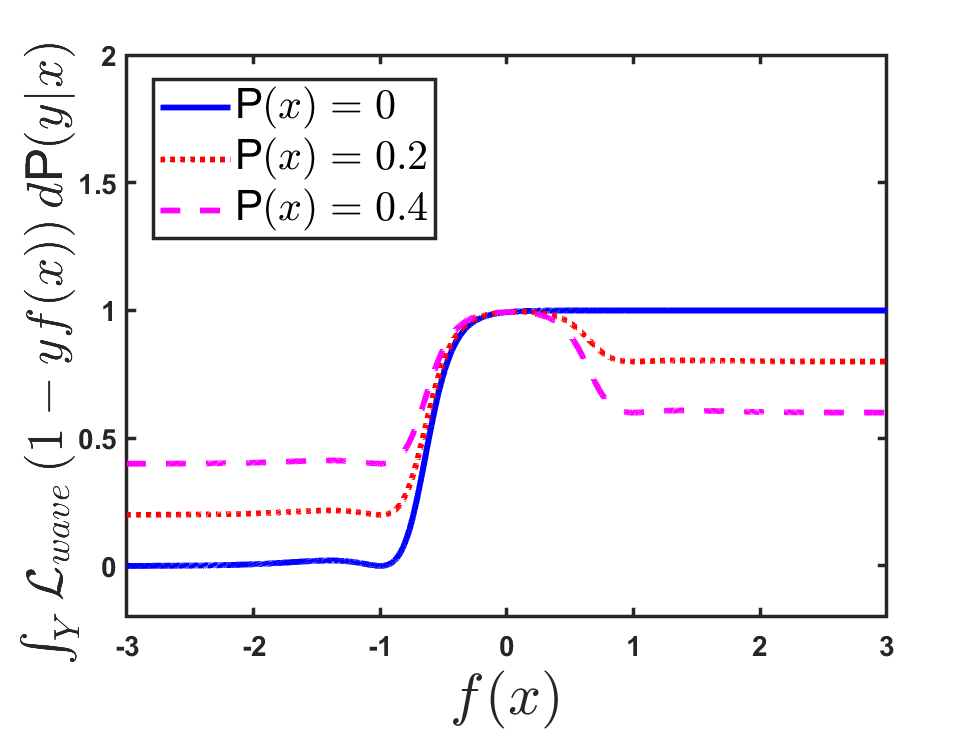}}
      \caption{Illustrate the plot of $\int_{Y}\mathfrak{L}_{wave}\left(1-yf(x)\right) d\mathsf{P} (y\vert x)$ over $f(x)$ with varying values of $\mathsf{P}(x)$. (a) Depict the case where $\mathsf{P}(x)$ is greater than $1/2$,  and (b) illustrate the scenario where $\mathsf{P}(x)$ less than $1/2$.}
    \label{fig:Classification calibrated figures}
 \end{figure*}

Consider that the training data $\mathcal{D}=\left\{x_i,y_i\right\}_{i=1}^l$ is independently sampled from a probability distribution $\mathsf{P}$. The distribution $\mathsf{P}$ is defined over a combined space of inputs and corresponding labels, where the input space $X$ is a subset of $\mathbb{R}^n$ and the label space $Y$ consists of two possible labels: $-1$ and $1$. In this context, the main objective is to develop a binary classifier $\mathsf{C}$, which takes inputs from the space $X$ and assigns them to one of the two labels in $Y$. The central goal of this classification problem is to design a classifier in a way that minimizes the associated error. The risk associated with a specific classifier $\mathsf{C}$ is quantified by the following mathematical expression:
$$
\mathsf{R}(\mathsf{C})=\int_{X}\mathsf{P} (y\ne \mathsf{C}(x)\vert x)d \mathsf{P}_{X}.
$$
Here $\mathsf{P} (y\vert x)$ signifies the probability distribution of the label $y$ given an input $x$, and $d \mathsf{P}_{X}$ represents the marginal distribution of the input $x$ according to the distribution $\mathsf{P}$. Moreover, the conditional distribution $\mathsf{P} (y\vert x)$ is a binary distribution, implying it is determined by the probabilities ${\tt Prob}(y=1 \vert x)$ and ${\tt Prob}(y=-1 \vert x)$. To ease in writing, we further utilize ${\tt P}(x)$ and $1 - {\tt P} (x)$ to represent ${\tt Prob}(y=1 \vert x)$  and ${\tt Prob}(y=-1 \vert x)$, respectively.
In simpler terms, the objective is to construct a classifier that accurately assigns labels to inputs while minimizing the overall error.\\
Now, the Bayes classifier, for ${\tt P}(x) \neq 1/2$, is defined as follows:
\begin{align} \label{Bayes classifier}     
f_{\mathsf{C}}(x)=
\begin{cases}
-1, & {\tt P}(x) < 1/2,\\
~~1, & {\tt P}(x) > 1/2.
\end{cases}
\end{align}
It can be verified that the Bayes classifier effectively minimizes the classification error. The following expression mathematically expresses the previous statement:
$$
f_{\mathsf{C}}=\arg\min_{{\sf C}{:}X\rightarrow Y}{\sf R}({\sf C}).
$$
Now, for any given loss function $\mathfrak{L}$, the expected error associated with a classifier $f{:} X \rightarrow \mathbb{R}$ is formulated as follows:
\begin{align}
    {\sf R}_{\mathfrak{L},{\sf P}}(f)=\int_{X\times Y}\mathfrak{L}(1-yf(x)) d{\sf P}.
\end{align}
The function $f_{\mathfrak{L},\mathsf{P}}$, which minimizes the expected error over all measurable functions, is formulated as:
\begin{align}{}
    f_{\mathfrak{L},\mathsf{P}}(x)=\arg\min_{f(x)\in\mathbb{R}}\int_{Y}\mathfrak{L}\left(1-yf(x)\right) d\mathsf{P} (y\vert x),~\forall x\in X.
\end{align}
Subsequently, for the proposed wave loss $\mathfrak{L}_{wave}(\cdot)$, we can derive Theorem \ref{classification calibration theorem}, providing evidence of the classification-calibrated nature \cite{bartlett2006convexity} of the wave loss function. This characteristic, a valuable aspect of a loss function, ensures that the minimizer of expected error aligns with the sign of the Bayes classifier.
\begin{theorem} \label{classification calibration theorem}
 The proposed loss function $\mathfrak{L}_{wave}(u)$ is classification-calibrated, i.e., $f_{\mathfrak{L}_{wave},\mathsf{P}}$ has the same sign as the Bayes classifier.   
\end{theorem}
\begin{proof}
After simple computation, we arrive at the following outcome:
\begin{align*}
&\int_{Y}\mathfrak{L}_{wave}\left(1-yf(x)\right) d\mathsf{P} (y\vert x)\\
&=\mathfrak{L}_{wave}(1-f(x)) \mathsf{P}(x)+\mathfrak{L}_{wave}(1+f(x)) (1-\mathsf{P}(x))\\
&= \frac{1}{\lambda} \biggl(1-\frac{1}{1+\lambda (1-f(x))^2e^{a(1-f(x))}}\biggr) \mathsf{P}(x) + \frac{1}{\lambda} \biggl(1-\frac{1}{1+\lambda (1+f(x))^2e^{a(1+f(x))}}\biggr) (1-\mathsf{P}(x)).
\end{align*}
In Figures \ref{fig:Calibrated1 } and \ref{fig:calibrated2 }, the graphical representations of $\int_{Y}\mathfrak{L}_{wave}\left(1-yf(x)\right) d\mathsf{P} (y\vert x)$ are depicted as functions of $f(x)$, considering the cases when $\mathsf{P}(x)$ is greater than $1/2$ and when it is less than $1/2$, respectively. As discernible from Figure \ref{fig:Classification calibrated figures}, when $\mathsf{P}(x)$ exceeds $1/2$, the minimum value of $\int_{Y}\mathfrak{L}_{wave}\left(1-yf(x)\right) d\mathsf{P} (y\vert x)$ is achieved for a positive value of $f(x)$. Conversely, when $\mathsf{P}(x)$ is less than $1/2$, the minimum value corresponds to a negative value of $f(x)$.

Hence, it becomes evident from the patterns observed in Figures \ref{fig:Calibrated1 } and \ref{fig:calibrated2 } that the proposed loss function $\mathfrak{L}_{wave}(u)$ holds the characteristic of being classification-calibrated.
\end{proof}

\subsection{Formulation of Wave-SVM}   
Through the amalgamation of the proposed wave loss function into the least squares setting of SVM, we construct a novel support vector classifier for large-scale problems that manifest robustness against outliers, insensitivity to noise, and smoothness characteristics. This advanced classifier is denoted as the Wave-SVM. For simplification, throughout the Wave-SVM formulation, we use the terminology $w$ for $\left[w^\intercal,b\right]$ and $x_i$ for $\left[x_i,1\right]^\intercal$. The formulation of non-linear Wave-SVM is given as: 
\begin{align} \label{proposedSVMprimary}
\underset{ w, \xi}{min}  \hspace{0.5cm}~& \frac{1}{2}\|w\|^2+C\sum_{i=1}^l \frac{1}{\lambda} \bigg(1-\frac{1}{1+\lambda \xi_i^2e^{a\xi_i}}\bigg), \nonumber \\
\text { subject to }\hspace{0.2cm}  & y_i\left(w^\intercal \phi(x_i)\right) = 1-\xi_i, ~\forall~ i=1,2, \ldots,l,
\end{align}
where $C>0$ is the regularization parameter, which determines the trade-off between margin and the penalty, $\lambda$ and $a$ are the parameters of the wave loss function, and $\phi(\cdot)$ is feature mapping associated with the kernel function.  The characteristics of various SVM models, along with the Wave-SVM, are presented in Table \ref{tab:SVM with different loss function}, which showcases that the proposed Wave-SVM possesses most of the desirable attributes required for a better classifier.
\begin{table}[]
\centering
\caption{Comparative Analysis of SVM Models Utilizing Different Loss Functions.}

\label{tab:SVM with different loss function}
\resizebox{\textwidth}{!}{%
\begin{tabular}{|l|l|l|l|l|l|l|l|}
\hline
\multicolumn{1}{|c|}{\textbf{Models $\downarrow$ \textbackslash Characteristics $\rightarrow$}} &
  \multicolumn{1}{c|}{\textbf{Robust to outliers}} &
  \multicolumn{1}{c|}{\textbf{Insensitive to noise}} &
  \multicolumn{1}{c|}{\textbf{Sparse}} &
  \multicolumn{1}{c|}{\textbf{Convex}} &
  \multicolumn{1}{c|}{\textbf{Bounded}} &
  \multicolumn{1}{c|}{\textbf{Smooth}} &
  \multicolumn{1}{c|}{\textbf{Fast}} \\ \hline
\textbf{C-SVM \cite{cortes1995support}}     &\hspace{1.8cm}{\color{red}\ding{55}}  &~~~~~~~~~~~~{\color{red}\ding{55}}  &~~~{\color{blue}\ding{51}}  &~~~~{\color{blue}\ding{51}}  &~~~~{\color{red}\ding{55}}  &~~~~{\color{red}\ding{55}}  &~~{\color{red}\ding{55}}  \\ \hline
\textbf{Pin-SVM \cite{huang2013support}}   &\hspace{1.8cm}{\color{red}\ding{55}}  &~~~~~~~~~~~~{\color{blue}\ding{51}}  &~~~{\color{red}\ding{55}}  &~~~~{\color{blue}\ding{51}}  &~~~~{\color{red}\ding{55}}  &~~~~{\color{red}\ding{55}}  &~~{\color{red}\ding{55}}  \\ \hline
\textbf{LINEX-SVM \cite{ma2019linex}} &\hspace{1.8cm}{\color{red}\ding{55}}  &~~~~~~~~~~~~{\color{blue}\ding{51}}  &~~~{\color{red}\ding{55}}  &~~~~{\color{blue}\ding{51}}  &~~~~{\color{red}\ding{55}}  &~~~~{\color{blue}\ding{51}}  &~~{\color{blue}\ding{51}}  \\ \hline
\textbf{QTLS \cite{zhao2022asymmetric}}      &\hspace{1.8cm}{\color{red}\ding{55}}  &~~~~~~~~~~~~{\color{blue}\ding{51}}  &~~~{\color{red}\ding{55}}  &~~~~{\color{red}\ding{55}}  &~~~~ {\color{red}\ding{55}} &~~~~{\color{blue}\ding{51}}  &~~{\color{blue}\ding{51}}  \\ \hline
\textbf{FP-SVM \cite{kumari2024diagnosis}}      &\hspace{1.8cm}{\color{red}\ding{55}}  &~~~~~~~~~~~~{\color{blue}\ding{51}}  &~~~{\color{red}\ding{55}}  &~~~~{\color{blue}\ding{51}}  &~~~~ {\color{red}\ding{55}} &~~~~{\color{red}\ding{55}}  &~~{\color{red}\ding{55}}  \\ \hline
\textbf{Wave-SVM (Proposed)}  &\hspace{1.8cm}{\color{blue}\ding{51}}  &~~~~~~~~~~~~{\color{blue}\ding{51}}  &~~~{\color{red}\ding{55}}  &~~~~{\color{red}\ding{55}}  &~~~~{\color{blue}\ding{51}}  &~~~~{\color{blue}\ding{51}}  &~~{\color{blue}\ding{51}} \\ \hline
\end{tabular}%
}
\end{table}  

The dual problem of the Wave-SVM is challenging to optimize due to the non-convexity of the wave loss function. However, the inherent smoothness of the Wave-SVM permits us to utilize a gradient-based algorithm to optimize the model. Gradient-based optimization with Wave-SVM offers several advantages. It facilitates faster convergence during training, as gradient-based methods generally converge at a faster rate than quadratic programming solvers \cite{bottou2018optimization}. In this paper, we adopt the Adam \cite{kingma2014adam} optimization technique for solving the Wave-SVM. It is an improved version of stochastic gradient descent with adaptive learning rates. This is the first time the Adam algorithm has been used to solve an SVM model. The Adam algorithm converges more quickly and stays stable during learning due to the adjustable learning rates. It combines the benefits of two other popular optimization algorithms: AdaGrad \cite{duchi2011adaptive} and RMSProp \cite{tieleman2012lecture}. The core idea behind Adam involves the computation of a progressively diminishing average for previous gradients and the squared gradients of the weights. Subsequently, based on these estimations, the algorithm establishes an adaptive learning rate for each weight parameter. It is important to note that the Adam algorithm only requires the gradient of the unconstrained optimization problem, and thus no modification was made to the Adam algorithm for its application to Wave-SVM.

\subsubsection{Adam for linear Wave-SVM}
For linear Wave-SVM, put $\phi(x)=x$ in equation (\ref{proposedSVMprimary}), the linear Wave-SVM is defined as:
\begin{align} \label{linearproposedSVMprimary}
\underset{ w}{min} ~f(w)= ~& \frac{1}{2}\|w\|^2+C\sum_{i=1}^l \frac{1}{\lambda} \bigg(1-\frac{1}{1+\lambda \xi_i^2e^{a\xi_i}}\bigg), \nonumber \\
\text { subject to }\hspace{0.2cm}  & y_i\left(w^\intercal x_i\right) = 1-\xi_i, ~\forall~ i=1,2, \ldots,l.
\end{align}
At each iteration $t$, $k$ samples are chosen, and at that time the value of gradient,  exponentially decaying averages of past gradients (first moment), and past squared gradients (second moment) are calculated using (\ref{gradient}), (\ref{first moment}), and (\ref{second moment}), respectively as follows:
\begin{align} 
g_t=\nabla f(w)=&w-\sum_{i=1}^k \frac{ C y_i x_i \xi_i   \left( 2+\xi_i\right) \exp\left(a \xi_i\right)}{\left[1+\lambda  \xi_i^2  \exp \left(a \xi_i\right)\right]^2}, \label{gradient}\\
m_t=& \beta_1 m_{t-1} + \left(1-\beta_1\right) \nabla f(w_t), \label{first moment}\\
v_t=& \beta_2 v_{t-1} + \left(1-\beta_2\right) \nabla f(w_t)^2, \label{second moment}
\end{align}
where $\beta_1$ and $\beta_2$ are exponential decay rates for the first and second moment estimates, usually set to $0.9$ and $0.999$, respectively. Then, we compute the bias-corrected first and second moment estimates using (\ref{corrected first moment}) and (\ref{corrected second moment}), respectively.
 \begin{align} 
\hat{m_t}=& \frac{m_t}{\left(1-\beta_1^t\right)}, \label{corrected first moment}\\
\hat{v_t}=& \frac{v_t}{\left(1-\beta_2^t\right)}. \label{corrected second moment}
\end{align}
In the end, the parameter $w$ is updated as:
\begin{align} \label{update rule}
w_t= w_{t-1} - \alpha \frac{\hat{m_t}}{\sqrt{\hat{v_t}}+\epsilon},
\end{align}
where $\alpha$ is the learning rate, and $\epsilon$ is a small constant used to avoid division by zero. Once the optimal $w$ is achieved, the prediction regarding the label of the unknown sample $x$ can be made using the decision function as:
\begin{align} \label{decision function}
\hat{y}= {\operatorname{\text{sign}}}(f(x))= {\operatorname{\text{sign}}}\left(w^\intercal x\right).
\end{align}
\subsubsection{Adam for non-linear Wave-SVM}
The determination of the dual problem in the proposed Wave-SVM poses challenges that limit the applicability of kernel methods. In this case, to enhance the capacity of Wave-SVM for non-linear adaptation, we use the representer theorem \cite{dinuzzo2012representer}, which allows us to express $w$ in equation (\ref{proposedSVMprimary}) as follows:
\begin{align} \label{representer theorem}
    w= \sum_{j=1}^l \gamma_j \phi(x_j),
\end{align}
where $\gamma = \left(\gamma_1, \ldots, \gamma_l \right)^\intercal$ is the coeffecient vector.\\
Substituting (\ref{representer theorem}) into (\ref{proposedSVMprimary}), we obtain:
\begin{align}
{\underset{\gamma}{min}}~ f(\gamma)=\sum_{i=1}^l \sum_{j=1}^l \frac{1}{2} \gamma_i \gamma_j \mathcal{K}\left(x_i, x_j\right)+C\sum_{i=1}^l \frac{1}{\lambda} \bigg(1-\frac{1}{1+\lambda \xi_i^2e^{a\xi_i}}\bigg),
\end{align}
where $\xi_i=1- y_i\left(\sum_{j=1}^l \gamma_{j} \mathcal{K}\left(x_j, x_i\right)\right) $, and $\mathcal{K}\left(x_{j},x_{i}\right)=\left(\phi\left(x_{j}\right) \cdot \phi\left(x_{i}\right)\right)$ is the Kernel function.
\begin{algorithm}
  \caption{Non-linear Wave-SVM}
  \label{algorithm1}
   \begin{algorithmic}
  \State \textbf{Input:}
     \State The dataset:  $\left\{x_i,y_i\right\}_{i=1}^l$, $y_i \in\{-1,1\}$;
     \State The parameters: Regularization parameter $C$, wave loss parameters $\lambda$ and $a$, mini-batch size $k$, decay rates $\beta_1$ and $\beta_2$, learning rate $\alpha$, constant $\epsilon$, error tolerance $\eta$, maximum iteration number $T$;    
      \State Initialize: $\gamma_0$ and $t$;
     \State \textbf{Output:}
       \State The classifiers parameters: $\gamma$;
       \State $1:$ Select $k$ samples $\left\{x_i,y_i\right\}_{i=1}^k$ uniformly at random.
\State $2:$  Computing $\xi_i$ :
\begin{align}
\xi_i= 1 -y_i\left(\sum_{j=1}^k \gamma_{j} \mathcal{K}\left(x_j, x_i\right)\right), \quad i=1, \ldots, k;
\end{align}
\State $3:$ Compute $\nabla f(\gamma_t)$: (\ref{ non-linear gradient});
\State $4:$ Compute $m_t'$: (\ref{non-linear first moment});
\State $5:$ Compute $v_t'$: (\ref{non-linear second moment});
\State $6:$ Compute $\hat{m_t}'$: (\ref{non-linear corrected first moment});
\State $7:$ Compute $\hat{v_t}'$: (\ref{non-linear corrected second moment});
\State $8:$ Update SVM solution $\gamma_t$: (\ref{non-linear update rule});
\State $9:$ Update iteration number: $t=t+1$.
\State \textbf{Until:}
\State $|\gamma_t - \gamma_{t-1}| < \eta$~~ or~ $t=T$
\State \textbf{Return:} $\gamma_t$.
 \end{algorithmic}
\end{algorithm}

In the same way, as in linear Wave-SVM, we use the following formulas to obtain the gradient,  exponentially decaying averages of past gradients and past squared gradients, and then update the parameter $\gamma$. The Adam algorithm structure for non-linear Wave-SVM is clearly described in Algorithm \ref{algorithm1}.
\begin{align} \label{ non-linear gradient}
\nabla f(\gamma_t)=\mathcal{K} \gamma -\sum_{i=1}^k \frac{ C y_i \mathcal{K}_i \xi_i   \left( 2+\xi_i\right) \exp\left(a \xi_i\right)}{\left[1+\lambda  \xi_i^2  \exp \left(a \xi_i\right)\right]^2},
\end{align}
where $\mathcal{K}$ is the Gaussian Kernel matrix of the randomly selected training dataset and $\mathcal{K}_i$ is the $i^{th}$ row of the matrix $\mathcal{K}$.
\begin{align} 
m_t'=& \beta_1 m_{t-1}' + \left(1-\beta_1\right) \nabla f(\gamma_t), \label{non-linear first moment}\\
v_t'=& \beta_2 v_{t-1}' + \left(1-\beta_2\right) \nabla f(\gamma_t)^2, \label{non-linear second moment}\\
\hat{m_t}'=& \frac{m_t'}{\left(1-\beta_1^t\right)}, \label{non-linear corrected first moment}\\
\hat{v_t}'=& \frac{v_t'}{\left(1-\beta_2^t\right)}.  \label{non-linear corrected second moment}
\end{align}
Eventually, the parameter $\gamma$ is updated as follows:
\begin{align} \label{non-linear update rule}
\gamma_t= \gamma_{t-1} - \alpha \frac{\hat{m_t}'}{\sqrt{\hat{v_t}'}+\epsilon}.
\end{align}
When the optimal $\gamma$ is obtained, the following decision function can be utilized to predict the label of a new sample $x$.
\begin{align} \label{non-linear decision function}
\hat{y}= {\operatorname{\text{sign}}}(f(x))= {\operatorname{\text{sign}}}\left(\sum_{j=1}^k \gamma_{j} \mathcal{K}\left(x_j, x\right)\right).
\end{align}

\subsection{Formulation of Wave-TSVM}
In this subsection, we amalgamate the proposed asymmetric wave loss function into the least squares setting of TSVM and propose a robust and smooth classifier referred to as Wave-TSVM. Let ${X}_{+}=\left({x}_1, \ldots, {x}_{l_{+}}\right)^{\top} \in \mathbb{R}^{l_{+} \times n}$ and ${X}_{-}=\left({x}_1, \ldots, {x}_{l_{-}}\right)^{\top} \in$ $\mathbb{R}^{l_{-} \times n}$ represent matrices containing positive and negative instances, where $l_{+}$ and $l_{-}$ denote the count of positive and negative instances, respectively, and $l=l_{+}+l_{-}$.
Further, $e_1$ and $e_2$ are identity vectors of appropriate size, and $I$ is the identity matrix of appropriate size.

\subsubsection{Linear Wave-TSVM}
Given a training dataset $\mathcal{D}$, the goal of Wave-TSVM is to obtain the positive and negative hyperplanes as follows:
\begin{align}\label{linear twin hyperplanes}
{w}_{+}^{\top} {x}+b_{+}=0 ~~\text { and } \quad {w}_{-}^{\top} {x}+b_{-}=0,
\end{align}
where ${w}_{+}$, ${w}_{-}$ $\in$ $\mathbb{R}^{n}$
and $b_{+}$, $b_{-}$ $\in$ $\mathbb{R}$
are the weight vectors and the bias terms,
respectively. To obtain the hyperplanes (\ref{linear twin hyperplanes}), the primal of linear Wave-TSVM are formed as follows:\\
(Linear Wave-TSVM-1)
\begin{align} \label{linear wave TSVM-1}
& \min _{{w}_{+}, b_{+}, \xi^{-}} \frac{1}{2} \sum_{i=1}^{l_{+}}\left({w}_{+}^{\top} {x}_i+b_{+}\right)^2+\frac{1}{2} C_1\left(\left\|{w}_{+}\right\|_2^2+b_{+}^2\right)+C_2 \sum_{j=1}^{l_{-}} \xi_j^{-} \nonumber \\
& \text {s.t. } \xi_j^{-}= \frac{1}{\lambda} \biggl( 1- \frac{1}{1+\lambda \left(1+{w}_{+}^{\top} {x}_j+b_{+}\right)^2 \exp \{a\left(1+{w}_{+}^{\top} {x}_j+b_{+} \right)\}}\biggr), \quad j=1, \ldots, l_{-},
\end{align}
(Linear Wave-TSVM-2)
\begin{align} \label{linear wave TSVM-2}
& \min _{{w}_{-}, b_{-}, \xi^{+}} \frac{1}{2} \sum_{j=1}^{l_{-}}\left({w}_{-}^{\top} {x}_j+b_{-}\right)^2+\frac{1}{2} C_3\left(\left\|{w}_{-}\right\|_2^2+b_{-}^2\right)+C_4 \sum_{i=1}^{l_{+}} \xi_i^{+}, \nonumber \\ 
& \text {s.t. } \xi_i^{+}= \frac{1}{\lambda} \biggl( 1- \frac{1}{1+\lambda \left(1-{w}_{-}^{\top} {x}_i-b_{-}\right)^2 \exp \{a\left(1-{w}_{-}^{\top} {x}_i-b_{-} \right)\}}\biggr), \quad i=1, \ldots, l_{+}
\end{align}
where ${\xi}_{+}=\left(\xi_1^{+}, \ldots, \xi_{l_{+}}^{+}\right)^{\top} \in \mathbb{R}^{l_{+}}, {\xi}_{-}=\left(\xi_1^{-}, \ldots, \xi_{l_{-}}^{-}\right)^{\top} \in$ $\mathbb{R}^{l_{-}}$. For the sake of brevity, we solely discuss optimization problem (\ref{linear wave TSVM-1}), noting that optimization problem (\ref{linear wave TSVM-2}) follows a similar structure. The objective function provided in equation (\ref{linear wave TSVM-1}) comprises three distinct terms. More precisely, the initial term aims to minimize the distance between the positive hyperplane and the positive instances. The second term, a regularization term, is incorporated to enhance the generalization performance of the model. The third term represents the cumulative penalty of all negative samples, leveraging the proposed wave loss function.

Given the non-convex nature of optimization problems (\ref{linear wave TSVM-1}) and (\ref{linear wave TSVM-2}), and utilizing their inherent smoothness, we devise an iterative algorithm to solve them. Initially, we convert (\ref{linear wave TSVM-1}) and (\ref{linear wave TSVM-2}) into vector-matrix form in the following manner:
\begin{align} \label{optimizationproblem1}
\min _{{w}_1} P_1\left({w}_1\right)=\frac{1}{2}\left\|{G}^{\top} {w}_1\right\|_2^2+\frac{1}{2} C_1\left\|{w}_1\right\|_2^2+C_2 \mathfrak{L}_1\left({w}_1\right),
\end{align}
and
\begin{align} \label{optimizationproblem2}
\min _{{w}_2} P_2\left({w}_2\right)=\frac{1}{2}\left\|{H}^{\top} {w}_2\right\|_2^2+\frac{1}{2} C_3\left\|{w}_2\right\|_2^2+C_4 \mathfrak{L}_2\left({w}_2\right),
\end{align}
where $\mathfrak{L}_1\left({w}_1\right)=\sum_{j=1}^{l_{-}}
\frac{1}{\lambda} \biggl( 1- \frac{1}{1+\lambda \left(1+{H}_j^{\top} {w}_1\right)^2 \exp \{a\left(1+{H}_j^{\top} {w}_1 \right)\}}\biggr),~ j=1, \ldots, l_{-}$; $\mathfrak{L}_2\left({w}_2\right)=\sum_{i=1}^{l_{+}}
 \frac{1}{\lambda} \biggl( 1- \frac{1}{1+\lambda \left(1-{G}_i^{\top} {w}_2\right)^2 \exp \{a\left(1-{G}_i^{\top} {w}_2 \right)\}}\biggr),~ i=1, \ldots, l_{+}$. ${G}_i$ is the $i^{th}$ column of matrix ${G}$ and ${H}_j$ is the $j^{th}$ column of matrix ${H}$. ${G}=\left[{X}_{+}, {e}_1\right]^{\top} \in \mathbb{R}^{(n+1) \times l_{+}}$, ${H}=\left[{X}_{-}, {e}_2\right]^{\top} \in \mathbb{R}^{(n+1) \times l_{-}}$ ; ${w}_1=\left[{w}_{+}^{\top}, b_{+}\right]^{\top} \in$ $\mathbb{R}^{n+1}$, ${w}_2=\left[{w}_{-}^{\top}, b_{-}\right]^{\top} \in \mathbb{R}^{n+1}$. Further, for simplification, we use $A_j$ and $B_i$ to represent $\left(1+{H}_j^{\top} {w}_1 \right)$ and $\left(1-{G}_i^{\top} {w}_2 \right)$, respectively.

In accordance with the optimality condition, we obtain the following:
\begin{align}
& \nabla P_1\left({w}_1\right)=\left({G} {G}^{\top}+C_1 {I}\right) {w}_1+ \hat{{H}} {s}_1=0, \label{problem1}\\
& \nabla P_2\left({w}_2\right)=\left({H} {H}^{\top}+C_3 {I}\right) {w}_2 - \hat{{G}} {s}_2=0, \label{problem2}
\end{align}
where $\hat{{G}}=\left[C_4  {G}_1, \ldots, C_4 {G}_{l_{+}}\right] \in \mathbb{R}^{(n+1) \times l_{+}},~ \hat{{H}}=\left[C_2 {H}_1, \ldots, C_4 {H}_{l_{-}}\right] \in \mathbb{R}^{(n+1) \times l_{-}}$. ${s}_1=\left[s_{11}, \ldots, s_{1 l_{-}}\right]^{\top} \in \mathbb{R}^{l_{-}}$,~ ${s}_2=\left[s_{21}, \ldots, s_{2 l_{+}}\right]^{\top} \in \mathbb{R}^{l_{+}};~ s_{1 j}= \frac{A_j\left(aA_j+2\right) \exp\left(aA_j\right)}{\{1+\lambda A_j^2 \exp\left(aA_j\right)\}^2},~ j=1, \ldots, l_{-};~ s_{2 i}= \frac{B_i\left(aB_i+2\right) \exp\left(aB_i\right)}{\{1+\lambda B_i^2 \exp\left(aB_i\right)\}^2},~i=1, \ldots, l_{+}$.

Now, we use equations (\ref{problem1}) and (\ref{problem2}) to formulate iterative equations for optimization problems (\ref{optimizationproblem1}) and (\ref{optimizationproblem2}) in the following manner:
\begin{align} 
&{w}_1^{t+1}=-\left({G} {G}^{\top}+C_1 {I}\right)^{-1} \hat{{H}} {s}_1^t,\label{iterativeequation1}\\
&{w}_2^{t+1}=\left({H} {H}^{\top}+C_3 {I}\right)^{-1} \hat{{G}} {s}_2^t \label{iterativeequation2}.
\end{align}
Here, $t$ represents the number of iteration. The iterative procedure involves iterating through equations (\ref{iterativeequation1}) and (\ref{iterativeequation2}) until convergence is achieved. After obtaining the solutions,  we can proceed to find the pair of hyperplanes (\ref{linear twin hyperplanes}). 

To ascertain the class of a unseen sample ${\widetilde{x}} \in \mathbb{R}^n$, we use the following decision rule: 
\begin{align}
\text{Class of}~ {\widetilde{x}}= \begin{cases}+1, & \text { if } \frac{\left|{w}_{+}^{\top} {\widetilde{x}}+b_{+}\right|}{\left\|{w}_{+}\right\|} \leq \frac{\left|{w}_{-}^{\top} {\widetilde{x}}+b_{-}\right|}{\left\|{w}_{-}\right\|}, \\ -1, & \text { otherwise. }\end{cases}
\end{align}

\subsubsection{Non-linear Wave-TSVM}
In the case of non-linearity, data points are separated linearly in a higher-dimensional feature space by utilizing the kernel trick to map them to a higher feature space. The objective of non-linear Wave-TSVM is to identify a pair of hypersurfaces in the following manner:
\begin{align} \label{hypersurface for nonlinear twin wave}
\mathcal{\mathcal{K}}\left({x}, {X}^{\top}\right) {v}_{+}+b_{+}=0 \quad \text { and } \quad \mathcal{K}\left({x}, {X}^{\top}\right) {v}_{-}+b_{-}=0,
\end{align}
where ${X}=\left[{X}_{+} ; {X}_{-}\right]^{\top}$ and $\mathcal{K}(\cdot, \cdot)$ is the kernel function.
To determine the hypersurfaces (\ref{hypersurface for nonlinear twin wave}), we formulate the following optimization problems:\\
(Non-linear Wave-TSVM-1)
\begin{align} \label{Non-linear Wave-TSVM-1}
& \min _{{v}_{+}, b_{+}, \zeta^{-}} \sum_{i=1}^{l_{+}} \frac{1}{2}\left(\mathcal{K}\left({x}_i, {X}^{\top}\right) {v}_{+}+b_{+}\right)^2+\frac{1}{2} C_1\left(\left\|{v}_{+}\right\|_2^2+b_{+}^2\right)+C_2 \sum_{j=1}^{l_{-}} \zeta_j^{-}, \nonumber \\
& \text {s.t. } \quad \zeta_j^{-}= \frac{1}{\lambda} \biggl( 1- \frac{1}{1+\lambda \left(1+\mathcal{K}\left({x}_j, {X}^{\top}\right) {v}_{+}+b_{+}\right)^2 \exp \{a\left(1+\mathcal{K}\left({x}_j, {X}^{\top}\right) {v}_{+}+b_{+} \right)\}}\biggr), \nonumber\\
& \quad j=1, \ldots, l_{-},
\end{align}
(Non-linear Wave-TSVM-2)
\begin{align} \label{Non-linear Wave-TSVM-2}
&\min _{{v}_{-}, b_{-}, \zeta^{+}} \sum_{j=1}^{l_{-}} \frac{1}{2}\left(\mathcal{K}\left({x}_j, {X}^{\top}\right) {v}_{-}+b_{-}\right)^2+\frac{1}{2} C_3\left(\left\|{v}_{-}\right\|_2^2+b_{-}^2\right)+C_4 \sum_{i=1}^{l_{+}} {\zeta}_i^{+}, \nonumber\\
&\text {s.t. } \quad \zeta_i^{+}= \frac{1}{\lambda} \biggl( 1- \frac{1}{1+\lambda \left(1-\mathcal{K}\left({x}_i, {X}^{\top}\right) {v}_{-}-b_{-}\right)^2 \exp \{a\left(1-\mathcal{K}\left({x}_i, {X}^{\top}\right) {v}_{-}-b_{-} \right)\}}\biggr), \nonumber\\
&\quad i=1, \ldots, l_{+}.
\end{align}
The method for solving problems (\ref{Non-linear Wave-TSVM-1}) and
(\ref{Non-linear Wave-TSVM-2}) is akin to the linear scenario. The iterative method corresponds to equations (\ref{Non-linear Wave-TSVM-1}) and
(\ref{Non-linear Wave-TSVM-2}) can be derived as follows:
\begin{align} 
{v}_1^{{t}+1}= &- \left({M M}^{\top}+C_1 {I}\right)^{-1} \nonumber\\
&\left(\sum_{j=1}^{l_{-}} C_2 {N}_j \frac{\left(1+{N}_j^{\top}{v}_1^t\right)\left(a\left(1+{N}_j^{\top}{v}_1^t\right)+2\right) \exp\left(a\left(1+{N}_j^{\top}{v}_1^t\right)\right)}{\{1+\lambda \left(1+{N}_j^{\top}{v}_1^t\right)^2 \exp\left(a\left(1+{N}_j^{\top}{v}_1^t\right)\right)\}^2}\right),\label{iterative equation1 nonlinear}\\
{v}_2^{t+1}= & \left({N} {N}^{\top}+C_3 {I}\right)^{-1} \nonumber\\
& \left(\sum_{i=1}^{l_{+}} C_4 {M}_i \frac{\left(1-{M}_i^{\top} {v}_2^t\right)\left(a\left(1-{M}_i^{\top} {v}_2^t\right)+2\right) \exp\left(a\left(1-{M}_i^{\top} {v}_2^t\right)\right)}{\{1+\lambda \left(1-{M}_i^{\top} {v}_2^t\right)^2 \exp\left(a\left(1-{M}_i^{\top} {v}_2^t\right)\right)\}^2} \right). \label{iterative equation2 nonlinear} 
\end{align}
Here ${M}=\left[\mathcal{K}\left({X}_{+}, {X}^{\top}\right), {e}_1\right]^{\top} \in \mathbb{R}^{(l+1) \times l_{+}}$ , ${N}=\left[\mathcal{K}\left({X}_{-}, {X}^{\top}\right), {e}_2\right]^{\top} \in \mathbb{R}^{(l+1) \times l_{-}}$; ${M}_i$ is the $i^{th}$ column of the matrix ${M}$, ${N}_j$ is the $j^{th}$ column of the matrix ${N}$. ${v}_1=\left[{v}_{+}^{\top}, b_{+}\right]^{\top}$, ${v}_2=\left[{v}_{-}^{\top}, b_{-}\right]^{\top}$. Further, for simplification, we use $E_j$ and $F_i$ to represent $\left(1+{N}_j^{\top} {v}_1 \right)$ and $\left(1-{M}_i^{\top} {v}_2 \right)$, respectively.

It's important to highlight that equations (\ref{iterative equation1 nonlinear}) and (\ref{iterative equation2 nonlinear}) entail the intricate calculation of matrix inversion. Therefore, we utilized the Sherman-Morrison-Woodbury (SMW) theorem \cite{kumar2009least} to alleviate computational complexity. Subsequently, in equations (\ref{iterative equation1 nonlinear}) and (\ref{iterative equation2 nonlinear}), the inverse matrices are substituted with the following matrices:
\begin{align} 
{Q_1}=&\frac{1}{C_1}\left(I-{M}\left(C_1 {I}+{M}^{\top} {M}\right)^{-1} {M}^{\top}\right),\label{new inverse matrix 1}\\
{Q_2}=&\frac{1}{C_3}\left({I}-{N}\left(C_3 {I}+{N}^{\top} {N}\right)^{-1} {N}^{\top}\right).\label{new inverse matrix 2}
\end{align}
Using the equations (\ref{new inverse matrix 1}) and (\ref{new inverse matrix 2}), the iterative approach can be derived in the following manner:
\begin{align} 
{v}_1^{{t}+1}= &- Q_1 \hat{{N}} {s}_1^t,\label{updated iterative equation1 nonlinear}\\ 
{v}_2^{t+1}= & ~Q_2 \hat{{M}} {s}_2^t, \label{updated iterative equation2 nonlinear}
\end{align}
where $t$ denotes the number of iteration. Also, $\hat{{N}}=\left[C_2 {N}_1, \ldots, C_2  {N}_{l_{-}}\right] \in \mathbb{R}^{(l+1) \times l_{-}}$, and $\hat{{M}}=\left[C_4 {M}_1, \ldots, C_4 {M}_{l_{+}}\right]$ $\in \mathbb{R}^{(l+1) \times l_{+}}$. Additionally, ${s}_1^t \in \mathbb{R}^{l_{-}}$, $s_{1j}^t=\frac{E_j^{t}\left(aE_j^{t}+2\right) \exp\left(aE_j^{t}\right)}{\{1+\lambda {E_j^{t}}^2 \exp\left(aE_j^{t}\right)\}^2}$, $j=1, \ldots, l_{-}$ ; ${s}_2^t \in \mathbb{R}^{l_{+}}$, $s_{2i}^t=\frac{F_i^{t}\left(aF_i^{t}+2\right) \exp\left(aF_i^{t}\right)}{\{1+\lambda {F_i^{t}}^2 \exp\left(aF_i^{t}\right)\}^2}$, $i=1, \ldots, l_{+}$. The iterative procedure can be established by iterating through equations (\ref{updated iterative equation1 nonlinear}) and (\ref{updated iterative equation2 nonlinear}) until convergence is achieved. Consequently, upon obtaining the solutions ${v}_{+}, b_{+}$ and ${v}_{-}, b_{-}$, we can then determine the positive and negative hypersurfaces generated by the kernel.

For a new sample ${\widetilde{x}} \in \mathbb{R}^n$, we use the following decision function:
\begin{align}
\text {Class of }~ {\widetilde{x}}= \begin{cases}+1, & \text { if } \frac{\left|\mathcal{K}\left({\widetilde{x}}, {X}^{\top}\right) {v}_{+}+b_{+}\right|}{\sqrt{{v}_{+}^{\top} \mathcal{K}\left({X}, {X}^{\top}\right) {v}_{+}}} \leq \frac{\left|\mathcal{K}\left({\widetilde{x}}, {X}^{\top}\right) {v}_{-}+b_{-}\right|}{\sqrt{{v}_{-}^{\top} \mathcal{K}\left({X}, {X}^{\top}\right) {v}_{-}}}, \\ -1, & \text { otherwise. }\end{cases}
\end{align}
The iterative algorithm structure for non-linear Wave-TSVM subproblem (\ref{Non-linear Wave-TSVM-1}) is clearly described in Algorithm \ref{algorithm2}. The structure for subproblem (\ref{Non-linear Wave-TSVM-2}) is similar to it.
\begin{tiny}
\begin{algorithm}
  \caption{ Non-linear Wave-TSVM}
  \label{algorithm2}
 
   \begin{algorithmic}
  \State \textbf{Input:}
     \State Training dataset:  $\left\{x_i,y_i\right\}_{i=1}^l$, $y_i \in\{-1,1\}$;
     \State The parameters: Convergence precision ($\eta$), maximum iteration number ($T$), parameter $C_1$ and $C_2$, wave loss parameters $a$ and $\lambda$, iteration number $t=0$;    
\State Initialize: $v_1^{0}$;
\State \textbf{Ensure:} $v_{+}$, $b_{+}$;  
\State $1:$ ${M}=\left[\mathcal{K}\left({X}_{+}, {X}^{\top}\right), {e}_1\right]^{\top}$ , ${N}=\left[\mathcal{K}\left({X}_{-}, {X}^{\top}\right), {e}_2\right]^{\top}$.
\State $2:$  \textbf{while} $t \leq T$ \textbf{do}
\State $3:$  ~~~~~\textbf{for} $j \leftarrow 1$ to $l_{-}$ \textbf{do}
\State $4:$ ~~~~~~~~~$s_{1j}^t  \leftarrow \frac{E_j^{t}\left(aE_j^{t}+2\right) \exp\left(aE_j^{t}\right)}{\{1+\lambda {E_j^{t}}^2 \exp\left(aE_j^{t}\right)\}^2}.$
\State $5:$  ~~~~~\textbf{end for}
\State $6:$  ~~~~~${v}_1^{{t}+1} \leftarrow - Q_1 \hat{{N}} {s}_1^t$
\State $7:$  ~~~~~\textbf{if }$\left\|{v}_1^{t+1}-{v}_1^t\right\|<\eta$ \textbf{then}
\State $8:$ ~~~~~~~\text{break}
\State $9:$  ~~~~~\textbf{else}
\State $10:$  ~~~~~~~$t \leftarrow t+1$
\State $11:$  ~~~~\textbf{end if}
\State $12:$  \textbf{end while}
\State $13:$  $\left({v}_{+}^{\top}, b_{+}\right)^{\top}={v}_1^{t+1}$.
 \end{algorithmic}
\end{algorithm}
\end{tiny}

\subsection{Computational Complexity}
In this subsection, we discuss the computational complexity of the proposed Wave-SVM and Wave-TSVM.  Let $l$ and $n$ denote the number of samples and features in training dataset, respectively, and $k$ denote the size of the mini-batch. The $l_{+}$ and $l_{-}$ denote the count of positive and negative samples, respectively. The computational complexity of Wave-SVM is dominated by the computation of the gradients, characterized by a complexity of $\mathcal{O}(k^2)$. Further, the process of computing the moving averages of the gradients' first and second moments, as well as performing bias correction for these averages, both require $\mathcal{O}(k)$ operations, as we have $k$ model parameter to compute. Therefore, the overall computational complexity of the Wave-SVM can be considered as $\mathcal{O}(T(k^{2} + k))$, where $T$ represents the maximum number of iterations. This encapsulates the fundamental operations involved in training the Wave-SVM and underscores its scalability concerning the number of iterations and features. The computational complexity of Wave-TSVM primarily arises from the computation of matrix inversion. In the linear case, the algorithm requires solving the inverse of a $(n + 1) \times (n + 1)$ matrix, which results in a time complexity of $\mathcal{O}((n + 1)^3)$. For the non-linear case, the algorithm needs to compute the inverse of two matrices: one of size $l_{+} \times l_{+}$ and the other of size $l_{-} \times l_{-}$, with computational complexities of $\mathcal{O}(l_{+}^3)$ and $\mathcal{O}(l_{-}^3)$, respectively.  Hence, for non-linear Wave-TSVM, the time complexity is $\mathcal{O}(T (l_{+}^3 + l_{-}^3))$. It is evident that non-linear Wave-TSVM is not well-suited for large-scale datasets due to its cubic computational complexity growth in relation to the sample size.

\section{\textbf{Numerical Experiments}}
In this section, we assess the proposed Wave-SVM and Wave-TSVM against the baseline models, including C-SVM \cite{cortes1995support}, Pin-SVM \cite{huang2013support}, LINEX-SVM \cite{ma2019linex}, FP-SVM \cite{kumari2024diagnosis}, TSVM \cite{khemchandani2007twin}, Pin-GTSVM \cite{tanveer2019general}, SLTSVM \cite{si2023symmetric}, and IF-RVFL \cite{malik2022alzheimer}. For experiments, we downloaded $42$ binary datasets of diverse domains from the UCI \cite{dua2017uci} and KEEL \cite{derrac2015keel} repositories. The detailed description of the datasets is provided in Table S.I of the supplementary file. Additionally, we assess the models on the Alzheimer’s disease (AD) dataset, available on the Alzheimer’s Disease Neuroimaging Initiative (ADNI) $(adni.loni.usc.edu)$. 
\subsection{Experimental Setup and Parameter Selection}
All the experiments are implemented using MATLAB R2023a on window 10 running on a PC with configuration  Intel(R) Core(TM) i7-6700 CPU @ 3.40GHz, 3408 Mhz, 4 Core(s), 8 Logical Processor(s) with 16 GB of RAM. For each SVM type model (C-SVM \cite{cortes1995support}, Pin-SVM \cite{huang2013support}, LINEX-SVM \cite{ma2019linex}, FP-SVM \cite{kumari2024diagnosis}, and proposed Wave-SVM), the regularization parameter $C$ is selected from the set $\{10^i\,|\,i=-6, -5,\ldots, 5, 6\}$. For Pin-SVM, the hyperparameter $\tau$ is choosen from the set $\{0,0.1,\ldots, 0.9,1\}$. For LINEX-SVM and FP-SVM, the loss hyperparameters are selected the same as in \cite{ma2019linex} and \cite{kumari2024diagnosis}, respectively. For the proposed Wave-SVM, loss hyperparameters $\lambda$ and $a$ are selected from the ranges $\left[0.1:0.2:2\right]$ and  $\left[-2:0.1:5\right]$, respectively. For the non-linear case, Gaussian kernel function $\mathcal{K}\left(x_j,x_k\right)=\exp\left(-\left\|x_j-x_k\right\|^2 / \sigma^2\right)$ (where $\sigma$ is the kernel parameter) is considered. The Gaussian kernel parameter $\sigma$ for each SVM type model is selected from the set $\{10^i\,|\,i=-6, -5, \ldots, 5, 6\}$. The parameters for the Adam algorithm are experimentally set as: (\RNum{1}) initial weight $w_{0}= 0.01$, (\RNum{2}) initial first moment $m_{0}=0.01$, (\RNum{3}) initial second moment $v_{0}=0.01$, (\RNum{4}) initial learning rate $\alpha$ is selected from the set $\{0.0001,0.001, 0.01\}$, (\RNum{5}) exponential decay rate of first order $\beta_{1}=0.9$, (\RNum{6}) exponential decay rate of second order $\beta_{2}=0.999$, (\RNum{7}) error tolerance $\eta= 10^{-5}$, (\RNum{8}) division constant $\epsilon=10^{-8}$, (\RNum{9}) mini-batch size $k=2^{5}$, (\RNum{10}) maximum iteration number $T=1000$.

For each TSVM type model (TSVM \cite{khemchandani2007twin}, Pin-GTSVM \cite{tanveer2019general}, SLTSVM \cite{si2023symmetric}, and proposed Wave-TSVM), the regularization parameters ($C_1$ and $C_3$) and kernel parameter ($\sigma$) are selected from the set $\{10^i\,|\,i=-6, -4,\ldots, 4, 6\}$. For Pin-GTSVM, the hyperparameter $\tau$ is selected the same as in \cite{tanveer2019general}. For SLTSVM and the proposed Wave-TSVM, the structural risk minimization parameters $C_2$ and $C_4$ are selected from the set $\{10^i\,|\,i=-6, -4,\ldots, 4, 6\}$. For SLTSVM and the proposed Wave-TSVM, we set $C_1=C_3$ and $C_2=C_4$. For iterative algorithm, to solve SLTSVM and the proposed Wave-TSVM, we set $\eta= 10^{-5}$ and $T=50$. To suppress the tuning time for the proposed Wave-TSVM on UCI and KEEL datasets, the loss hyperparameter $\lambda$ is fixed to $1$, and $a$ is selected from the range $\left[-2:0.5:2\right]$. For IF-RVFL \cite{malik2022alzheimer}, the regualrization parameter ($C$) and kernel parameter ($\sigma$) are selected from the set $\{10^i\,|\,i=-6, -4,\ldots, 4, 6\}$, and the number of hidden nodes ($N_d$) is selected from the range $\left[3:20:203\right]$ \cite{zhang2016comprehensive}.

The performance of the models significantly depends on the selection of parameters \cite{cristianini2000introduction}. In order to adjust them, we tune the parameters based on k-fold (k$=4$) cross-validation and grid search. The benefit behind k-fold cross-validation is that it ensures that every sample will eventually be included in both the training and testing sets. More specifically, the dataset is divided into four non-overlapping subsets at random, one of which is set aside as a test set and the other three as train set. This process is conducted four times, and at every setup, a different test set is chosen while the other three sets are used as training, and the best of the four testing results is used as the performance measure for each model. For all the experiments, we normalized the data in the range of $[0, 1]$.

To assess the efficacy of the proposed Wave-SVM and Wave-TSVM against baseline models, we employed the accuracy metric, defined as follows:
\begin{align}{}
 \text {Accuracy}=\frac{\text { T.P.+T.N.  }}{\text { T.P.+T.N.+F.P.+F.N.}} \times 100.   
\end{align}
Here, T.P., T.N., F.P., and F.N. represent the true positive, true negative, false positive, and false negative, respectively. Further, we conducted a comparative analysis of the training time between the proposed Wave-SVM and the baseline models. The reported times only include
the time taken for training the models with the optimized hyperparameters.

\subsection{Evaluation on UCI and KEEL datasets}
For a fair comparison of the proposed Wave-SVM and Wave-TSVM with baseline models, we compare Wave-SVM with C-SVM, Pin-SVM, LINEX-SVM, and FP-SVM, and Wave-TSVM with TSVM, Pin-GTSVM, and SLTSVM. In essence, we compared Wave-SVM and Wave-TSVM with SVM type models and TSVM type models, respectively. Additionally, we compare Wave-TSVM with IF-RVFL.

First, we discuss the experimental outcomes of Wave-SVM. For linear case, the average classification accuracy and training time of the proposed Wave-SVM and the baseline models (C-SVM, Pin-SVM, LINEX-SVM, and FP-SVM) are presented in Table \ref{tab:lineartable}. The detailed experimental results for each dataset are presented in Table S.VI of the supplementary file. The average accuracy of the existing C-SVM, Pin-SVM, LINEX-SVM, and FP-SVM are $82.21 \%$, $89.1 \%$, $77.41\%$, and $89.73\%$ respectively, whereas, the average accuracy of the proposed Wave-SVM is  $91.15 \%$, surpassing the baseline models.
The average training time, expressed in seconds, of C-SVM, Pin-SVM, LINEX-SVM, FP-SVM, and the proposed Wave-SVM are $4.6266$s,	$7.074$s, $0.0033$s, $2.254$, and $0.0049$s, respectively. This observation strongly underscores the substantial superiority of the proposed Wave-SVM over C-SVM and Pin-SVM. However, its training time is marginally extended compared to LINEX-SVM. Nonetheless, it still demonstrates commendable performance in terms of training time. The optimal parameters of linear Wave-SVM and baseline models corresponding to the accuracy values are presented in Table S.VII of the supplementary file.
For the non-linear case, the average accuracy and training time of the proposed Wave-SVM and baseline models are presented in Table \ref{tab:non-linear Wave-SVM table}. The detailed experimental results of non-linear Wave-SVM against baseline models for each dataset are presented in Table S.VIII of the supplementary file. The average accuracy of C-SVM, Pin-SVM, LINEX-SVM, FP-SVM, and the proposed Wave-SVM are  $89.46 \%$, $89.97 \%$, $86.63 \%$, $90.18 \%$, and $91.92 \%$, respectively. Evidently, the proposed Wave-SVM achieves the highest classification accuracy in comparison with the baseline models. The average training time of C-SVM, Pin-SVM, LINEX-SVM, FP-SVM, and the proposed Wave-SVM are $7.57$s, $8.64$s, $0.0137$s, $1.39$s, and	$0.0219$s, respectively.
The outcomes manifest that the proposed Wave-SVM exhibits remarkable efficiency. Its training time is significantly less than C-SVM and Pin-SVM, indicating its enhanced computational efficiency. However, the proposed Wave-SVM takes a slightly longer time to train than the LINEX-SVM, although the difference in training time is minor. It is noteworthy that the comparative evaluation of accuracy and training time shows the overall performance of the models. Thus, the proposed Wave-SVM's ability to strike a balance between classification accuracy and training time makes it a superior model in comparison to the baseline models. The optimal parameters of non-linear Wave-SVM and baseline models corresponding to the accuracy values are presented in Table S.IX of the supplementary file.

To assess the TSVM type models, given the substantial computational complexity of the proposed Wave-TSVM due to the matrix inversion, we excluded datasets with either over $5000$ samples or more than $25$ features. This exclusion resulted in a remaining set of $32$ datasets. The average experimental outcomes for the non-linear proposed Wave-TSVM and baseline models, including TSVM, Pin-GTSVM, SLTSVM, and IF-RVFL, across these $3$2 datasets are presented in Table \ref{tab:non-linear Twin Table}. For detailed results on each dataset, refer to Table S.X of the supplementary file. The average classification accuracies of the existing TSVM, Pin-GTSVM, IF-RVFL, and SLTSVM are $86.49\%$, $83.24\%$, $83.8\%$, and $85.98\%$, respectively. In contrast, the average accuracy of the proposed Wave-TSVM is $87.91\%$, surpassing that of baseline models. These findings strongly emphasize the significant superiority of the proposed Wave-TSVM over the baseline models. The optimal parameters of non-linear Wave-TSVM and baseline models corresponding to the accuracy values are outlined in Table S.XI of the supplementary file.

Further, to offer additional substantiation for the enhanced efficacy of the proposed Wave-SVM and Wave-TSVM, we conducted a thorough statistical analysis of the models, for which we followed four tests: ranking scheme, Friedman test, Nemenyi post hoc test, and Win-Tie-Loss sign test. A detailed discussion of the statistical tests and their results is presented in Section S.II of the supplementary file.
\begin{table}[htp]
\centering
\caption{Average accuracy, training time and rank for linear Wave-SVM against baseline models on benchmark UCI and KEEL datasets.}
\label{tab:lineartable}
\begin{tabular}{lccccc}
\hline
Dataset & C-SVM \cite{cortes1995support} & Pin-SVM \cite{huang2013support} & LINEX-SVM \cite{ma2019linex} & FP-SVM \cite{kumari2024diagnosis} & Wave-SVM$^{\dagger}$ \\ \hline
\textbf{Avg. Acc.} & 82.21 & 89.1 & 77.41 & 89.73 & \textbf{91.15} \\ \hline
\textbf{Avg. Time} & 4.6266 & 7.074 & \textbf{0.0033} & 2.254 & 0.0049 \\ \hline
\textbf{Avg. Rank} & 3.73 & 2.52 & 4.29 & 2.49 & \textbf{1.68} \\ \hline
\multicolumn{6}{l}{$^{\dagger}$ represents the proposed model.}\\
\multicolumn{6}{l}{Here, Avg. and Acc. are acronyms used for average and accuracy, respectively.}
\end{tabular}
\end{table}
\begin{table}[htp]
\centering
\caption{Average accuracy, training time and rank for non-linear Wave-SVM against baseline models using Gaussian kernel function on benchmark UCI and KEEL datasets.}
\label{tab:non-linear Wave-SVM table}
\begin{tabular}{lccccc}
\hline
Dataset & C-SVM \cite{cortes1995support} & Pin-SVM \cite{huang2013support} & LINEX-SVM \cite{ma2019linex} & FP-SVM \cite{kumari2024diagnosis} & Wave-SVM$^{\dagger}$ \\ \hline
\textbf{Avg. Acc.} & 89.46 & 89.97 & 86.63 & 90.18 & \textbf{91.92} \\ \hline
\textbf{Avg. Time} & 7.57 & 8.64 & \textbf{0.0137} & 1.39 & 0.0219 \\ \hline
\textbf{Avg. Rank}& 3.37 & 2.85 & 4.16 & 2.53 & \textbf{1.8} \\ \hline
\multicolumn{6}{l}{$^{\dagger}$ represents the proposed model.}\\
\multicolumn{6}{l}{Here, Avg. and Acc. are acronyms used for average and accuracy, respectively.}
\end{tabular}
\end{table}
\begin{table}[htp]
\centering
\caption{Average accuracy and rank for non-linear Wave-TSVM against baseline models on benchmark UCI and KEEL datasets.}
\label{tab:non-linear Twin Table}
\begin{tabular}{lccccc}
\hline
Dataset & TSVM \cite{khemchandani2007twin} & Pin-GTSVM \cite{tanveer2019general} & IF-RVFL \cite{malik2022alzheimer} & SLSTSVM \cite{si2023symmetric} & Wave-TSVM$^{\dagger}$ \\ \hline
\textbf{Avg. Acc.} & 86.49 & 83.24 & 83.8 & 85.98 & \textbf{87.91} \\ \hline
\textbf{Avg. Rank} & 2.78 & 3.55 & 3.75 & 2.92 & \textbf{2} \\ \hline
\multicolumn{6}{l}{$^{\dagger}$ represents the proposed model.}\\
\multicolumn{6}{l}{Here, Avg. and Acc. are acronyms used for average and accuracy, respectively.}
\end{tabular}
\end{table}
\begin{table}[htp]
\centering
\caption{Performance comparison of the proposed Wave-SVM compare baseline models on benchmark UCI and KEEL datasets with Gaussian noise.}
\label{tab:Wave-SVM noise table}
\resizebox{\textwidth}{!}{%
\begin{tabular}{lcccccc}
\hline
Model & \multicolumn{1}{l}{} & C-SVM \cite{cortes1995support} & Pin-SVM \cite{huang2013support} & LINEX-SVM \cite{ma2019linex} & FP-SVM \cite{kumari2024diagnosis} & Wave-SVM$^{\dagger}$ \\
Dataset & Noise & Accuracy & Accuracy & Accuracy & Accuracy & Accuracy \\ \hline
breast\_cancer\_wisc & 5\% & 93.68 & 97.7 & 100 & 100 & 100 \\
 & 10\% & 94.25 & 97.7 & 100 & 100 & 100 \\
 & 20\% & 87.36 & 97.7 & 100 & 100 & 100 \\
 & 30\% & 85.14 & 98.85 & 100 & 100 & 100 \\ \hline
\textbf{Avg. Acc.} & \multicolumn{1}{l}{} & 90.11 & \underline{97.99} & \textbf{100} & \textbf{100} & \textbf{100} \\ \hline
credit\_approval & 5\% & 85.55 & 84.88 & 91.33 & 94.8 & 97.11 \\
 & 10\% & 86.13 & 85.47 & 89.02 & 95.38 & 96.53 \\
 & 20\% & 84.97 & 86.13 & 89.02 & 95.38 & 92.49 \\
 & 30\% & 84.39 & 84.97 & 89.02 & 94.8 & 93.64 \\ \hline
\textbf{Avg. Acc.} & \multicolumn{1}{l}{} & 85.26 & 85.36 & 89.6 & \textbf{95.09} & \underline{94.94} \\ \hline
horse\_colic & 5\% & 70.65 & 76.09 & 81.52 & 85.87 & 86.96 \\
 & 10\% & 70.65 & 73.91 & 82.61 & 84.78 & 86.96 \\
 & 20\% & 69.57 & 71.74 & 80.43 & 88.04 & 86.96 \\
 & 30\% & 68.48 & 69.57 & 80.43 & 86.96 & 84.78 \\ \hline
\textbf{Avg. Acc.} & \multicolumn{1}{l}{} & 69.84 & 72.83 & \underline{81.25} & \textbf{86.41} & \textbf{86.41} \\ \hline
led7digit-0-2-4-5-6-7-8-9\_vs\_1 & 5\% & 96.4 & 96.4 & 96.4 & 97.3 & 100 \\
 & 10\% & 96.4 & 96.4 & 93.69 & 98.2 & 98.2 \\
 & 20\% & 96.4 & 96.4 & 95.45 & 97.3 & 98.2 \\
 & 30\% & 96.36 & 96.4 & 94.59 & 97.3 & 99.1 \\ \hline
\textbf{Avg. Acc.} & \multicolumn{1}{l}{} & 96.39 & 96.4 & 95.03 & \underline{97.52} & \textbf{98.88} \\ \hline
monk1 & 5\% & 56.12 & 53.96 & 56.83 & 59.71 & 65.47 \\
 & 10\% & 54.68 & 57.55 & 53.96 & 58.27 & 66.19 \\
 & 20\% & 55.4 & 56.83 & 56.12 & 58.27 & 64.03 \\
 & 30\% & 58.27 & 56.83 & 56.83 & 61.87 & 64.75 \\ \hline
\textbf{Avg. Acc.} & \multicolumn{1}{l}{} & 56.12 & 56.29 & 55.93 & \underline{59.53} & \textbf{65.11} \\ \hline
yeast2vs8 & 5\% & 98.35 & 98.35 & 98.35 & 99.17 & 99.17 \\
 & 10\% & 98.35 & 98.35 & 98.35 & 99.17 & 100 \\
 & 20\% & 98.35 & 98.35 & 98.35 & 99.17 & 100 \\
 & 30\% & 98.35 & 98.35 & 98.35 & 99.17 & 100 \\ \hline
\textbf{Avg. Acc.} & \multicolumn{1}{l}{} & 98.35 & 98.35 & 98.35 & \underline{99.17} & \textbf{99.79} \\ \hline
\multicolumn{7}{l}{$^{\dagger}$ represents the proposed model.}\\
\multicolumn{7}{l}{Here, Avg. and Acc. are acronyms used for average and accuracy, respectively.}\\
\multicolumn{7}{l}{The boldface and underline indicate the best and second-best models, respectively, in terms of average accuracy.}
\end{tabular}}
\end{table}
\subsection{Evaluation on UCI and KEEL datasets with Gaussian noise}
While the UCI and KEEL datasets used in our evaluation reflect real-world circumstances, it is essential to acknowledge that outliers or noise can arise due to various factors. To showcase the effectiveness of the proposed Wave-SVM and Wave-TSVM, even in adverse conditions, we deliberately added feature noise to selected datasets. We selected 6 diverse datasets for our comparative analysis, namely breast\_cancer\_wisc, credit\_approval, horse\_colic, led7digit-0-2-4-5-6-7-8-9\_vs\_1, monk1, and yeast2vs8. To ensure impartiality in evaluating the models, we selected 2 datasets where the proposed Wave-SVM achieves the highest performance compared to baseline models at 0\% noise level (refer to Table S.VIII of supplementary). Further, we selected 2 datasets where the proposed Wave-SVM does not achieve the highest performance and 2 datasets where the proposed Wave-SVM ties with an existing model. To carry out a comprehensive evaluation, we added Gaussian noise at varying levels of 5\%, 10\%, 20\%, and 30\% to corrupt the features of these datasets. The accuracy values of Wave-SVM against baseline models for the selected datasets with 5\%, 10\%, 20\%, and 30\% level of noise are presented in Table \ref{tab:Wave-SVM noise table}. Out of 6 diverse datasets, the proposed Wave-SVM outperforms the baseline models on 5 datasets and attains second position on 1 dataset, as per the average accuracies at different levels of noise. This outcome unequivocally underscores the significance of the proposed Wave-SVM as a robust model. The optimal parameters of Wave-SVM and baseline models corresponding to the accuracy values on noisy datasets are presented in Table S.IX of the supplementary. Similarly, the accuracy values of Wave-TSVM against baseline models for the selected datasets with 5\%, 10\%, 20\%, and 30\% noise levels are outlined in Table \ref{tab:Twin noise table}. Out of the 6 diverse datasets, the proposed Wave-TSVM surpasses the baseline algorithms on 4 datasets, achieves the second position on 1 dataset, and secures the third position on 1 dataset, based on the average accuracies at different noise levels. This result demonstrate the significance and robustness of Wave-TSVM. The optimal parameters of Wave-TSVM and baseline models corresponding to the accuracy values on noisy datasets are presented in Table S.XI of the supplementary. By subjecting the models to rigorous conditions, we demonstrate the
exceptional performance and superiority of the proposed Wave-SVM and Wave-TSVM, especially in adverse scenarios.
\begin{table}[htp]
\centering
\caption{Performance comparison of the proposed Wave-TSVM compare baseline models on benchmark UCI and KEEL datasets with Gaussian noise.}
\label{tab:Twin noise table}
\resizebox{\textwidth}{!}{%
\begin{tabular}{lcccccc}
\hline
Model & \multicolumn{1}{l}{} & TSVM \cite{khemchandani2007twin} & Pin-GTSVM \cite{tanveer2019general} & IF-RVFL \cite{malik2022alzheimer} & SLSTSVM \cite{si2023symmetric} & Wave-TSVM$^{\dagger}$ \\ \hline
Dataset & Noise & Accuracy & Accuracy & Accuracy & Accuracy & Accuracy \\ \hline
breast\_cancer\_wisc & 5\% & 88.01 & 88.68 & 83.57 & 78.16 & 87.36 \\
 & 10\% & 88.01 & 88.68 & 83.57 & 78.16 & 85.63 \\
 & 20\% & 86.9 & 87.37 & 83.57 & 78.16 & 83.91 \\
 & 30\% & 84.37 & 85.5 & 83.57 & 78.16 & 80.46 \\ \hline
\textbf{Avg. Acc.} & \multicolumn{1}{l}{} & \underline{86.82} & \textbf{87.56} & 83.57 & 78.16 & 84.34 \\ \hline
credit\_approval & 5\% & 91.84 & 90.18 & 80.92 & 80.92 & 91.33 \\
 & 10\% & 89.53 & 87.57 & 80.92 & 80.81 & 89.02 \\
 & 20\% & 86.37 & 87.57 & 80.92 & 79.65 & 88.44 \\
 & 30\% & 85.95 & 84.08 & 80.92 & 73.41 & 86.13 \\ \hline
\textbf{Avg. Acc.} & \multicolumn{1}{l}{} & \underline{88.42} & 87.35 & 80.92 & 78.7 & \textbf{88.73} \\ \hline
horse\_colic & 5\% & 75.04 & 82.25 & 77.61 & 75 & 77.17 \\
 & 10\% & 75.04 & 79.9 & 77.61 & 71.74 & 84.78 \\
 & 20\% & 73.22 & 79.9 & 77.61 & 70.65 & 76.09 \\
 & 30\% & 69.04 & 68.86 & 77.61 & 72.83 & 76.09 \\ \hline
\textbf{Avg. Acc.} & \multicolumn{1}{l}{} & 73.1 & \underline{77.73} & 77.61 & 72.55 & \textbf{78.53} \\ \hline
led7digit-0-2-4-5-6-7-8-9\_vs\_1 & 5\% & 95.3 & 93.81 & 89.37 & 95.5 & 98.2 \\
 & 10\% & 95.3 & 87.09 & 89.37 & 95.5 & 95.5 \\
 & 20\% & 93.5 & 87.09 & 89.37 & 93.69 & 95.45 \\
 & 30\% & 91.4 & 84.5 & 89.37 & 93.69 & 94.59 \\ \hline
\textbf{Avg. Acc.} & \multicolumn{1}{l}{} & 93.87 & 88.12 & 89.37 & \underline{94.6} & \textbf{95.93 }\\ \hline
monk1 & 5\% & 55.71 & 52.15 & 61.15 & 61.15 & 58.99 \\
 & 10\% & 56.99 & 52.15 & 61.15 & 60.43 & 60.43 \\
 & 20\% & 58.15 & 53.98 & 61.15 & 57.55 & 60.43 \\
 & 30\% & 59.71 & 51.18 & 61.15 & 61.15 & 60.43 \\ \hline
\textbf{Avg. Acc.} & \multicolumn{1}{l}{} & 57.64 & 52.37 & \textbf{61.15} & \underline{60.07} & \underline{60.07} \\ \hline
yeast2vs8 & 5\% & 99.17 & 88.55 & 89.92 & 97.52 & 98.35 \\
 & 10\% & 98.35 & 88.55 & 89.92 & 98.35 & 99.17 \\
 & 20\% & 97.69 & 88.55 & 89.92 & 98.35 & 98.35 \\
 & 30\% & 96.04 & 88.1 & 89.92 & 97.52 & 98.35 \\ \hline
\textbf{Avg. Acc.} & \multicolumn{1}{l}{} & 97.81 & 88.44 & 89.92 & \underline{97.93} & \textbf{98.55} \\ \hline
\multicolumn{7}{l}{$^{\dagger}$ represents the proposed model.}\\
\multicolumn{7}{l}{Here, Avg. and Acc. are acronyms used for average and accuracy, respectively.}\\
\multicolumn{7}{l}{The boldface and underline indicate the best and second-best models, respectively, in terms of average accuracy.}
\end{tabular}}
\end{table}
\subsection{Results Analysis on ADNI dataset}
Alzheimer's disease (AD) is a widely recognized neurodegenerative illness that progressively results in memory loss and cognitive impairments. AD development is irreversible and manifests as atrophy in the brain's inner regions. It is predicted that by the year $2050$, one out of every $85$ individuals will suffer from AD \cite{porsteinsson2021diagnosis}. Multiple studies indicate that early detection and intervention can slow down the progression of AD. Therefore, to mitigate further growth, treatment should commence at the earliest possible stage. The ADNI project, initiated by Michael W. Weiner in $2003$, aims to examine various neuroimaging techniques, including positron emission tomography (PET), magnetic resonance imaging (MRI), and other diagnostic tests for AD, particularly at the mild cognitive impairment (MCI) stage.
AD data is publicly available on the ADNI repository, which is accessible at $adni.loni.usc.edu$. The pipeline for feature extraction opted in this paper is the same as followed in \cite{richhariya2021efficient}. The dataset contains three cases: control normal (CN) versus AD, CN versus MCI, and MCI versus AD.

The accuracy of the proposed Wave-SVM, as well as baseline models for AD diagnosis, are presented in Table \ref{tab:AD table}. We examine that with an average accuracy of $79.11 \%$, the proposed Wave-SVM is the best model. The average accuracies of baseline algorithms, C-SVM, Pin-SVM, and LINEX-SVM are $70.55$\%, $78.16 \%$, and $75.99 \%$, respectively. With an accuracy of $88.46 \%$ and $79.45 \%$, the proposed Wave-SVM is the most accurate classifier for CN versus AD and MCI versus AD cases. For CN versus MCI case, Pin-SVM came out on top followed by the proposed Wave-SVM. Overall, the proposed Wave-SVM has emerged as the best classifier in comparison to the baseline models for AD diagnosis.

\begin{table}[htp]
\centering
\caption{Results of the proposed non-linear Wave-SVM against baseline models on ADNI dataset.}
\label{tab:AD table}
\resizebox{\textwidth}{!}{%
\begin{tabular}{lcccccccccc}
\hline
Model & \multicolumn{2}{c}{C-SVM \cite{cortes1995support}} & \multicolumn{2}{c}{Pin-SVM \cite{huang2013support}} & \multicolumn{2}{c}{LINEX-SVM \cite{ma2019linex}} & \multicolumn{2}{c}{FP-SVM \cite{kumari2024diagnosis}} & \multicolumn{2}{c}{Wave-SVM$^{\dagger}$} \\
Dataset & Accuracy & Time & Accuracy & Time & Accuracy & Time & Accuracy & Time & Accuracy & Time \\
(\# of samples, \# of features) & ($C$, $\sigma$) & \multicolumn{1}{l}{} & ($\tau$, $C$, $\sigma$) & \multicolumn{1}{l}{} & (a, $C$, $\sigma$) & \multicolumn{1}{l}{} & ($C$, $\sigma$, $\tau_1$, $\tau_2$) & \multicolumn{1}{l}{} & ($\lambda$, a, $C$, $\sigma$) & \multicolumn{1}{l}{} \\ \hline
CN versus AD & 82.26 & 0.0206 & 84.61 & 0.006 & 86.54 & 0.0011 & 84.62 & 0.0668 & 88.46 & 0.0019 \\
(415 ,91) & 1,1 & \multicolumn{1}{l}{} & 0,0.1,0.001 & \multicolumn{1}{l}{} & -1,0.001,1 & \multicolumn{1}{l}{} & 0.000001, 0.000001, 0.5, 0 & \multicolumn{1}{l}{} & 0.1,-0.6,1000,10 & \multicolumn{1}{l}{} \\
CN versus MCI & 63.1 & 0.0247 & 74.52 & 0.0119 & 68.15 & 0.0018 & 75.15 & 0.0241 & 69.43 & 0.0027 \\
(626, 91) & 10,1 & \multicolumn{1}{l}{} & 0.2,0.1,0.001 & \multicolumn{1}{l}{} & -1,1000,10 & \multicolumn{1}{l}{} & 0.0001, 0.01, 0, 0.4 & \multicolumn{1}{l}{} & 0.7,3.2,0.001,10 & \multicolumn{1}{l}{} \\
MCI versus AD & 66.28 & 0.0176 & 75.34 & 0.0502 & 73.29 & 0.0014 & 73.97 & 0.0444 & 79.45 & 0.0024 \\
(585, 91) & 1,1 & \multicolumn{1}{l}{} & 0,10,0.01 & \multicolumn{1}{l}{} & -1,0.001,1 & \multicolumn{1}{l}{} & 1000000, 0.000001, 0, 0.8 & \multicolumn{1}{l}{} & 0.1,0.3,100,10 & \multicolumn{1}{l}{} \\ \hline
\textbf{Avg. Acc. and Avg. Time} & 70.55 & 0.021 & 78.16 & 0.0227 & 75.99 & \textbf{0.0014} & 77.91 & 0.0451 & \textbf{79.11} & 0.0023 \\ \hline
\multicolumn{6}{l}{$^{\dagger}$ represents the proposed model.}\\
  \multicolumn{6}{l}{Here, Avg. and Acc. are acronyms used for average and accuracy, respectively.}
\end{tabular}%
}
\end{table}

\section{Conclusions and Future work}
In this paper, we have emphasized the pivotal role of the loss function within the realm of supervised learning, highlighting how the choice of the appropriate loss function significantly influences the proficiency of the developed models. We introduced an innovative asymmetric loss function named wave loss. This novel loss function is distinguished by its aptitude for handling outliers, resilience against noise, bounded characteristics, and an essential smoothness property. These distinctive attributes collectively position the wave loss function to effectively address the complex challenges posed by real-world data. Further, by showing the possession of classification-calibrated property, we have laid the foundation for its broader adoption in real-world applications. Additionally, the amalgamation of the proposed wave loss into the least squares setting of SVM and TSVM led to the development of two innovative models: Wave-SVM and Wave-TSVM, respectively.  These models offer a robust and smooth alternative to existing classifiers, contributing to the advancement of classification techniques. Notably, the incorporation of the Adam algorithm for optimizing the Wave-SVM brings further innovation, as this is the first time that Adam has been utilized to solve an SVM model. This strategic use of Adam not only improves the efficiency of Wave-SVM but also offers a solution for handling large-scale problems. To tackle the optimization problem of Wave-TSVM, we utilize an iterative algorithm that requires matrix inversion at each iteration, which is intractable for large datasets. In the future, one can reformulate the Wave-TSVM to circumvent the need to compute matrix inversion. The empirical results garnered from a diverse spectrum of UCI and KEEL datasets (with and without feature noise) undeniably affirm the efficacy of the proposed Wave-SVM and Wave-TSVM against baseline algorithms.
\par
However, we did not explore the application of the wave loss function within deep learning models, which presents an avenue for future research. Looking ahead, the crucial smoothness property inherent to the wave loss function suggests promising possibilities for its fusion with advanced machine learning and deep learning techniques. In future, researchers can explore the fusion of the wave loss function with cutting-edge methodologies like support matrix machines to tackle complex real-world problems. The source codes of the proposed models are publicly available at \url{https://github.com/mtanveer1/Wave-SVM}.
\section*{Acknowledgment}
This project is supported by the Indian government's Science and Engineering Research Board (MTR/\\2021/000787) under the Mathematical Research Impact-Centric Support (MATRICS) scheme. The Council of Scientific and Industrial Research (CSIR), New Delhi, provided a fellowship for Mushir Akhtar’s research under grant no. 09/1022(13849)/2022-EMR-I. The collection and sharing of data for this project were supported by funding from the Alzheimer's Disease Neuroimaging Initiative (ADNI) through the National Institutes of Health Grant U01 AG024904 and the DOD ADNI through the Department of Defense award number W81XWH-12-2-0012. ADNI receives funding from the National Institute on Aging, the National Institute of Biomedical Imaging and Bioengineering, and generous contributions from the following sources: AbbVie, Alzheimer’s Association; Alzheimer’s Drug Discovery Foundation; Araclon Biotech; BioClinica, Inc.; Biogen;
Bristol-Myers Squibb Company; CereSpir, Inc.; Cogstate; Eisai Inc.; Elan Pharmaceuticals, Inc.; Eli Lilly and Company; EuroImmun; F. Hoffmann-La Roche Ltd and its affiliated company Genentech, Inc.; Fujirebio; GE Healthcare; IXICO Ltd.; Janssen Alzheimer Immunotherapy Research \& Development, LLC.; Johnson \& Johnson Pharmaceutical Research \& Development LLC.; Lumosity; Lundbeck; Merck \& Co., Inc.; Meso
Scale Diagnostics, LLC.; NeuroRx Research; Neurotrack Technologies; Novartis Pharmaceuticals
Corporation; Pfizer Inc.; Piramal Imaging; Servier; Takeda Pharmaceutical Company; and Transition
Therapeutics. The Canadian Institutes of Health Research is funding ADNI clinical sites in Canada. Private sector donations are managed by the Foundation for the National Institutes of Health (\url{www.fnih.org}). The Northern California Institute for Research and Education is the grantee organization, and the study is coordinated by the Alzheimer’s Therapeutic Research Institute at the University of Southern California. ADNI data are distributed by the Laboratory for Neuro Imaging at the University of Southern California.

\bibliography{refs.bib}
 \bibliographystyle{unsrtnat}
\end{document}


\begin{frontmatter}
\title{Supplementary Material for the Manuscript:\\``Advancing Supervised Learning with the Wave Loss Function: A Robust and Smooth Approach"}
\author[inst1]{Mushir Akhtar}
\author[inst1]{M. Tanveer\corref{Correspondingauthor}}
\author[inst1]{Mohd. Arshad}
\author[]{for the Alzheimer’s Disease Neuroimaging
Initiative\corref{ADNI information}}
\affiliation[inst1]{organization={Department of Mathematics},
            addressline={Indian Institute of Technology Indore}, 
            city={Simrol, Indore},
            postcode={453552}, 
            country={India}}
            \cortext[Correspondingauthor]{Corresponding author}
            \cortext[ADNI information]{Data used in preparation of this article were obtained from the Alzheimer’s Disease Neuroimaging Initiative
(ADNI) database (adni.loni.usc.edu). As such, the investigators within the ADNI contributed to the design and implementation of ADNI and/or provided data but did not participate in analysis or writing of this report.}
\end{frontmatter}
\section{Proposed Work}
\textbf{Mathematical justification for the smoothness of wave loss function:}\\
The proposed wave loss function is formulated with continuous and exponential components, ensuring its overall differentiability. However, to alleviate concerns regarding its differentiability, we provide a rigorous mathematical justification. Specifically, we demonstrate that for any arbitrary $u \in \mathbb{R}$, the left-hand derivative ($L.H.D.$) and right-hand derivative ($R.H.D.$) are equivalent, affirming the function's smoothness and analytical tractability.
\begin{proof}
The left-hand and right-hand derivatives of a function $\mathfrak{L}(u)$ at an arbitrary point $u \in \mathbb{R}$ are defined as: $\lim _{h \rightarrow 0^{-}} \frac{\mathfrak{L}(u+h)-\mathfrak{L}(u)}{h}$ and $\lim _{h \rightarrow 0^{+}} \frac{\mathfrak{L}(u+h)-\mathfrak{L}(u)}{h}$, respectively.
First, we compute the left-hand derivative of the wave loss function $\mathfrak{L}_{wave}(u)$:
\begin{align}
L.H.D.=& \lim_{h \rightarrow 0^{-}} \frac{\mathfrak{L}_{wave}(u+h)-\mathfrak{L}_{wave}(u)}{h} \nonumber\\
=&\lim_{h \rightarrow 0^{-}} \frac{\frac{1}{\lambda}\left(1-\frac{1}{1+\lambda(u+h)^2 e^{a(u+h)}}\right)-\frac{1}{\lambda}\left(1-\frac{1}{1+\lambda u^2 e^{a u}}\right)}{h} \nonumber \\
=& \lim_{h \rightarrow 0^{-}} \frac{1}{\lambda h}\left(\frac{1}{1+\lambda u^2 e^{a u}}-\frac{1}{1+\lambda(u+h)^2 e^{a(u+h)}}\right) \nonumber \\
=& \lim_{h \rightarrow 0^{-}} \frac{(u+h)^2 e^{a(u+h)}- u^2 e^{a u}}{ h\left(1+\lambda u^2 e^{a u}\right)\left(1+\lambda(u+h)^2 e^{a(u+h)}\right)} \hspace{2cm} (\frac{0}{0}~\text{form})\nonumber\\
=& \lim_{h \rightarrow 0^{-}} \frac{a(u+h)^2 e^{a(u+h)} + 2(u+h) e^{a(u+h)}}{(1+\lambda u^2 e^{a u}) [ 1+\lambda (u+h)^2 e^{a (u+h)} + \lambda h (u+h) e^{a(u+h)} (a(u+h)+2)] } \nonumber \\
&\hspace{9.6cm}(\text{Using L'hospital rule}) \nonumber\\
=&~ \frac{u e^{a u} (au+2)}{(1+\lambda u^2 e^{a u})^2}. \nonumber
\end{align}
Now, we compute the right-hand derivative of the wave loss function $\mathfrak{L}_{wave}(u)$:
\begin{align}
R.H.D.=& \lim_{h \rightarrow 0^{+}} \frac{\mathfrak{L}_{wave}(u+h)-\mathfrak{L}_{wave}(u)}{h}. \hspace{7cm} \nonumber
\end{align}
In a similar manner, we can find that the right-hand derivative is also $\frac{u e^{a u} (au+2)}{(1+\lambda u^2 e^{a u})^2}$. Since, the left-hand derivative and right-hand derivative of the wave loss function are equal for any arbitrary $u \in \mathbb{R}$. Therefore, we have shown that the wave loss function is smooth in the whole domain.
\end{proof}
\begin{table}[htp]
\centering
\caption{Description of UCI and KEEL datasets employed in the experiments.}
\label{tab:Data Description Table}
\begin{tabular}{lccc}
\hline
Dataset & Number of samples & Number of features & I.R. \\ \hline
breast\_cancer & 286 & 9 & 2.36 \\
breast\_cancer\_wisc & 699 & 9 & 1.9 \\
breast\_cancer\_wisc\_diag & 569 & 30 & 1.68 \\
congressional\_voting & 435 & 16 & 1.59 \\
conn\_bench\_sonar\_mines\_rocks & 208 & 60 & 1.14 \\
credit\_approval & 690 & 15 & 1.25 \\
cylinder\_bands & 512 & 35 & 1.56 \\
echocardiogram & 131 & 10 & 2.05 \\
haberman\_survival & 306 & 3 & 2.78 \\
hepatitis & 155 & 19 & 3.84 \\
horse\_colic & 368 & 25 & 1.71 \\
monks\_1 & 556 & 6 & 1 \\
oocytes\_trisopterus\_nucleus\_2f & 1022 & 41 & 1.37 \\
pima & 768 & 8 & 1.86 \\
pittsburg\_bridges\_T\_OR\_D & 102 & 7 & 6.28 \\
spect & 265 & 22 & 1.41 \\
statlog\_heart & 270 & 13 & 1.25 \\
vertebral\_column\_2clases & 310 & 6 & 2.1 \\
abalone9-18 & 731 & 7 & 16.39 \\
aus & 690 & 14 & 1.25 \\
cmc & 1473 & 9 & 1.35 \\
ecoli0137vs26 & 311 & 7 & 4.76 \\
ecoli01vs5 & 240 & 7 & 11 \\
ecoli3 & 336 & 7 & 8.6 \\
led7digit-0-2-4-5-6-7-8-9\_vs\_1 & 443 & 7 & 10.98 \\
monk1 & 556 & 6 & 1.04 \\
monk3 & 556 & 6 & 1.01 \\
shuttle-6\_vs\_2-3 & 230 & 9 & 21.98 \\
shuttle-c0-vs-c4 & 1829 & 9 & 13.87 \\
sonar & 208 & 60 & 1.14 \\
transfusion & 748 & 4 & 3.2 \\
votes & 435 & 16 & 1.59 \\
vowel & 988 & 10 & 9.99 \\
yeast-0-2-5-6\_vs\_3-7-8-9 & 1004 & 8 & 9.14 \\
yeast-0-3-5-9\_vs\_7-8 & 506 & 8 & 9.12 \\
yeast1vs7 & 459 & 8 & 14.3 \\
yeast2vs8 & 483 & 8 & 23.15 \\
ringnorm & 7400 & 20 & 1.02 \\
twonorm & 7400 & 20 & 1 \\
mushroom & 8124 & 21 & 1.07 \\
musk\_2 & 6598 & 166 & 5.49 \\
EEG Eye State & 14980 & 14 & 1.23 \\ \hline
\end{tabular}
\end{table}

\section{Statistical Analysis}
To further support the efficacy of the proposed Wave-SVM and Wave-TSVM, we conducted a comprehensive statistical analysis of the models, for which we followed four tests: ranking scheme, Friedman test, Nemenyi post hoc test, and Win-Tie-Loss sign test  \cite{demvsar2006statistical}.

\textbf{Ranking scheme:} To conduct a more comprehensive evaluation of the Wave-SVM, we employed a ranking scheme. Relying solely on average accuracy as a single metric could be problematic, as exceptional performance on specific datasets might mask inadequate performance on others. To mitigate this concern, it becomes essential to individually rank each model with respect to each dataset, allowing for a comprehensive assessment of their respective capabilities. In the ranking scheme \cite{demvsar2006statistical}, the model with the poorest performance on a dataset is assigned a higher rank, while the model achieving the best performance is assigned a lower rank. Consider $p$ models are being assessed using $D$ datasets, and  $R_e^f$ denotes the rank of the $e^{th}$  of $p$ models on the $f^{th}$ of $D$ datasets. The $e^{th}$ model's mean rank is obtained by the following formula:
\begin{align}{}
R_e=\frac{1}{D} \sum_f R_e^f.
\end{align}
For SVM type models, the mean rank of the baseline models C-SVM, Pin-SVM, LINEX-SVM, and FP-SVM are $3.37$, $2.85$, $4.16$, and $2.53$, respectively. Whereas, the mean rank of the proposed Wave-SVM is  $1.8$, the most favorable position when compared to the baseline models. For TSVM type models, the mean rank of the baseline models including, TSVM, Pin-GTSVM, IF-RVFL, and SLTSVM are $2.78$, $3.55$, $3.75$, and $2.92$, respectively. In contrast, the mean rank of the proposed Wave-TSVM is $2$, the lowest in comparison to baseline models (see Table \ref{tab:non-linear Twin Table}). Since the lower rank denotes heightened accomplishment, the proposed Wave-SVM and Wave-TSVM securely occupy the foremost positions in comparison to baseline models.

\textbf{Friedman test:} 
 The Friedman test \cite{friedman1940comparison} is utilized to statistically assess the significance of various models. In this test, each individual model is ranked independently for each dataset. The highest-performing model receives a rank of $1$, followed by the second-best model with a rank of $2$, and so forth. The null hypothesis posits that all models are essentially identical, implying that the average rank of each model is equivalent. The Friedman statistic adheres to the chi-squared $\chi^2_F$ distribution with degrees of freedom (d.f.) equal to $p-1$, where $p$ represents the number of models. This statistic is computed as follows:
\begin{align}{} \label{chisquareequation}
\chi_F^2=\frac{12 D}{p(p+1)}\left[\sum_e R_e^2-\frac{p(p+1)^2}{4}\right].
\end{align}
Here, $D$ signifies the total number of datasets, and $R_e$ denotes the mean rank of the $e^{th}$ model out of the $p$ models. However, the Friedman statistic is overly cautious in nature. To address this, \citet{iman1980approximations} introduced a more robust statistic:
\begin{align}{} \label{ffequation}
F_F=\frac{(D-1) \chi_F^2}{D(p-1)-\chi_F^2},
\end{align}
which follows $F$ distribution with $((p-1),(p-1)(D-1))$ d.f.. For the proposed Wave-SVM and baseline models, we have $p=5$ and $D=42$. Thus, we obtain $\chi_F^2 = 23.95$ and $F_F= 6.82$. Also, $F_F$ is distributed with $(4,164)$ d.f.. From the statistical F-distribution table, at $5 \%$ level of significance, the value of $F_F(4,164)= 2.43$. Since, $F_F > 2.43$, thus we reject the null hypothesis. For the proposed Wave-TSVM and baseline models, we have $p=5$ and $D=32$. Thus, we obtain $\chi_F^2 = 24.58$ and $F_F= 7.36$. Also, $F_F$ is distributed with $(4,124)$ d.f.. From the statistical F-distribution table, at $5 \%$ level of significance, the value of $F_F(4,124)= 2.45$. Since, $F_F > 2.45$, thus we reject the null hypothesis. Hence, substantial differences exist among the models.

\textbf{Nemenyi post hoc test:}
In the Nemenyi post hoc test \cite{demvsar2006statistical}, a pairwise comparison is conducted among all models. The disparity in performance between the two models is considered significant if the corresponding average ranks display a noticeable difference surpassing a specific threshold value known as the critical difference ($C.D.$). When the discrepancy in mean ranks between the two models exceeds the $C.D.$, the model with the higher mean rank is deemed statistically superior to the model with the lower mean rank. The computation of $C.D.$ follows the following formula:
\begin{align}{}
    C.D.=q_\alpha \sqrt{\frac{p(p+1)}{6D}}.
\end{align}
Here, $q_\alpha$ is derived from the studentized range statistic divided by $\sqrt{2}$. This value is referred to as the critical value for the two-tailed Nemenyi test. From distribution table, for $\alpha=0.1$, the value of $q_\alpha$ is $2.29$. Thus, after calculation, we obtain $C.D.= 0.85$ and $0.97$ for SVM type models and TSVM type models, respectively. The mean rank difference of proposed Wave-SVM from C-SVM, Pin-SVM, LINEX-SVM, and FP-SVM are $1.57$, $1.05$, $2.36$, and $0.73$, respectively. Notably, Wave-SVM differs significantly from the baseline models C-SVM, Pin-SVM, and LINEX-SVM as the difference in the average rank is greater than $C.D.$, however, the rank difference of the proposed Wave-SVM with FP-SVM is less than $C.D.$. Similarly, the mean rank difference of proposed Wave-TSVM from TSVM, Pin-GTSVM, IF-RVFL, and SLTSVM are $0.78$, $1.55$, $1.75$, and $0.92$, respectively. Therefore, the proposed Wave-TSVM showcases significant disparity from the baseline models Pin-GTSVM and IF-RVFL, as the average rank difference surpasses the $C.D.$. Though, the rank difference of the proposed Wave-SVM with TSVM and SLTSVM is below the $C.D.$, the efficacy of the proposed Wave-TSVM over TSVM and SLTSVM is underscored by the average rank, average accuracy, and other statistical tests. These findings are supported by the data presented in Table \ref{tab:Nemenyi-table-UCI and KEEL for Wave-SVM} and \ref{tab:Nemenyi-table-UCI and KEEL for Wave-TSVM}, which clearly indicates the superior performance of the proposed Wave-SVM and Wave-TSVM in comparison to the baseline models.
\begin{table}[htp]
\centering
\caption{Differences in the rankings of the non-linear proposed Wave-SVM
against baseline models on UCI and KEEL datasets.}
\label{tab:Nemenyi-table-UCI and KEEL for Wave-SVM}
\begin{tabular}{|l|c|c|c|}
\hline
Model & Average rank & Rank difference & \begin{tabular}[c]{@{}c@{}}Significant difference\\ (As per Nemenyi post hoc test)\end{tabular} \\ \hline
C-SVM \cite{cortes1995support}     & 3.37 & 1.57 & Yes \\ \hline
Pin-SVM \cite{huang2013support}    & 2.85 & 1.05 & Yes  \\ \hline
LINEX-SVM \cite{ma2019linex}    & 4.16 & 2.36 & Yes \\ \hline
FP-SVM \cite{kumari2024diagnosis}   & 2.53 & 0.73 & No \\ \hline
Wave-SVM$^{\dagger}$& 1.8 & -    & N/A \\ \hline
\multicolumn{4}{l}{$^{\dagger}$ represents the proposed model.}
\end{tabular}
\end{table}

\begin{table}[htp]
\centering
\caption{Differences in the rankings of the non-linear proposed Wave-TSVM
against baseline models on UCI and KEEL datasets.}
\label{tab:Nemenyi-table-UCI and KEEL for Wave-TSVM}
\begin{tabular}{|l|c|c|c|}
\hline
Model & Average rank & Rank difference & \begin{tabular}[c]{@{}c@{}}Significant difference\\ (As per Nemenyi post hoc test)\end{tabular} \\ \hline
TSVM \cite{khemchandani2007twin}     & 2.78 & 0.78 & No \\ \hline
Pin-GTSVM \cite{tanveer2019general}   & 3.55 & 1.55 & Yes  \\ \hline
IF-RVFL \cite{malik2022alzheimer}    & 3.75 & 1.75 & Yes \\ \hline
SLTSVM \cite{si2023symmetric}   & 2.92 & 0.92 & No \\ \hline
Wave-TSVM$^{\dagger}$& 2 & -    & N/A \\ \hline
\multicolumn{4}{l}{$^{\dagger}$ represents the proposed model.}
\end{tabular}
\end{table}

\begin{table}[htp]
\centering
\caption{Non-linear Wave-SVM results of pairwise Win-Tie-Loss test on UCI and KEEL datasets. Here, in [x/y/z], x, y, and z correspond to the count of wins, ties, and losses of the row model
over column method, respectively.}
\label{tab:wintie table Wave-SVM}
\begin{tabular}{lllll}
\hline
 & C-SVM \cite{cortes1995support} & Pin-SVM \cite{huang2013support} & LINEX-SVM \cite{ma2019linex} & FP-SVM \cite{kumari2024diagnosis} \\ \hline
Pin-SVM \cite{huang2013support} & {[}13, 28, 1{]} &  &  &  \\
LINEX-SVM \cite{ma2019linex} & {[}5, 12, 25{]} & {[}4, 10, 28{]} &  &  \\
FP-SVM \cite{kumari2024diagnosis} & {[}19, 17, 6{]} & {[}14, 18, 10{]} & {[}29, 6, 7{]} &  \\
Wave-SVM$^{\dagger}$ & {[}28, 11, 3{]} & {[}25, 12, 5{]} & {[}32, 7, 3{]} & {[}27, 12, 3{]} \\ \hline
\multicolumn{5}{l}{$^{\dagger}$ represents the proposed model.}
\end{tabular}
\end{table}

\begin{table}[htp]
\centering
\caption{Non-linear Wave-TSVM results of pairwise Win-Tie-Loss test on UCI and KEEL datasets. Here, in [x/y/z], x, y, and z correspond to the count of wins, ties, and losses of the row model
over column method, respectively.}
\label{tab:wintie table Wave-TSVM}
\begin{tabular}{lllll}
\hline
 & TSVM \cite{khemchandani2007twin} & Pin-GTSVM \cite{tanveer2019general} & IF-RVFL \cite{malik2022alzheimer} & SLTSVM \cite{si2023symmetric}\\ \hline
Pin-GTSVM \cite{tanveer2019general} & {[}10, 4, 18{]} &  &  &  \\
IF-RVFL \cite{malik2022alzheimer} & {[}6, 5, 21{]} & {[}15, 2, 15{]} &  &  \\
SLTSVM \cite{si2023symmetric} & {[}11, 5, 16{]} & {[}21, 2, 9{]} & {[}19, 4, 9{]} &  \\
Wave-TSVM$^{\dagger}$ & {[}19, 8, 5{]} & {[}22, 3, 7{]} & {[}26, 3, 3{]} & {[}17, 10, 5{]} \\ \hline
\multicolumn{5}{l}{$^{\dagger}$ represents the proposed model.}
\end{tabular}
\end{table}

\textbf{Win-Tie-Loss sign test:} It is a well-known pairwise statistical test to evaluate whether there is a statistical difference between the outcomes of two models. In this test, it is assumed that the two models are equivalent under the null hypothesis, i.e., each model wins on $D/2$ out of $D$ datasets. Two models are considered significantly different at 5\% level of significance if one model has at least $D/2 + 1.96 \sqrt{D}/2$ wins. If there is a tie, the score is shared equally among the two models. For Wave-SVM, we have $D=42$, $D/2 + 1.96 \sqrt{D}/2= 27.35$. Thus, two SVM type models are statistically different, if one of them wins on at least $27$ datasets.  Table \ref{tab:wintie table Wave-SVM} presents the pairwise Win-Tie-Loss performance of the proposed Wave-SVM and the baseline models. It has been determined that the proposed Wave-SVM has $28$ (\emph{w.r.t}. C-SVM), $25$ (\emph{w.r.t}. Pin-SVM), $32$ (\emph{w.r.t}. LINEX-SVM), and $27$ (\emph{w.r.t}. FP-SVM) wins out of $42$ datasets. For Wave-TSVM, we have $D=32$, $D/2 + 1.96 \sqrt{D}/2= 21.54$. Thus, two models are statistically different, if one of them wins on at least $21$ datasets. Table \ref{tab:wintie table Wave-TSVM} showcases the Win-Tie-Loss results of the proposed Wave-TSVM and the baseline models. It can be seen that the proposed Wave-TSVM has $19$ (\emph{w.r.t}. TSVM), $22$ (\emph{w.r.t}. Pin-GTSVM), $26$ (\emph{w.r.t}. IF-RVFL), and $17$ (\emph{w.r.t}. SLTSVM) wins out of $32$ datasets. After sharing the ties equally among the models, it can be established that the proposed Wave-SVM and Wave-TSVM models perform superiorly over the baseline models.
\begin{table}[htp]
\centering
\caption{Accuracy and training time for linear Wave-SVM against baseline models on UCI and KEEL datasets.}
\label{tab:detailed-linear-Wave-SVM-Table}

\resizebox{\textwidth}{!}{%
\begin{tabular}{lcccccccccc}
\hline
Dataset & \multicolumn{2}{c}{C-SVM \cite{cortes1995support}} & \multicolumn{2}{c}{Pin-SVM \cite{huang2013support}} & \multicolumn{2}{c}{LINEX-SVM \cite{ma2019linex}} & \multicolumn{2}{c}{FP-SVM \cite{kumari2024diagnosis}} & \multicolumn{2}{c}{Wave-SVM$^{\dagger}$} \\
 & Accuracy & Time & Accuracy & Time & Accuracy & Time & Accuracy & Time & Accuracy & Time \\ \hline
breast\_cancer & 100 & 0.004 & 100 & 0.543 & 94.44 & 0.0059 & 100 & 0.3012 & 100 & 0.0059 \\
breast\_cancer\_wisc & 95.43 & 0.0351 & 100 & 0.1989 & 99.43 & 0.003 & 99.43 & 0.0759 & 99.43 & 0.003 \\
breast\_cancer\_wisc\_diag & 96.5 & 0.1831 & 99.3 & 0.1108 & 93.66 & 0.0044 & 98.59 & 1.316 & 99.3 & 0.0044 \\
congressional\_voting & 64.22 & 0.011 & 64.22 & 0.0702 & 58.72 & 0.0032 & 64.22 & 0.2424 & 66.97 & 0.0032 \\
conn\_bench\_sonar\_mines\_rocks & 61.54 & 0.018 & 100 & 0.0168 & 90.38 & 0.0075 & 100 & 0.0717 & 100 & 0.0075 \\
credit\_approval & 69.94 & 0.0407 & 99.42 & 0.196 & 63.01 & 0.0013 & 99.42 & 1.8022 & 99.42 & 0.0013 \\
cylinder\_bands & 67.97 & 0.0127 & 73.44 & 0.0735 & 72.66 & 0.0043 & 74.22 & 0.7981 & 75.78 & 0.0043 \\
echocardiogram & 81.82 & 0.002 & 93.75 & 0.0094 & 93.75 & 0.0035 & 93.75 & 0.7435 & 96.97 & 0.0035 \\
haberman\_survival & 80.52 & 0.0081 & 80.52 & 0.0429 & 83.12 & 0.005 & 81.82 & 12.139 & 84.42 & 0.005 \\
hepatitis & 94.87 & 0.0038 & 94.87 & 0.0103 & 69.23 & 0.0057 & 94.87 & 0.1911 & 97.44 & 0.0057 \\
horse\_colic & 82.61 & 0.0087 & 92.39 & 0.0423 & 51.09 & 0.0021 & 88.04 & 0.4263 & 90.22 & 0.0021 \\
monks\_1 & 50.36 & 0.0174 & 63.31 & 0.1267 & 67.63 & 0.0013 & 74.1 & 1.0576 & 87.05 & 0.0013 \\
oocytes\_trisopterus\_nucleus\_2f & 74.56 & 0.3941 & 76.75 & 0.5174 & 67.98 & 0.0014 & 76.75 & 1.5174 & 78.95 & 0.0014 \\
pima & 75.52 & 0.0358 & 81.25 & 0.2001 & 70.31 & 0.0011 & 79.69 & 2.2016 & 82.29 & 0.0011 \\
pittsburg\_bridges\_T\_OR\_D & 92.31 & 0.0019 & 96.15 & 0.0028 & 92.31 & 0.0009 & 96 & 0.0105 & 100 & 0.0009 \\
spect & 71.64 & 0.0053 & 77.27 & 0.0264 & 65.67 & 0.0017 & 78.79 & 0.2066 & 81.82 & 0.0017 \\
statlog\_heart & 80.88 & 0.0056 & 89.55 & 0.0232 & 77.94 & 0.0015 & 88.06 & 0.021 & 92.54 & 0.0015 \\
vertebral\_column\_2clases & 100 & 0.219 & 100 & 0.0274 & 93.51 & 0.0009 & 100 & 0.1121 & 100 & 0.0009 \\
abalone9-18 & 91.8 & 0.0955 & 100 & 0.2191 & 100 & 0.0015 & 100 & 0.5245 & 100 & 0.0129 \\
aus & 87.86 & 0.0731 & 91.28 & 0.0926 & 82.56 & 0.0022 & 90.7 & 1.7916 & 92.44 & 0.0025 \\
cmc & 10.3 & 0.3185 & 100 & 0.4872 & 100 & 0.0092 & 100 & 0.4161 & 100 & 0.0036 \\
ecoli0137vs26 & 93.59 & 0.0243 & 97.44 & 0.0254 & 84.62 & 0.0017 & 93.59 & 0.3243 & 96.15 & 0.0014 \\
ecoli01vs5 & 100 & 0.0122 & 100 & 0.0246 & 95 & 0.0022 & 100 & 0.4122 & 100 & 0.0021 \\
ecoli3 & 91.67 & 0.0363 & 91.67 & 0.0422 & 91.67 & 0.0018 & 91.67 & 0.3363 & 92.86 & 0.0019 \\
led7digit-0-2-4-5-6-7-8-9\_vs\_1 & 93.69 & 0.0183 & 97.3 & 0.0452 & 67.27 & 0.0024 & 98.2 & 0.415 & 96.4 & 0.0037 \\
monk1 & 53.96 & 0.0179 & 53.96 & 0.0802 & 55.4 & 0.0019 & 56.12 & 0.6479 & 58.99 & 0.0036 \\
monk3 & 51.8 & 0.0253 & 51.08 & 0.0677 & 56.12 & 0.002 & 55.4 & 13.2818 & 59.71 & 0.0022 \\
shuttle-6\_vs\_2-3 & 100 & 0.0117 & 100 & 0.0195 & 82.76 & 0.0021 & 100 & 0.0081 & 100 & 0.0072 \\
shuttle-c0-vs-c4 & 100 & 0.8325 & 100 & 1.1571 & 92.58 & 0.0019 & 100 & 0.8325 & 100 & 0.0026 \\
sonar & 78.85 & 0.0095 & 84.62 & 0.0154 & 61.54 & 0.0025 & 78.85 & 0.0095 & 86.54 & 0.0059 \\
transfusion & 59.36 & 0.0469 & 87.17 & 0.2267 & 40.64 & 0.0022 & 87.7 & 10.1106 & 87.17 & 0.0012 \\
votes & 96.33 & 0.0245 & 97.25 & 0.032 & 77.06 & 0.0022 & 97.25 & 0.0858 & 97.25 & 0.0026 \\
vowel & 95.55 & 0.1324 & 96.76 & 0.266 & 91.09 & 0.0019 & 95.55 & 0.1324 & 96.76 & 0.0026 \\
yeast-0-2-5-6\_vs\_3-7-8-9 & 94.02 & 0.1471 & 94.02 & 0.2317 & 93.23 & 0.0028 & 96.41 & 21.9554 & 94.42 & 0.002 \\
yeast-0-3-5-9\_vs\_7-8 & 92.13 & 0.0298 & 95.24 & 0.0416 & 91.34 & 0.0019 & 96.83 & 6.0563 & 95.24 & 0.0021 \\
yeast1vs7 & 94.78 & 0.0428 & 94.78 & 0.0415 & 94.74 & 0.0031 & 94.78 & 1.428 & 94.78 & 0.0033 \\
yeast2vs8 & 98.35 & 0.1581 & 98.35 & 0.0513 & 47.11 & 0.0039 & 99.17 & 1.3929 & 99.17 & 0.0056 \\
ringnorm & 77.51 & 18.7724 & 78.05 & 22.3657 & 78.16 & 0.00147 & \multicolumn{1}{l}{\hspace{0.7cm}*} & \multicolumn{1}{l}{\hspace{0.7cm}*} & 79.09 & 0.0105 \\
twonorm & 98.1 & 13.1825 & 98.81 & 20.0629 & 34.59 & 0.0057 & \multicolumn{1}{l}{\hspace{0.7cm}*} & \multicolumn{1}{l}{\hspace{0.7cm}*} & 98.92 & 0.0089 \\
mushroom & 94.69 & 17.1168 & 95.27 & 66.9571 & 79.02 & 0.0027 & \multicolumn{1}{l}{\hspace{0.7cm}*} & \multicolumn{1}{l}{\hspace{0.7cm}*} & 99.21 & 0.0044 \\
musk\_2 & 99.15 & 12.3079 & 100 & 17.5185 & 93.21 & 0.0207 & \multicolumn{1}{l}{\hspace{0.7cm}*} & \multicolumn{1}{l}{\hspace{0.7cm}*} & 98.24 & 0.0513 \\
EEG Eye State & 58.83 & 129.8739 & 56.98 & 164.8006 & 56.82 & 0.0051 & \multicolumn{1}{l}{\hspace{0.7cm}*} & \multicolumn{1}{l}{\hspace{0.7cm}*} & 72.55 & 0.0089 \\ \hline
\textbf{Avg. Acc. and Avg. time} & 82.21 & 4.6266 & 89.1 & 7.074 & 77.41 & \textbf{0.0033} & 89.73 & 2.254 & \textbf{91.15} & 0.0049 \\ \hline
\textbf{Avg. Rank} & 3.73 & \multicolumn{1}{l}{} & 2.52 & \multicolumn{1}{l}{} & 4.29 & \multicolumn{1}{l}{} & 2.49 & \multicolumn{1}{l}{} & \textbf{1.68} & \multicolumn{1}{l}{} \\ \hline
\multicolumn{11}{l}{$^{\dagger}$ represents the proposed model.}\\
\multicolumn{11}{l}{* denotes that we stopped the implementation after 36 hours.}\\
\multicolumn{11}{l}{Here, Avg. and Acc. are acronyms used for average and accuracy, respectively.}
\end{tabular}}
\end{table}
\begin{table}[htp]
\centering
\caption{The optimal parameters of the proposed linear Wave-SVM and baseline models corresponding to accuracy values.}
\label{tab:linear-optimal-parameters-table}

\resizebox{\textwidth}{!}{%
\begin{tabular}{lccccc}
\hline
Dataset & C-SVM\cite{cortes1995support} & Pin-SVM\cite{huang2013support} & LINEX-SVM\cite{ma2019linex} & FP-SVM \cite{kumari2024diagnosis} & Wave-SVM$^{\dagger}$ \\
 & ($C$) & ($\tau$, $C$) & ($a$, $C$) & ($C$, $\tau_1$, $\tau_2$) & ($\lambda$, $a$, $C$)  \\ \hline
breast\_cancer & 0.000001 & 0,0.001 & -6,1 & 0.000001, 0.000001, -1 & 0.5,-1.9,0.0001 \\
breast\_cancer\_wisc & 0.1 & 0,0.001 & -10,0.000001 & 0.000001, 0.01, 0.1 & 0.1,-2,0.000001 \\
breast\_cancer\_wisc\_diag & 0.1 & 0,0.1 & -10,0.000001 & 0.000001, 1, 0 & 1.1,-1.9,0.001 \\
congressional\_voting & 0.01 & 0,0.01 & -9,1 & 0.01, 1, 0 & 1.1,2.9,0.1 \\
conn\_bench\_sonar\_mines\_rocks & 1 & 0.5,0.1 & -5,0.01 & 0.000001, 0.000001, 0 & 1.3,-1.3,1000 \\
credit\_approval & 0.01 & 0,0.01 & -10,0.000001 & 0.0001, 0.01, 0 & 0.3,-0.3,0.001 \\
cylinder\_bands & 0.000001 & 0.3,0.1 & -1,01,00,000 & 100, 1, 0.1 & 1.5,-1.8,10000 \\
echocardiogram & 0.01 & 0,0.01 & -10,0.000001 & 0.000001, 1, 0 & 0.5,-0.4,0.001 \\
haberman\_survival & 0.01 & 0,0.01 & -5,0.0001 & 0.000001, 0.01, 0 & 1.7,-1.1,0.1 \\
hepatitis & 10 & 0,10 & -10,0.000001 & 0.000001, 0.000001, -1 & 0.5,-1.7,1000 \\
horse\_colic & 0.01 & 0.3,0.01 & -9,1 & 100, 1, 0.1 & 1.1,1.1,0.00001 \\
monks\_1 & 0.001 & 0,0.001 & -3,1 & 1000000, 1, 0 & 1.9,3.6,0.01 \\
oocytes\_trisopterus\_nucleus\_2f & 1000 & 0.5,0.01 & -9, 1000000 & 0.000001, 0.000001, -0.7 & 1.5,4.6,0.00001 \\
pima & 0.01 & 0.9,0.01 & -10,0.000001 & 1, 1, 0.1 & 0.9,1,0.01 \\
pittsburg\_bridges\_T\_OR\_D & 0.000001 & 0,0.1 & -6,0.0001 & 1000000, 1, 0.5 & 1.3,-0.7,0.0001 \\
spect & 0.001 & 0.2,1 & -6,0.0001 & 100, 0.01, 0 & 0.3,-2,0.01 \\
statlog\_heart & 0.001 & 0.5,0.01 & -8, 1000000 & 10000, 1, 0 & 1.7,2.1,0.0001 \\
vertebral\_column\_2clases & 0.000001 & 0.9,0.1 & -8, 1000000 & 0.000001, 0.000001, -0.7 & 1.5,-2,0.001 \\
abalone9-18 & 0.1 & 0,0.001 & -10,0.000001 & 0.000001, 0.000001, -1 & 0.1,-2,0.000001 \\
aus & 0.01 & 0.5,0.001 & -8,0.00001 & 0.000001, 1, 0.5 & 1.9,-0.2,1 \\
cmc & 0.1 & 0.2,0.001 & -10,0.00001 & 0.000001, 0.000001, 0 & 0.1,-1.4,10 \\
ecoli0137vs26 & 1 & 0,1 & -10,0.000001 & 0.000001, 0.000001, -0.7 & 1.1,2.3,0.000001 \\
ecoli01vs5 & 1 & 0,1 & -10,0.000001 & 100, 0.01, 0 & 1.5,4.2,0.00001 \\
ecoli3 & 10 & 0,0.001 & -10,0.000001 & 0.000001, 1, 0.5 & 1.1,4.5,0.00001 \\
led7digit-0-2-4-5-6-7-8-9\_vs\_1 & 0.000001 & 0,0.1 & -10,0.000001 & 100, 1, 0 & 0.9,3,0.000001 \\
monk1 & 0.000001 & 0,0.001 & -7, 100000 & 10000, 100, 0.1 & 0.3,1.2,0.000001 \\
monk3 & 0.01 & 0,0.001 & -7, 1000 & 10000, 0.01, 0.5 & 0.5,1,10 \\
shuttle-6\_vs\_2-3 & 0.1 & 0,0.001 & -10,0.0001 & 0.000001, 0.000001, 0.5 & 0.1,-2,0.000001 \\
shuttle-c0-vs-c4 & 0.001 & 0,0.001 & -9, 10000 & 0.000001, 1, 0 & 1.5,0.6,0.000001 \\
sonar & 0.1 & 0,0.1 & -10,0.01 & 0.000001, 1, 0 & 0.3,-0.8,1000000 \\
transfusion & 0.000001 & 0,0.001 & -10,0.00001 & 0.000001, 0.01, 0 & 0.1,-2,0.000001 \\
votes & 0.1 & 0.2,0.01 & -10,0.000001 & 0.000001, 1, 0.5 & 1.9,-1.9,0.1 \\
vowel & 0.01 & 0,0.01 & -8,0.000001 & 0.000001, 1, 0.5 & 1.7,-1.8,10000 \\
yeast-0-2-5-6\_vs\_3-7-8-9 & 0.01 & 0,1 & -10,0.000001 & 1, 1, 0 & 1.5,1.2,0.000001 \\
yeast-0-3-5-9\_vs\_7-8 & 0.01 & 0.9,0.1 & -10,0.000001 & 0.000001, 0.000001, 0.5 & 1.9,0.9,0.000001 \\
yeast1vs7 & 0.000001 & 0,0.001 & -10,0.000001 & 0.000001, 1, 0 & 0.1,-2,0.000001 \\
yeast2vs8 & 0.000001 & 0,0.001 & -10,0.000001 & 0.000001, 0.01, 0 & 1.9,1.2,0.000001 \\
ringnorm & 0.001 & 0,0.01 & -10,0.001 & \multicolumn{1}{l}{\hspace{2cm}$\_\_\_$} & 1.5,4.3,0.1 \\
twonorm & 0.0001 & 0,0.00001 & -10,0.000001 & \multicolumn{1}{l}{\hspace{2cm}$\_\_\_$} & 0.5,-0.6,0.01 \\
mushroom & 10 & 0,0.000001 & -10,10 & \multicolumn{1}{l}{\hspace{2cm}$\_\_\_$} & 1.7,0.7,100 \\
musk\_2 & 0.001 & 0.2,100 & -10,0.000001 & \multicolumn{1}{l}{\hspace{2cm}$\_\_\_$} & 0.1,0.3,0.01 \\
EEG Eye State & 1 & 0, 1000 & -1,0.000001 & \multicolumn{1}{l}{\hspace{2cm}$\_\_\_$} & 1.3,0,0.001 \\ \hline
\multicolumn{6}{l}{$^{\dagger}$ represents the proposed model.}
\end{tabular}}
\end{table}
\begin{table}[htp]
\centering
\caption{Accuracy and training time for non-linear Wave-SVM against baseline models using Gaussian kernel function on UCI and KEEL datasets.}
\label{tab:detailed-non-linear Wave-SVM table}

\resizebox{\textwidth}{!}{%
\begin{tabular}{lcccccccccc}
\hline
Dataset & \multicolumn{2}{c}{C-SVM \cite{cortes1995support}} & \multicolumn{2}{c}{Pin-SVM \cite{huang2013support}} & \multicolumn{2}{c}{LINEX-SVM \cite{ma2019linex}} & \multicolumn{2}{c}{FP-SVM \cite{kumari2024diagnosis}} & \multicolumn{2}{c}{Wave-SVM$^{\dagger}$} \\
 & Accuracy & Time & Accuracy & Time & Accuracy & Time & Accuracy & Time & Accuracy & Time \\ \hline
breast\_cancer & 100 & 0.2574 & 100 & 0.4802 & 100 & 0.0108 & 100 & 0.137 & 100 & 0.0069 \\
breast\_cancer\_wisc & 100 & 0.0644 & 100 & 0.1535 & 98.28 & 0.0038 & 100 & 25.4471 & 100 & 0.0031 \\
breast\_cancer\_wisc\_diag & 97.89 & 0.0633 & 97.89 & 0.0664 & 96.48 & 0.0049 & 98.59 & 0.0246 & 98.59 & 0.0082 \\
congressional\_voting & 65.14 & 0.0281 & 65.14 & 0.033 & 63.3 & 0.0044 & 65.74 & 0.0134 & 67.89 & 0.0078 \\
conn\_bench\_sonar\_mines\_rocks & 100 & 0.0206 & 100 & 0.0173 & 100 & 0.0053 & 100 & 0.0052 & 100 & 0.0033 \\
credit\_approval & 95.38 & 0.138 & 96.53 & 0.1659 & 90.75 & 0.005 & 95.38 & 0.0404 & 95.95 & 0.0108 \\
cylinder\_bands & 71.09 & 0.0305 & 71.09 & 0.0478 & 71.09 & 0.0046 & 74.22 & 0.0315 & 76.56 & 0.015 \\
echocardiogram & 84.38 & 0.0036 & 87.5 & 0.006 & 84.38 & 0.0034 & 93.75 & 0.0019 & 96.97 & 0.0072 \\
haberman\_survival & 80.52 & 0.018 & 81.82 & 0.0475 & 79.22 & 0.0054 & 80.52 & 0.0077 & 85.71 & 0.0045 \\
hepatitis & 92.31 & 0.0142 & 92.31 & 0.0052 & 94.87 & 0.0041 & 92.31 & 0.0144 & 97.44 & 0.007 \\
horse\_colic & 76.09 & 0.0201 & 83.7 & 0.0314 & 75 & 0.0048 & 85.87 & 2.0631 & 85.87 & 0.0039 \\
monks\_1 & 72.66 & 0.0791 & 74.1 & 0.0642 & 68.35 & 0.0046 & 79.14 & 0.0545 & 85.61 & 0.0066 \\
oocytes\_trisopterus\_nucleus\_2f & 72.63 & 0.2335 & 72.63 & 0.2542 & 67.54 & 0.0051 & 72.63 & 0.3335 & 75 & 0.0062 \\
pima & 79.69 & 0.0822 & 80.21 & 0.1601 & 73.44 & 0.0045 & 79.69 & 0.0632 & 80.73 & 0.0083 \\
pittsburg\_bridges\_T\_OR\_D & 96 & 0.0047 & 96 & 0.005 & 96 & 0.0046 & 96.15 & 0.4534 & 100 & 0.0016 \\
spect & 74.24 & 0.0162 & 75.76 & 0.0167 & 72.73 & 0.0042 & 77.27 & 0.0049 & 86.36 & 0.0062 \\
statlog\_heart & 86.57 & 0.021 & 86.57 & 0.0164 & 83.58 & 0.0041 & 88.06 & 0.0074 & 89.71 & 0.0043 \\
vertebral\_column\_2clases & 100 & 0.0139 & 100 & 0.0143 & 100 & 0.0045 & 100 & 0.0399 & 100 & 0.0071 \\
abalone9-18 & 100 & 0.0747 & 100 & 0.1922 & 100 & 0.0054 & 100 & 0.3086 & 100 & 0.0076 \\
aus & 84.97 & 0.1197 & 89.53 & 0.14 & 87.21 & 0.0049 & 88.37 & 19.5134 & 90.12 & 0.0085 \\
cmc & 99.73 & 2.3671 & 100 & 1.5839 & 87.5 & 0.0099 & 100 & 0.1654 & 100 & 0.0118 \\
ecoli0137vs26 & 98.7 & 0.0172 & 98.72 & 0.0199 & 84.62 & 0.0044 & 98.72 & 0.0199 & 98.72 & 0.0104 \\
ecoli01vs5 & 100 & 0.0097 & 100 & 0.0173 & 95 & 0.004 & 100 & 0.0173 & 100 & 0.0083 \\
ecoli3 & 95.24 & 0.0163 & 96.43 & 0.0447 & 91.67 & 0.0043 & 96.43 & 0.0447 & 100 & 0.0088 \\
led7digit-0-2-4-5-6-7-8-9\_vs\_1 & 98.2 & 0.0435 & 98.2 & 0.0311 & 96.4 & 0.0046 & 98.2 & 0.024 & 98.2 & 0.0073 \\
monk1 & 54.68 & 0.0892 & 54.68 & 0.1369 & 58.99 & 0.0048 & 56.12 & 0.0886 & 65.47 & 0.0073 \\
monk3 & 52.52 & 0.0785 & 51.08 & 0.0678 & 57.55 & 0.0045 & 57.55 & 1.2189 & 64.03 & 0.0095 \\
shuttle-6\_vs\_2-3 & 100 & 0.0132 & 100 & 0.0099 & 100 & 0.0047 & 100 & 0.0231 & 100 & 0.0095 \\
shuttle-c0-vs-c4 & 99.78 & 0.5597 & 100 & 0.5717 & 99.78 & 0.0053 & 100 & 0.2061 & 99.78 & 0.0042 \\
sonar & 92.31 & 0.0078 & 92.31 & 0.0215 & 75 & 0.0049 & 92.31 & 0.0043 & 88.46 & 0.0078 \\
transfusion & 87.17 & 0.0562 & 87.17 & 0.1295 & 87.17 & 0.0038 & 87.7 & 0.034 & 87.17 & 0.0063 \\
votes & 96.33 & 0.0285 & 96.33 & 0.0274 & 94.5 & 0.0053 & 96.17 & 0.1345 & 96.33 & 0.0067 \\
vowel & 97.98 & 0.3369 & 97.98 & 0.3629 & 95.55 & 0.0135 & 97.98 & 0.9215 & 100 & 0.007 \\
yeast-0-2-5-6\_vs\_3-7-8-9 & 95.22 & 0.2745 & 95.22 & 0.4243 & 93.23 & 0.0047 & 96.02 & 0.0356 & 97.21 & 0.0065 \\
yeast-0-3-5-9\_vs\_7-8 & 95.24 & 0.0448 & 95.24 & 0.094 & 92.86 & 0.004 & 95.24 & 0.1141 & 97.62 & 0.0036 \\
yeast1vs7 & 97.39 & 0.0557 & 97.39 & 0.0553 & 94.78 & 0.0052 & 97.39 & 0.0553 & 98.26 & 0.0038 \\
yeast2vs8 & 98.35 & 0.0187 & 98.35 & 0.0469 & 98.35 & 0.0058 & 99.17 & 0.0141 & 100 & 0.0145 \\
ringnorm & 100 & 12.801 & 100 & 11.9823 & 97.44 & 0.0891 & \multicolumn{1}{l}{\hspace{0.7cm}*} & \multicolumn{1}{l}{\hspace{0.7cm}*} & 79.51 & 0.1245 \\
twonorm & 100 & 4.9025 & 100 & 6.4691 & 100 & 0.189 & \multicolumn{1}{l}{\hspace{0.7cm}*} & \multicolumn{1}{l}{\hspace{0.7cm}*} & 98.65 & 0.303 \\
mushroom & 97.89 & 29.4162 & 98.05 & 15.2824 & 60.27 & 0.0284 & \multicolumn{1}{l}{\hspace{0.7cm}*} & \multicolumn{1}{l}{\hspace{0.7cm}*} & 100 & 0.0614 \\
musk\_2 & 98.54 & 15.9498 & 98.54 & 5.851 & 95.95 & 0.0409 & \multicolumn{1}{l}{\hspace{0.7cm}*} & \multicolumn{1}{l}{\hspace{0.7cm}*} & 100 & 0.0617 \\
EEG Eye State & 72.52 & 249.8753 & 72.52 & 317.6164 & 79.87 & 0.0344 & \multicolumn{1}{l}{\hspace{0.7cm}*} & \multicolumn{1}{l}{\hspace{0.7cm}*} & 78.8 & 0.1007 \\ \hline
\textbf{Avg. Acc. and Avg. time} & 89.46 & 7.57 & 89.97 & 8.64 & 86.63 & \textbf{0.0137} & 90.18 & 1.39 & \textbf{91.92} & 0.0219 \\ \hline
\textbf{Avg. Rank} & 3.37 & \multicolumn{1}{l}{} & 2.85 & \multicolumn{1}{l}{} & 4.16 & \multicolumn{1}{l}{} & 2.53 & \multicolumn{1}{l}{} & \textbf{1.8} & \multicolumn{1}{l}{} \\ \hline
\multicolumn{6}{l}{$^{\dagger}$ represents the proposed model.}\\
\multicolumn{6}{l}{Here, Avg. and Acc. are acronyms used for average and accuracy, respectively.}
\end{tabular}}
\end{table}
\begin{table}[htp]
\centering
\caption{The optimal parameters of the proposed non-linear Wave-SVM and baseline models corresponding to accuracy values.}
\label{tab:non-linear-optimal-parameters-table}

\resizebox{14.5cm}{!}{
\begin{tabular}{lccccc}
\hline
Dataset & C-SVM \cite{cortes1995support} & Pin-SVM \cite{huang2013support} & LINEX-SVM \cite{ma2019linex} & FP-SVM \cite{kumari2024diagnosis} & Wave-SVM$^{\dagger}$ \\
 & ($C$, $\sigma$) & ($\tau$, $C$, $\sigma$) & ($a$, $C$, $\sigma$) & ($C$, $\sigma$, $\tau_1$, $\tau_2$) & ($\lambda$, $a$, $C$, $\sigma$) \\ \hline
breast\_cancer & 0.000001,0.000001 & 0,0.000001,0.000001 & -10,0.001,0.001 & 0.000001, 0.000001, -1, 1 & 0.1,-2,0.000001,0.000001 \\
breast\_cancer\_wisc & 10,0.01 & 0.2,0.01,0.1 & -1,0.001,1 & 0.000001, 0.01, 0.1, 0.1 & 1.3,-2,0.000001,0.1 \\
breast\_cancer\_wisc\_diag & 1,0.1 & 0,0.1,0.1 & -1,0.01,1 & 0.000001, 1, 0, 0.1 & 1.9,-1.6,0.00001,1 \\
congressional\_voting & 1,1 & 0,1,0.01 & -10,0.001,0.001 & 0.01, 1, 0, 0.1 & 1.3,-1.3,0.000001,1 \\
conn\_bench\_sonar\_mines\_rocks & 0.1,10 & 0,0.1,1 & -10,0.001,0.001 & 0.000001, 0.000001, 0, 0.8 & 0.1,-2,0.000001,0.000001 \\
credit\_approval & 0.1,0.1 & 0,0.1,0.01 & -1,0.01,1 & 0.0001, 0.01, 0, 0.5 & 0.7,4.7,0.000001,0.1 \\
cylinder\_bands & 0.000001,0.000001 & 0,0.000001,0.000001 & -10,0.001,0.001 & 100, 1, 0.1, 0 & 1.9,-0.1,0.00001,1 \\
echocardiogram & 0.000001,0.000001 & 0,1,0.01 & -1,0.001,0.001 & 0.000001, 1, 0, 0.1 & 0.3,-0.4,0.000001,1 \\
haberman\_survival & 1,1 & 0.4,1000000,0.01 & -1,0.001,0.001 & 0.000001, 0.01, 0, 0.1 & 1.1,2.9,0.001,10 \\
hepatitis & 0.000001,0.000001 & 0,0.000001,10 & -1,0.001,1 & 0.000001, 0.000001, -1, 1 & 1.9,-0.1,0.00001,1 \\
horse\_colic & 10,1 & 0,0.1,0.01 & -2,0.001,1 & 100, 1, 0.1, 0.1 & 1.3,-1.6,0.00001,1 \\
monks\_1 & 10,1 & 0,0.001,0.001 & -5,1000,10 & 1000000, 1, 0, 0.9 & 0.9,0.4,0.000001,10 \\
oocytes\_trisopterus\_nucleus\_2f & 10,1 & 0,1,1000 & -1,1,1 & 0.000001, 0.000001, -0.7, 0.7 & 0.9,3.1,0.00001,1 \\
pima & 0.1,1 & 0,1,0.01 & -1,0.001,1 & 1, 1, 0.1, 0 & 0.1,4.1,0.000001,10 \\
pittsburg\_bridges\_T\_OR\_D & 1,1 & 0,0.000001,0.000001 & -2,0.001,0.001 & 1000000, 1, 0.5, 0.5 & 1.5,0.8,0.00001,0.1 \\
spect & 0.000001,0.000001 & 0,1,0.001 & -1,0.001,0.1 & 100, 0.01, 0, 0.3 & 1.5,4.9,0.00001,1 \\
statlog\_heart & 1,0.1 & 0.7,0.1,0.001 & -1,0.001,1 & 10000, 1, 0, 0.9 & 1.3,-0.8,0.000001,1 \\
vertebral\_column\_2clases & 1,0.1 & 0,0.000001,0.000001 & -10,0.001,0.001 & 0.000001, 0.000001, -0.7, 0.7 & 0.1,-2,0.000001,0.000001 \\
abalone9-18 & 0.000001,0.000001 & 0,0.000001,0.1 & -2,0.001,0.001 & 0.000001, 0.000001, -1, 1 & 1.3,0.4,0.000001,0.1 \\
aus & 0.000001,0.000001 & 0,1,0.01 & -1,0.001,1 & 0.000001, 1, 0.5, 0 & 0.9,-1.3,0.000001,1 \\
cmc & 100,1 & 0.7,0.01,0.1 & -10,0.001,0.001 & 0.000001, 0.000001, 0, 0.8 & 1.4,-2,0.0001,10 \\
ecoli0137vs26 & 1,00,100 & 0,1,1 & -2,0.001,0.001 & 0.000001, 0.000001, -0.7, 0.7 & 0.1,-2,0.0001,10 \\
ecoli01vs5 & 1,1 & 0,1,1 & -2,0.001,0.001 & 100, 0.01, 0, 0.3 & 0.5,-2,0.000001,0.1 \\
ecoli3 & 1000, 10000 & 0,1,10 & -2,0.001,0.001 & 0.000001, 1, 0.5, 0 & 0.9,-1.8,0.00001,10 \\
led7digit-0-2-4-5-6-7-8-9\_vs\_1 & 1,10 & 0,1,0.000001 & -1,0.001,1 & 100, 1, 0, 0.9 & 0.1,-2,10,10 \\
monk1 & 1,1 & 0,10000,0.000001 & -5,0.001,10 & 10000, 100, 0.1, 0 & 1.7,2.8,0.0001,10 \\
monk3 & 10,1 & 0,0.000001,0.01 & -10,1,10 & 10000, 0.01, 0.5, 0 & 0.7,4.7,1000,10 \\
shuttle-6\_vs\_2-3 & 100000, 1000000 & 0,10,0.1 & -2,0.001,0.01 & 0.000001, 0.000001, 0.5, 0 & 0.3,-2,0.000001,0.01 \\
shuttle-c0-vs-c4 & 1,0.001 & 0,0.1,0.001 & -1,0.001,0.1 & 0.000001, 1, 0, 0.2 & 0.7,-2,0.000001,0.1 \\
sonar & 10,0.1 & 0,10,0.1 & -1,0.001,10 & 0.000001, 1, 0, 0.1 & 0.1,-0.8,10000,10 \\
transfusion & 10,1 & 0,0.000001,0.000001 & -1,0.001,0.001 & 0.000001, 0.01, 0, 0.2 & 0.1,-2,0.000001,0.000001 \\
votes & 0.000001,0.000001 & 0,1,0.1 & -1,0.001,1 & 0.000001, 1, 0.5, 0 & 0.9,0.3,0.000001,10 \\
vowel & 1,0.1 & 0,1,0.01 & -1,0.001,1 & 0.000001, 1, 0.5, 0 & 1.5,-2,0.0001,0.1 \\
yeast-0-2-5-6\_vs\_3-7-8-9 & 1,0.1 & 0,10,0.01 & -2,0.001,0.001 & 1, 1, 0, 1 & 0.3,-1.9,1000,10 \\
yeast-0-3-5-9\_vs\_7-8 & 10,1 & 0,1,0.01 & -2,0.001,0.001 & 0.000001, 0.000001, 0.5, 0 & 0.1,-0.2,100000,10 \\
yeast1vs7 & 1,1 & 0,10000,0.000001 & -2,0.001,0.001 & 0.000001, 1, 0, 0.2 & 1.5,1.7,100,10 \\
yeast2vs8 & 10000,1 & 0,0.000001,0.000001 & -2,0.001,0.001 & 0.000001, 0.01, 0, 0.2 & 1.9,-0.1,1,10 \\
ringnorm & 0.000001,0.000001 & 0.1,0.01,0.001 & -1,0.001,1 & \multicolumn{1}{l}{\hspace{2.5cm}$\_\_\_$} & 1.9,2.8,0.000001,0.1 \\
twonorm & 0.01,1 & 0,0.001,0.0001 & -1,0.001,1 & \multicolumn{1}{l}{\hspace{2.5cm}$\_\_\_$} & 1.5,4.3,0.000001,0.1 \\
mushroom & 1,1 & 0,1,10 & -1,0.001,1 & \multicolumn{1}{l}{\hspace{2.5cm}$\_\_\_$} & 0.1,-2,1000,10 \\
musk\_2 & 0.000001,0.000001 & 0,0.001,0.001 & -2,0.001,0.001 & \multicolumn{1}{l}{\hspace{2.5cm}$\_\_\_$} & 0.1,-2,0.000001,0.000001 \\
EEG Eye State & 0.001,10 & 0,0.001,10 & -10,10,1 & \multicolumn{1}{l}{\hspace{2.5cm}$\_\_\_$} & 1.7,4.1,0.00001,0.01 \\ \hline
\multicolumn{6}{l}{$^{\dagger}$ represents the proposed model.}
\end{tabular}}
\end{table}
\begin{table}[htp]
\centering
\caption{Accuracy for non-linear Wave-TSVM against baseline models on UCI and KEEL datasets.}
\label{tab:non-linear Twin Table}

\resizebox{\textwidth}{!}{%
\begin{tabular}{lccccc}
\hline
Dataset & TSVM \cite{khemchandani2007twin} & Pin-GTSVM \cite{tanveer2019general} & IF-RVFL \cite{malik2022alzheimer} & SLSTSVM \cite{si2023symmetric} & Wave-TSVM$^{\dagger}$ \\ \hline
breast\_cancer & 100 & 100 & 100 & 98.59 & 100 \\
breast\_cancer\_wisc & 88.01 & 88.68 & 83.57 & 78.16 & 87.93 \\
congressional\_voting & 63.3 & 64.04 & 65.14 & 65.14 & 66.06 \\
credit\_approval & 91.84 & 90.18 & 90.92 & 83.82 & 91.91 \\
echocardiogram & 87.2 & 80.74 & 85.63 & 87.5 & 87.5 \\
haberman\_survival & 80.52 & 67.97 & 80.52 & 79.22 & 81.82 \\
hepatitis & 82.31 & 84.29 & 76.67 & 94.87 & 97.44 \\
horse\_colic & 75.04 & 83.48 & 77.61 & 73.91 & 79.35 \\
monks\_1 & 73.38 & 78.48 & 74.82 & 89.93 & 87.77 \\
pima & 77.69 & 75.4 & 70.1 & 71.88 & 72.4 \\
pittsburg\_bridges\_T\_OR\_D & 96.15 & 95.83 & 93.08 & 96 & 96.15 \\
spect & 75.76 & 77.5 & 63.64 & 68.18 & 73.13 \\
statlog\_heart & 65.07 & 66.47 & 66.72 & 67.65 & 67.65 \\
vertebral\_column\_2clases & 100 & 100 & 100 & 100 & 100 \\
abalone9-18 & 100 & 88.01 & 90.48 & 100 & 100 \\
aus & 76.95 & 78.37 & 75.93 & 78.03 & 77.91 \\
cmc & 100 & 67.08 & 100 & 93.22 & 100 \\
ecoli0137vs26 & 93.12 & 91.44 & 91.82 & 92.31 & 94.87 \\
ecoli01vs5 & 94.35 & 90.36 & 86.67 & 95 & 95 \\
ecoli3 & 91.4 & 89.4 & 89.29 & 92.86 & 92.86 \\
led7digit-0-2-4-5-6-7-8-9\_vs\_1 & 95.3 & 93.31 & 89.37 & 93.69 & 96.4 \\
monk1 & 55.4 & 52.76 & 61.15 & 56.83 & 56.83 \\
monk3 & 57.55 & 56.58 & 56.83 & 56.83 & 55.4 \\
shuttle-6\_vs\_2-3 & 100 & 100 & 98.28 & 100 & 100 \\
shuttle-c0-vs-c4 & 95.76 & 95.76 & 94.09 & 97.16 & 96.72 \\
transfusion & 87.17 & 83.87 & 87.17 & 87.17 & 87.7 \\
votes & 90.17 & 90.48 & 78.53 & 79.82 & 92.66 \\
vowel & 92.19 & 96.43 & 95.08 & 91.5 & 95.14 \\
yeast-0-2-5-6\_vs\_3-7-8-9 & 94.42 & 84.6 & 94.02 & 93.63 & 93.23 \\
yeast-0-3-5-9\_vs\_7-8 & 95.24 & 81.62 & 88.9 & 95.24 & 95.24 \\
yeast1vs7 & 93.1 & 82.03 & 85.65 & 94.78 & 94.78 \\
yeast2vs8 & 99.17 & 88.66 & 89.92 & 98.35 & 99.17 \\ \hline
\textbf{Avg. Acc.} & 86.49 & 83.24 & 83.8 & 85.98 & \textbf{87.91} \\ \hline
\textbf{Avg. Rank} & 2.78 & 3.55 & 3.75 & 2.92 & \textbf{2} \\ \hline
\multicolumn{6}{l}{$^{\dagger}$ represents the proposed model.}\\
\multicolumn{6}{l}{Here, Avg. and Acc. are acronyms used for average and accuracy, respectively.}
\end{tabular}}
\end{table}
\begin{table}[htp]
\centering
\caption{The optimal parameters of the proposed Wave-TSVM and baseline models corresponding to accuracy values on UCI and KEEL datasets.}
\label{tab:optimal parameters twin model on UCI and KEEL datasets}

\resizebox{14.5cm}{!}{
\begin{tabular}{lccccc}
\hline
Dataset & TSVM \cite{khemchandani2007twin} & Pin-GTSVM \cite{tanveer2019general} & IF-RVFL \cite{malik2022alzheimer} & SLSTSVM \cite{si2023symmetric} & Wave-TSVM$^{\dagger}$ \\
 & ($C_1$, $C_3$, $\sigma$) & ($C_1$, $C_3$, $\sigma$, $\tau$) & ($C$,$ N_d$, $\sigma$) & ($C_1$, $C_2$, $\sigma$) & (a, $C_1$, $C_2$, $\sigma$) \\ \hline
breast\_cancer & 0.000001, 0.000001, 0.000001 & 10000, 0.000001, 100, 0.5 & 0.000001, 23, 0.000001 & 1000000, 100, 0.000001 & -2, 0.000001, 0.000001, 0.000001 \\
breast\_cancer\_wisc & 0.000001, 0.000001, 100 & 0.000001, 0.000001, 0.000001, 0.5 & 0.000001, 23, 0.000001 & 1000000, 100, 0.000001 & 0, 0.0001, 1, 0.000001 \\
congressional\_voting & 0.000001, 0.000001, 10000 & 10000, 0.000001, 0.01, 0.5 & 0.01, 143, 100 & 1, 1, 0.000001 & 1.5, 0.01, 1, 0.000001 \\
credit\_approval & 1, 0.000001, 100 & 1, 0.000001, 1, 0.5 & 0.000001, 83, 100 & 0.000001, 0.000001, 0.000001 & 1, 0.01, 1, 0.000001 \\
echocardiogram & 0.000001, 1, 1 & 0.000001, 0.000001, 1, 0.8 & 100, 183, 1000000 & 10000, 100, 0.000001 & 1, 0.0001, 1, 0.000001 \\
haberman\_survival & 1, 0.000001, 1 & 10000, 0.000001, 10000, 0.5 & 100, 3, 0.01 & 0.000001, 1, 0.000001 & -1.5, 0.01, 100, 0.000001 \\
hepatitis & 0.000001, 0.000001, 0.000001 & 1, 0.000001, 1000000, 1 & 0.000001, 203, 0.000001 & 0.000001, 1, 0.000001 & 0, 0.000001, 0.01, 0.000001 \\
horse\_colic & 1, 0.000001, 100 & 0.000001, 0.000001, 0.000001, 0.5 & 0.000001, 43, 0.000001 & 100, 1, 0.000001 & -0.5, 0.000001, 1, 0.000001 \\
monks\_1 & 1, 0.000001, 100 & 10000, 0.000001, 0.01, 0.8 & 1000000, 83, 100 & 0.000001, 100, 0.000001 & 2, 0.000001, 1, 0.000001 \\
pima & 0.000001, 0.000001, 100 & 0.000001, 0.000001, 10000, 0.5 & 0.000001, 3, 1 & 100, 1, 0.000001 & -2, 0.000001, 0.000001, 0.000001 \\
pittsburg\_bridges\_T\_OR\_D & 1, 0.000001, 100 & 0.000001, 0.000001, 0.000001, 0.5 & 0.000001, 23, 0.000001 & 0.000001, 0.01, 0.000001 & 1, 0.0001, 1, 0.000001 \\
spect & 0.000001, 0.000001, 100 & 0.000001, 0.000001, 1000000, 0.5 & 1, 3, 1 & 100, 1, 0.000001 & -1, 1, 1, 0.000001 \\
statlog\_heart & 100, 100, 100 & 0.000001, 0.000001, 1000000, 0.8 & 1000000, 123, 10000 & 0.000001, 100, 0.000001 & -2, 0.000001, 0.000001, 0.000001 \\
vertebral\_column\_2clases & 0.000001, 0.000001, 0.000001 & 0.000001, 0.000001, 0.000001, 0.5 & 0.000001, 63, 1 & 0.0001, 1, 0.000001 & 0, 0.0001, 1, 0.000001 \\
abalone9-18 & 0.000001, 0.000001, 0.000001 & 100, 0.000001, 0.01, 0.5 & 1000000, 163, 1000000 & 0.01, 100, 0.000001 & 1.5, 0.01, 1, 0.000001 \\
aus & 1, 0.000001, 100 & 0.000001, 0.000001, 0.000001, 0.5 & 0.000001, 23, 1 & 0.000001, 100, 0.000001 & 1, 0.01, 1, 0.000001 \\
cmc & 0.000001, 0.000001, 0.000001 & 0.000001, 0.000001, 100, 1 & 0.000001, 3, 0.000001 & 1000000, 100, 0.000001 & 1, 0.0001, 1, 0.000001 \\
ecoli0137vs26 & 0.000001, 0.000001, 0.000001 & 1, 0.000001, 1, 0.5 & 0.000001, 143, 1 & 10000, 10000, 0.000001 & -1.5, 0.01, 100, 0.000001 \\
ecoli01vs5 & 0.0001, 0.000001, 1000000 & 100, 0.000001, 10000, 1 & 0.000001, 83, 1 & 0.000001, 0.000001, 0.000001 & 0, 0.000001, 0.01, 0.000001 \\
ecoli3 & 1, 0.000001, 100 & 100, 0.000001, 1, 0.8 & 0.000001, 143, 1 & 1, 1, 0.000001 & -0.5, 0.000001, 1, 0.000001 \\
led7digit-0-2-4-5-6-7-8-9\_vs\_1 & 1, 0.000001, 100 & 1000000, 0.000001, 0.01, 0.5 & 0.000001, 23, 0.000001 & 0.000001, 1, 0.000001 & 2, 0.000001, 1, 0.000001 \\
monk1 & 0.000001, 0.000001, 100 & 100, 0.000001, 10000, 1 & 100, 83, 1000000 & 0.000001, 100, 0.000001 & -2, 0.000001, 0.000001, 0.000001 \\
monk3 & 0.0001, 0.000001, 1000000 & 100, 0.000001, 10000, 1 & 100, 43, 10000 & 0.000001, 100, 0.000001 & 1, 0.0001, 1, 0.000001 \\
shuttle-6\_vs\_2-3 & 0.000001, 0.000001, 100 & 0.000001, 0.000001, 0.000001, 0.5 & 0.000001, 43, 0.01 & 0.000001, 0.000001, 0.000001 & -1, 1, 1, 0.000001 \\
shuttle-c0-vs-c4 & 0.000001, 0.000001, 1 & 100, 0.000001, 1, 0.8 & 0.000001, 63, 0.01 & 100, 1, 0.000001 & -1.5, 0.000001, 100, 0.000001 \\
transfusion & 0.000001, 0.000001, 0.000001 & 0.000001, 0.000001, 0.01, 0.5 & 0.000001, 3, 0.01 & 0.000001, 100, 0.000001 & -2, 0.000001, 0.000001, 0.000001 \\
votes & 100, 1, 100 & 0.000001, 0.000001, 0.01, 0.5 & 0.000001, 183, 0.000001 & 100, 100, 0.000001 & -2, 0.000001, 0.000001, 0.000001 \\
vowel & 100, 100, 100 & 1, 0.000001, 0.01, 1 & 1000000, 163, 1 & 100, 1, 0.000001 & -0.5, 0.01, 1, 0.000001 \\
yeast-0-2-5-6\_vs\_3-7-8-9 & 1, 0.000001, 100 & 0.000001, 0.000001, 0.01, 1 & 0.000001, 3, 0.000001 & 0.000001, 1, 0.000001 & -2, 0.000001, 0.000001, 0.000001 \\
yeast-0-3-5-9\_vs\_7-8 & 1, 0.000001, 100 & 0.000001, 0.000001, 0.01, 0.8 & 1000000, 183, 100 & 0.000001, 0.000001, 0.000001 & 0.5, 100, 1, 0.000001 \\
yeast1vs7 & 1, 0.000001, 100 & 1, 0.000001, 0.01, 1 & 10000, 183, 1000000 & 0.000001, 0.000001, 0.000001 & -2, 0.000001, 0.000001, 0.000001 \\
yeast2vs8 & 1, 0.000001, 100 & 1000000, 0.000001, 0.01, 0.5 & 1000000, 163, 1000000 & 0.000001, 0.000001, 0.000001 & 0, 0.0001, 1, 0.000001 \\ \hline
\multicolumn{6}{l}{$^{\dagger}$ represents the proposed model.}
\end{tabular}}
\end{table}
\begin{table}[htp]
\centering
\caption{The optimal parameters of the proposed non-linear Wave-SVM and baseline models  corresponding to accuracy values on noisy datasets.}
\label{tab:non-linearWave-SVM-optimal-parameters-on-noisy-datasets-table}
\resizebox{\textwidth}{!}{
\begin{tabular}{lccccc}
\hline
Model & C-SVM \cite{cortes1995support} & Pin-SVM \cite{huang2013support} & LINEX-SVM \cite{ma2019linex} & FP-SVM \cite{kumari2024diagnosis} & Wave-SVM$^{\dagger}$ \\
Dataset & ($C$, $\sigma$) & ($\tau$, $C$, $\sigma$) & ($a$, $C$, $\sigma$) & ($C$, $\sigma$, $\tau_1$, $\tau_2$) & ($\lambda$, $a$, $C$, $\sigma$) \\ \hline
\multicolumn{6}{c}{5\% noise} \\ \hline
breast\_cancer\_wisc & 0.000001, 0.000001 & 0.000001, 0.000001, 0 & -3, 0.000001, 1 & 0.000001, 1, 0, 0.1 & -2, 1.5, 0.000001, 0.1 \\
credit\_approval & 1, 1 & 1, 1, 1 & -3, 0.000001, 1 & 0.0001, 0.01, 0, 0.5 & -2, 1.7, 0.00001, 0.1 \\
horse\_colic & 1, 1 & 100, 1, 0 & -3, 0.000001, 1 & 1000000, 1, 0, 0.7 & -0.1, 1.9, 0.000001, 1 \\
led7digit-0-2-4-5-6-7-8-9\_vs\_1 & 1, 1 & 1, 1, 0 & -5, 0.000001, 1 & 0.000001, 0.01, 0, 0.3 & 0.1, 0.7, 0.001, 10 \\
monk1 & 0.01, 1 & 0.000001, 0.000001, 0 & -1, 0.000001, 1 & 10000, 1, 0, 0.1 & 0.8, 1.7, 1000000, 10 \\
yeast2vs8 & 0.000001, 0.000001 & 0.000001, 0.000001, 0 & -7, 0.000001, 0.000001 & 0.000001, 0.01, 0, 0.2 & -2, 0.3, 100, 100 \\ \hline
\multicolumn{6}{c}{10\% noise} \\ \hline
breast\_cancer\_wisc & 1, 1 & 0.01, 1, 0 & -5, 0.000001, 1 & 0.000001, 1, 0.1, 0.1 & -0.7, 0.1, 10000, 10 \\
credit\_approval & 1, 1 & 1, 1, 1 & -2, 0.0001, 1 & 0.0001, 1, 0.5, 0 & 2.1, 0.7, 0.000001, 0.1 \\
horse\_colic & 1, 1 & 100, 1, 0 & -4, 0.000001, 1 & 0.01, 1, 0, 0.8 & -0.7, 1.5, 0.000001, 1 \\
led7digit-0-2-4-5-6-7-8-9\_vs\_1 & 1, 1 & 1, 1, 0 & -7, 0.000001, 0.000001 & 0.000001, 1, 0.5, 0 & -1.9, 0.3, 1000000, 10 \\
monk1 & 0.01, 1 & 1, 1, 0 & -5, 0.000001, 0.01 & 0.0001, 1, 0, 0.8 & 4.3, 0.7, 0.001, 10 \\
yeast2vs8 & 0.000001, 0.000001 & 0.000001, 0.000001, 0 & -7, 0.000001, 0.000001 & 0.000001, 0.01, 0, 0.2 & 0.6, 1.9, 10, 10 \\ \hline
\multicolumn{6}{c}{20\% noise} \\ \hline
breast\_cancer\_wisc & 1, 1 & 0.01, 1, 0 & -5, 0.000001, 1 & 0.0001, 1, 0, 0.2 & 1.9, 1.3, 0.000001, 0.1 \\
credit\_approval & 1, 1 & 1, 1, 0.1 & -6, 0.000001, 1 & 100, 1, 0, 0.1 & 3.7, 1.5, 0.00001, 1 \\
horse\_colic & 100, 1 & 1, 1, 0 & -5, 0.000001, 1 & 10000, 1, 0, 0.8 & 2.7, 0.3, 0.00001, 1 \\
led7digit-0-2-4-5-6-7-8-9\_vs\_1 & 1, 1 & 1, 1, 0 & -5, 0.000001, 1 & 100, 1, 0, 1 & -2, 0.1, 100, 10 \\
monk1 & 0.01, 1 & 0.01, 1, 0 & -6, 0.000001, 1 & 0.000001, 1, 0.7, 0 & 3.1, 0.1, 100000, 10 \\
yeast2vs8 & 0.000001, 0.000001 & 0.000001, 0.000001, 0 & -7, 0.000001, 0.000001 & 0.000001, 0.01, 0, 0.2 & -1.5, 0.5, 100000, 10 \\ \hline
\multicolumn{6}{c}{30\% noise} \\ \hline
breast\_cancer\_wisc & 100, 1 & 0.01, 1, 0 & -1, 0.000001, 1 & 0.000001, 1, 0.5, 0 & -0.9, 0.1, 10000, 10 \\
credit\_approval & 100, 1 & 1, 1, 0.5 & -4, 0.000001, 1 & 0.0001, 0.01, 0, 0.9 & -1.2, 0.9, 0.000001, 0.1 \\
horse\_colic & 0.000001, 0.000001 & 1, 1, 0 & -2, 0.000001, 1 & 10000, 1, 0, 0.7 & -2, 1.9, 0.00001, 1 \\
led7digit-0-2-4-5-6-7-8-9\_vs\_1 & 1, 1 & 1, 1, 0 & -5, 0.000001, 1 & 1, 0.01, 0, 0.1 & 2.7, 1.5, 0.01, 10 \\
monk1 & 1, 1 & 0.01, 1, 0 & -3, 0.000001, 1 & 100, 1, 0, 0.9 & 1.1, 0.5, 10000, 10 \\
yeast2vs8 & 0.000001, 0.000001 & 0.000001, 0.000001, 0 & -7, 0.000001, 0.000001 & 0.000001, 0.01, 0, 0.2 & -1.6, 0.1, 1000, 10 \\ \hline
\multicolumn{6}{l}{$^{\dagger}$ represents the proposed model.}
\end{tabular}}
\end{table}
\begin{table}[htp]
\centering
\caption{The optimal parameters of the proposed Wave-TSVM and baseline models corresponding to accuracy values on noisy datasets.}
\label{tab:Optimal parameters Twin models on noisy datasets}
\resizebox{14.5cm}{!}{
\begin{tabular}{lccccc}
\hline
Model & TSVM \cite{khemchandani2007twin} & Pin-GTSVM \cite{tanveer2019general} & IF-RVFL \cite{malik2022alzheimer} & SLSTSVM \cite{si2023symmetric} & Wave-TSVM$^{\dagger}$ \\
Dataset & ($C_1$, $C_3$, $\sigma$) & ($C_1$, $C_3$, $\sigma$, $\tau$) & ($C$,$ N_d$, $\sigma$) & ($C_1$, $C_2$, $\sigma$) & (a, $C_1$, $C_2$, $\sigma$) \\ \hline
\multicolumn{6}{c}{5\% noise} \\ \hline
breast\_cancer\_wisc & 0.000001, 0.000001, 100 & 1, 0.000001, 1, 0.5 & 0.000001, 23, 0.000001 & 1000000, 100, 0.000001 & 0, 0.000001, 1, 0.000001 \\
credit\_approval & 1, 0.000001, 100 & 1, 0.000001, 1, 1 & 0.000001, 83, 100 & 1, 0.000001, 0.000001 & -0.5, 0.01, 1, 0.000001 \\
horse\_colic & 1, 0.000001, 100 & 0.000001, 0.000001, 1000000, 0.5 & 0.000001, 43, 0.000001 & 1, 10000, 0.000001 & -0.5, 1, 1, 0.000001 \\
led7digit-0-2-4-5-6-7-8-9\_vs\_1 & 1, 1, 1 & 1000000, 0.000001, 0.01, 0.8 & 0.000001, 23, 0.000001 & 100, 1, 0.000001 & 0.5, 100, 1, 0.000001 \\
monk1 & 0.000001, 10000, 1000000 & 10000, 0.000001, 10000, 0.5 & 100, 83, 1000000 & 0.000001, 100, 0.000001 & 1, 1, 100, 0.000001 \\
yeast2vs8 & 1, 0.000001, 100 & 10000, 0.000001, 0.0001, 0.5 & 1000000, 163, 1000000 & 1000000, 1000000, 0.000001 & -2, 0.000001, 0.000001, 0.000001 \\ \hline
\multicolumn{6}{c}{10\% noise} \\ \hline
breast\_cancer\_wisc & 0.000001, 0.000001, 100 & 0.000001, 0.000001, 1, 0.5 & 0.000001, 23, 0.000001 & 1000000, 100, 0.000001 & 1, 0.01, 1, 0.000001 \\
credit\_approval & 1, 0.000001, 100 & 1, 0.000001, 1, 0.5 & 0.000001, 83, 100 & 0.000001, 100, 0.000001 & -0.5, 0.000001, 1, 0.000001 \\
horse\_colic & 1, 0.000001, 100 & 100, 0.000001, 0.01, 0.8 & 0.000001, 43, 0.000001 & 0.000001, 10000, 0.000001 & -1, 0.01, 1, 0.000001 \\
led7digit-0-2-4-5-6-7-8-9\_vs\_1 & 1, 0.000001, 100 & 1000000, 0.000001, 0.01, 1 & 0.000001, 23, 0.000001 & 100, 1, 0.000001 & 0.5, 100, 1, 0.000001 \\
monk1 & 0.01, 0.01, 1000000 & 1000000, 0.000001, 1000000, 0.5 & 100, 83, 1000000 & 0.000001, 0.000001, 0.000001 & -1.5, 0.000001, 0.000001, 0.000001 \\
yeast2vs8 & 1, 0.000001, 100 & 100, 0.000001, 0.01, 0.5 & 1000000, 163, 1000000 & 0.000001, 100, 0.000001 & 0, 0.000001, 1, 0.000001 \\ \hline
\multicolumn{6}{c}{20\% noise} \\ \hline
breast\_cancer\_wisc & 0.000001, 0.000001, 100 & 0.000001, 0.000001, 1, 0.5 & 0.000001, 23, 0.000001 & 10000, 1, 0.000001 & -1, 0.000001, 100, 0.000001 \\
credit\_approval & 0.000001, 0.000001, 100 & 1, 0.000001, 1, 0.5 & 0.000001, 83, 100 & 0.000001, 10000, 0.000001 & 0, 1, 100, 0.000001 \\
horse\_colic & 1, 0.000001, 100 & 100, 0.000001, 0.01, 0.8 & 0.000001, 43, 0.000001 & 100, 1, 0.000001 & -0.5, 0.01, 1, 0.000001 \\
led7digit-0-2-4-5-6-7-8-9\_vs\_1 & 0.000001, 0.000001, 1 & 1000000, 0.000001, 0.01, 1 & 0.000001, 23, 0.000001 & 0.000001, 1, 0.000001 & 1, 0.0001, 1, 0.000001 \\
monk1 & 10000, 0.0001, 1000000 & 1000000, 0.000001, 1000000, 0.5 & 100, 83, 1000000 & 10000, 0.000001, 0.000001 & 1, 1000000, 1000000, 0.000001 \\
yeast2vs8 & 1, 0.000001, 100 & 100, 0.000001, 0.01, 0.5 & 1000000, 163, 1000000 & 0.000001, 1, 0.000001 & -2, 0.000001, 0.000001, 0.000001 \\ \hline
\multicolumn{6}{c}{30\% noise} \\ \hline
breast\_cancer\_wisc & 100, 1, 100 & 0.000001, 0.000001, 0.01, 0.5 & 0.000001, 23, 0.000001 & 10000, 1, 0.000001 & 0, 1, 100, 0.000001 \\
credit\_approval & 0.000001, 0.000001, 100 & 1, 0.000001, 1, 0.8 & 0.000001, 83, 100 & 10000, 0.01, 0.000001 & -1, 0.01, 1, 0.000001 \\
horse\_colic & 1, 0.000001, 100 & 0.000001, 0.000001, 0.01, 0.8 & 0.000001, 43, 0.000001 & 0.000001, 100, 0.000001 & 0, 0.0001, 1, 0.000001 \\
led7digit-0-2-4-5-6-7-8-9\_vs\_1 & 0.000001, 1, 1 & 10000, 0.000001, 0.01, 0.8 & 0.000001, 23, 0.000001 & 0.000001, 1, 0.000001 & -1.5, 0.01, 1, 0.000001 \\
monk1 & 10000, 0.01, 1000000 & 1, 0.000001, 1, 1 & 100, 83, 1000000 & 100, 10000, 0.000001 & 0.5, 0.000001, 100, 0.000001 \\
yeast2vs8 & 1, 0.000001, 100 & 100, 0.000001, 0.01, 0.5 & 1000000, 163, 1000000 & 0.000001, 1, 0.000001 & -2, 0.000001, 0.000001, 0.000001 \\ \hline
\multicolumn{6}{l}{$^{\dagger}$ represents the proposed model.}
\end{tabular}}
\end{table}

\clearpage
\bibliography{refs.bib}
 \bibliographystyle{unsrtnat}